\pdfoutput=1

\documentclass[11pt]{article}
\usepackage[dvipsnames,svgnames,x11names]{xcolor} 

\usepackage[final]{acl}

\usepackage{times}
\usepackage{latexsym}

\usepackage[T1]{fontenc}

\usepackage[utf8]{inputenc}

\usepackage{microtype}

\usepackage{inconsolata}

\usepackage{graphicx}

\usepackage{amsthm}
\theoremstyle{plain}

\theoremstyle{definition}
\newtheorem{assumption}{Assumption}

\theoremstyle{remark}

\usepackage[most]{tcolorbox}
\usepackage{multicol}

 \makeatletter
\def\@fnsymbol#1{\ensuremath{\ifcase#1\or \dagger\or \ddagger\or
  \mathsection\or \mathparagraph\or \|\or **\or \dagger\dagger
  \or \ddagger\ddagger \else\@ctrerr\fi}}
\makeatother

\usepackage{dsfont}

\usepackage{multirow}
\usepackage{amsthm,amssymb,lipsum}
\usepackage{mathrsfs}
\usepackage{enumerate}
\usepackage{colortbl}
\usepackage{wrapfig}
\usepackage{bbding}
\usepackage{pifont}
\usepackage{xspace}
\usepackage{amsmath}
\usepackage{wasysym}
\usepackage{textcomp}
\usepackage[bottom]{footmisc}

\usepackage{makecell}

\usepackage{color}
\usepackage{tocloft}
\usepackage{caption}
\usepackage{float}
\usepackage{afterpage}

\usepackage{stackengine}

\usepackage[commandnameprefix=always]{changes}

\usepackage{bbm}

\usepackage{cuted}
\usepackage{duckuments}

\usepackage{tabularx}
\usepackage{booktabs}
\usepackage[percent]{overpic}
\usepackage{adjustbox}
\usepackage{utfsym}

\usepackage{epigraph}

\def\eg{{\it e.g.}\xspace}

\def\ie{{\it i.e.}\xspace}
\def\etc{{\it etc}\xspace}

\definecolor{drp-blue}{HTML}{1f77b4}
\definecolor{pretty-blue}{RGB}{0, 113, 188}
\definecolor{kaiming-green}{RGB}{57,181,74} 
\definecolor{mypurple}{RGB}{55,0,168} 
\definecolor{icmlblue}{rgb}{0,0.08,0.45} 
\definecolor{linecolor1}{HTML}{F1F7FB}
\definecolor{linecolor2}{HTML}{E3EFF7}
\definecolor{linecolor3}{HTML}{D5E4F0}

\definecolor{reconcolor}{HTML}{412F8A}
\definecolor{runpei-orange}{HTML}{F35F27}
\definecolor{runpei_blue}{HTML}{14294B}
\definecolor{datacolor}{HTML}{0009BF}
\definecolor{vitcolor}{HTML}{fc8e62}
\definecolor{xycolor}{HTML}{EF98AA}
\definecolor{lightpink}{HTML}{F9F0EE}
\definecolor{deeppink}{HTML}{EBD9E4}
\definecolor{lightpurple}{RGB}{123,107,143}
\definecolor{lightyellow}{RGB}{200,180,120}
\definecolor{MorandiLightBlue}{RGB}{150,180,220}
\definecolor{MorandiLighterBlue}{RGB}{230,240,250}
\definecolor{MorandiPink}{RGB}{200, 160, 180}
\definecolor{MorandiLightPink}{RGB}{250, 240, 245}

\newcommand{\pwait}{{\bs{p}_{\text{\texttt{wait}}}}}

\newcommand{\bs}[1]{\boldsymbol{#1}}
\def\baselineexp{{\scshape Base}\xspace}
\def\codexp{{\scshape CoD}\xspace}
\def\sone{{\scshape s1}\xspace}

\usepackage{enumitem}

\usepackage[capitalize]{cleveref} 
\usepackage{appendix}
\usepackage{titletoc}

\def\ours{{\scshape AlphaOne}\xspace}
\def\oursbf{{\scshape \textbf{AlphaOne}}\xspace}
\def\oursshort{{$\alpha$1}\xspace}
\def\oursshortbold{{$\bs\alpha$\textbf{1}}\xspace}
\def\oursshortexp{{$\bs\alpha$\textbf{1 (Ours)}}\xspace}

\title{\ours: Reasoning Models Thinking Slow and Fast at Test Time}

\author{
 \textbf{Junyu Zhang}\raisebox{1.0ex}{\includegraphics[height=1.1ex]{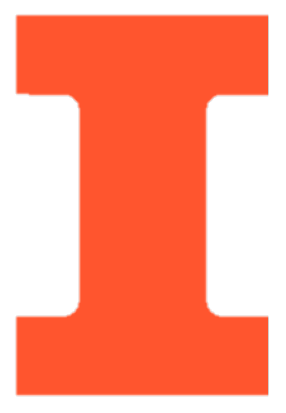}}$^\dagger$\qquad
 \textbf{Runpei Dong}\raisebox{1.0ex}{\includegraphics[height=1.1ex]{figures/src/uiuc_logo}}$^\dagger$\qquad
 \textbf{Han Wang}\raisebox{1.0ex}{\includegraphics[height=1.1ex]{figures/src/uiuc_logo}}\qquad
 \textbf{Xuying Ning}\raisebox{1.0ex}{\includegraphics[height=1.1ex]{figures/src/uiuc_logo}}\qquad
\\
 \textbf{Haoran Geng}\raisebox{1.0ex}{\includegraphics[height=1.1ex]{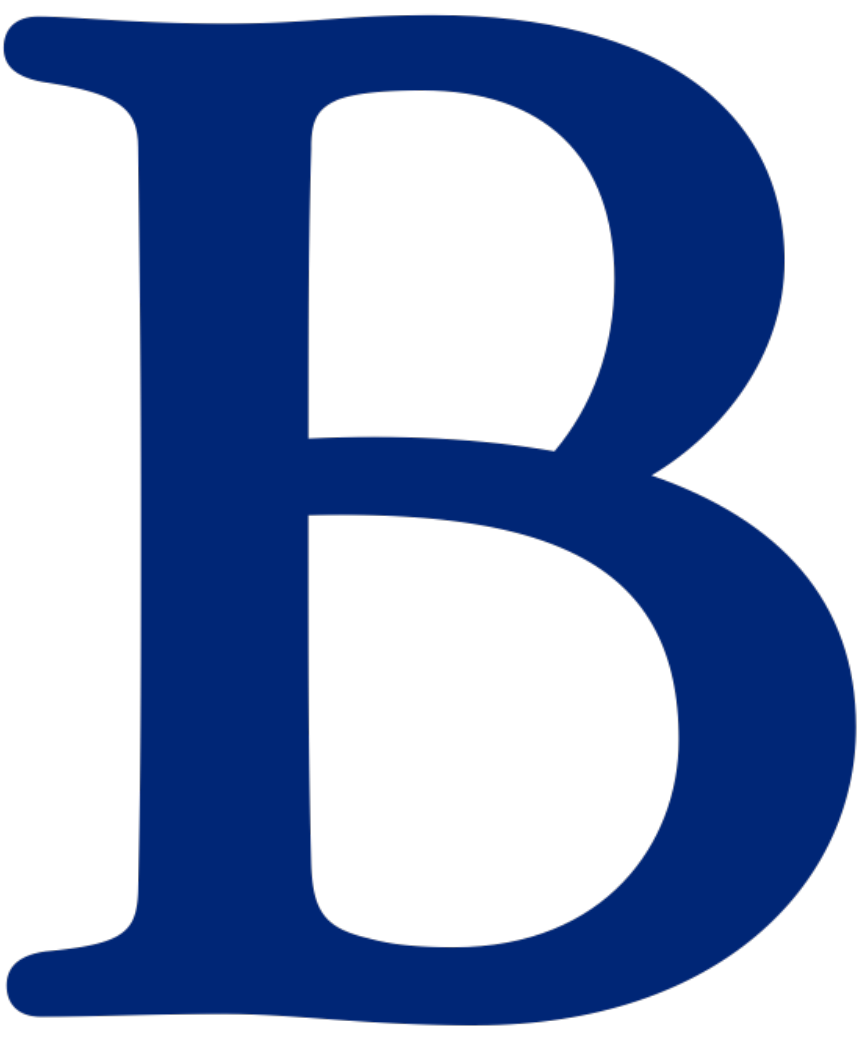}}\qquad
\textbf{Peihao Li}\raisebox{1.0ex}{\includegraphics[height=1.1ex]{figures/src/berkeley_logo}}\qquad
 \textbf{Xialin He}\raisebox{1.0ex}{\includegraphics[height=1.1ex]{figures/src/uiuc_logo}}\qquad
 \textbf{Yutong Bai}\raisebox{1.0ex}{\includegraphics[height=1.1ex]{figures/src/berkeley_logo}}\qquad
\\
 \textbf{Jitendra Malik}\raisebox{1.0ex}{\includegraphics[height=1.1ex]{figures/src/berkeley_logo}}\qquad
 \textbf{Saurabh Gupta}\raisebox{1.0ex}{\includegraphics[height=1.1ex]{figures/src/uiuc_logo}}\qquad
 \textbf{Huan Zhang}\raisebox{1.0ex}{\includegraphics[height=1.1ex]{figures/src/uiuc_logo}}
\\[0.7em]
 \raisebox{1.0ex}{\includegraphics[height=1.1ex]{figures/src/uiuc_logo}}University of Illinois Urbana-Champaign\qquad
 \raisebox{1.0ex}{\includegraphics[height=1.1ex]{figures/src/berkeley_logo}}UC Berkeley\qquad
\\
 \small{
    \textbf{$^\dagger$Equal contributions.}
 }
 \small{
 \textbf{Correspondence:} 
 \texttt{\{junyuz6, runpeid2, huanz\}@illinois.edu}
 }
}

\begin{document}
\maketitle

\begin{abstract}
This paper presents \ours ($\alpha$1), a universal framework for modulating reasoning progress in large reasoning models (LRMs) at test time. 
\oursshort first introduces $\alpha$ moment, which represents the scaled thinking phase with a universal parameter $\alpha$.
Within this scaled pre-$\alpha$ moment phase, it dynamically schedules slow thinking transitions by modeling the insertion of reasoning transition tokens as a Bernoulli stochastic process. 
After the $\alpha$ moment, $\alpha$1 deterministically terminates slow thinking with the end-of-thinking token, thereby fostering fast reasoning and efficient answer generation. 
This approach unifies and generalizes existing monotonic scaling methods by enabling flexible and dense slow-to-fast reasoning modulation. 
Extensive empirical studies on various challenging benchmarks across mathematical, coding, and scientific domains demonstrate \oursshort's superior reasoning capability and efficiency.
Project page: 
\href{alphaone-project.github.io}{\url{https://alphaone-project.github.io/}}
\end{abstract}

\section{Introduction}
\begin{figure}[t!]
  \includegraphics[width=1.0\columnwidth]{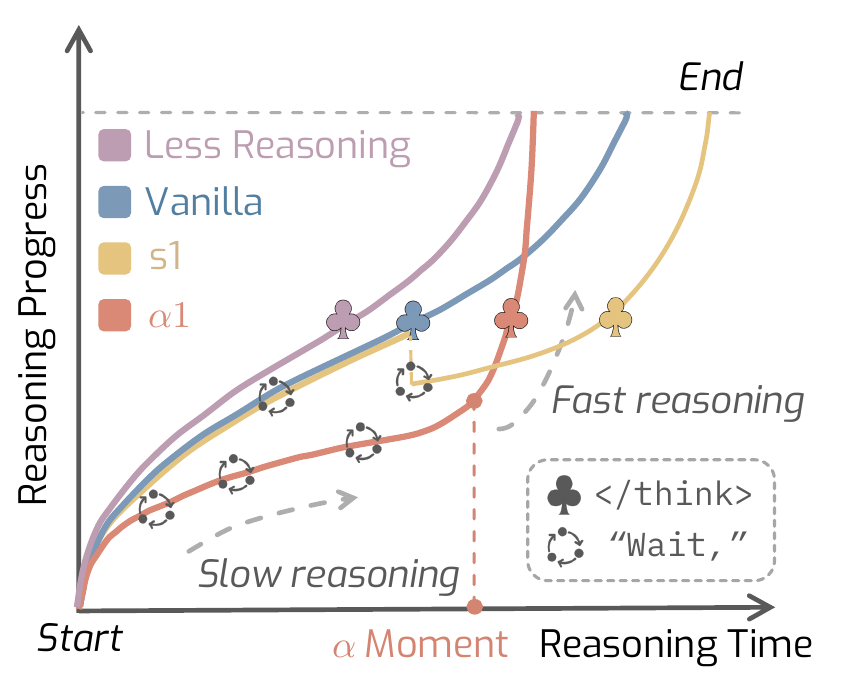}
  \vspace{-20pt}
  \caption{\textbf{Conceptual illustration} of reasoning modulation strategies. 
  Our \oursshort employs a \textit{slow-to-fast} reasoning schedule controlled by $\alpha$. 
  \oursshort scales more efficiently than \textit{monotonously increasing} method s1 (\textbf{\textcolor{lightyellow}{yellow}}) and generally outperforms \textit{monotonously decreasing} (\textbf{\textcolor{lightpurple}{purple}}) approaches.}
  \label{fig:teaser}
\end{figure}

\setlength{\epigraphwidth}{0.95\columnwidth}
\renewcommand{\epigraphflush}{center}
\renewcommand{\textflush}{flushepinormal}
\renewcommand{\epigraphsize}{\footnotesize}
\epigraph{\textcolor{black}{``The most effortful forms of slow thinking are those that require you to think fast.''}}
{\textcolor{black}{\textit{Thinking, Fast and Slow~\cite{ThinkingFastAndSlow}}}}
Large Reasoning Models (LRMs) such as OpenAI o1~\cite{OpenAIO1_24} and DeepSeek-R1~\cite{DeepSeekR125} have demonstrated unprecedented progress in approaching human-like system-2 reasoning capabilities, enabling \textit{slow thinking}—slowing down \textit{reasoning progress}\footnote{
Consider reasoning progress as a metric ranging from $0$ to $1$, indicating the start and the end of reasoning, respectively.
It increases slowly or fast at the pace of thinking.
See \cref{fig:teaser} and \cref{sec:background} for a more illustrative and detailed explanation.
}
at test time—for solving complex reasoning problems that require high-order cognitive processing.
These advanced models are trained to utilize slow thinking via reinforcement learning, enabling LRMs to slow down reasoning progress automatically.
Is such automatic slowing down of reasoning progress determined by LRMs sufficiently reliable?
According to \citet{ThinkingFastAndSlow}, humans typically think fast first and activate slow thinking when running into difficulty,
through a conscious control of system-1-to-2 reasoning transitioning, resulting in overall comprehensive but efficient reasoning.
While similar to human systems and interesting results have been observed, a lot of works have pointed out that the LRMs themselves are prone to \textit{overthinking}~\cite{OverThinkingO1Like24,StopOverthinking25,ThoughtTerminatorOverThinking25,SpeculativeThinking25} or \textit{underthinking}~\cite{BetweenUnderThinkingandOverthinking25,ThinkingOptimalScaling25,UnderthinkingO1Like25}.
This is because of the inability of LRMs to find the optimal human-like system-1-to-2 reasoning transitioning and limited reasoning capabilities, leading to unsatisfactory reasoning performance.

\begin{figure*}[ht!]
  \includegraphics[width=1.0\linewidth]{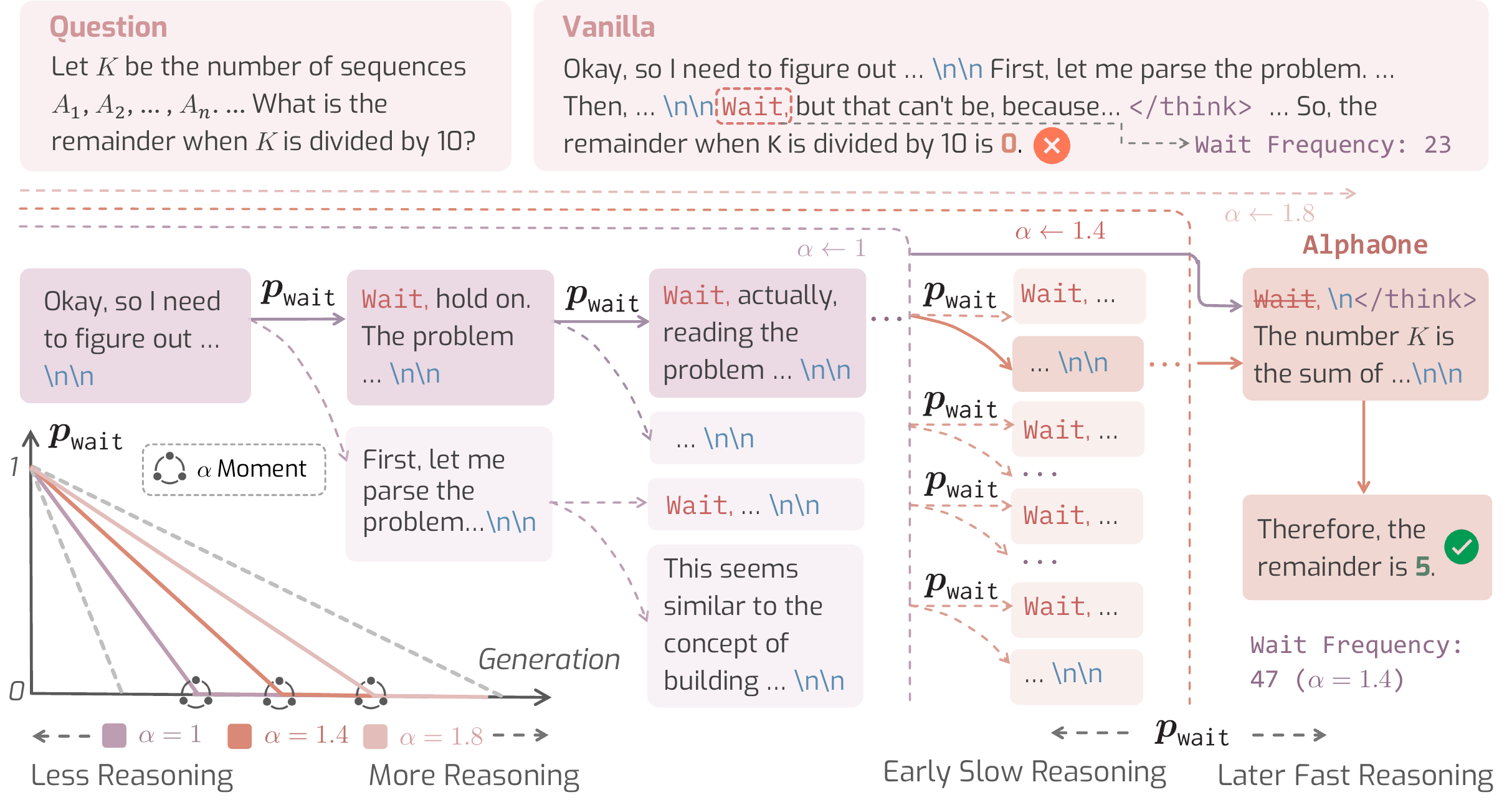}
  \caption {\textbf{Overview of \ours ($\bs \alpha$1)}. 
  Here \raisebox{-0.25em}{\includegraphics[height=1.2em]{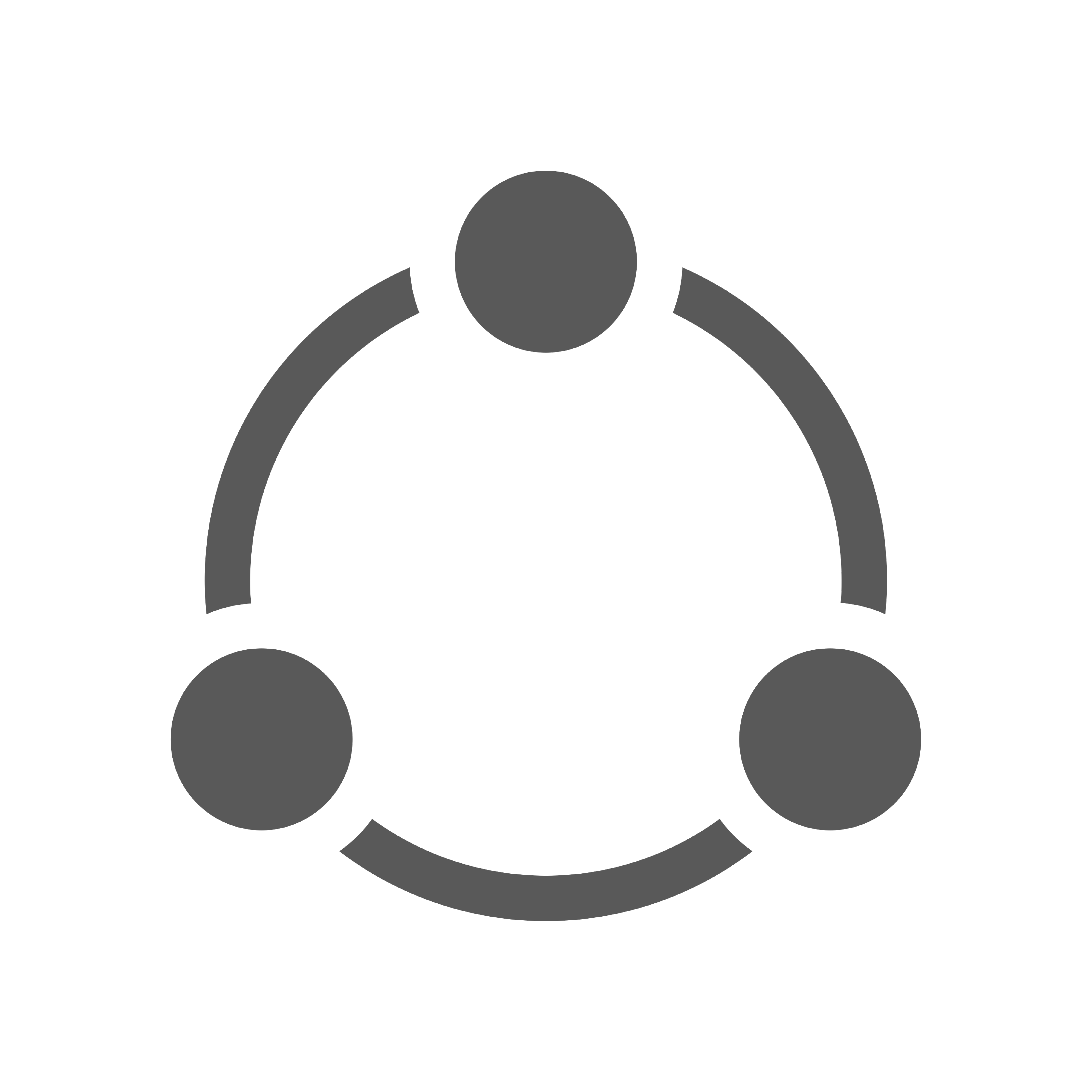}} represents $\alpha$ moment (\cref{sec:alpha_moment}). 
  \oursshort applies dense reasoning modulation via a user-defined slow thinking scheduling in pre-$\alpha$ moment. 
  In addition, \oursshort utilizes a post-$\alpha$ moment modulation by replacing slow thinking transitioning tokens ``\texttt{wait}'' to ``\texttt{</think>}'', which fosters fast thinking. Specifically, $\alpha$ determines when the slow-to-fast reasoning transition occurs. For example, reducing $\alpha$ from $1.4$ to $1.0$ shifts the $\alpha$ moment earlier, resulting in shorter slow reasoning phase and accelerating the annealing of $\pwait$.
  }\label{fig:framework}
\end{figure*}

To overcome this limitation, existing works scale LRMs at test time in mainly two ways. 
i) \textit{parallel scaling}: this line of research follows a \textit{Best of $N$} strategy and typically samples $N$ times and outputs the best answer using criteria such as self-consistency~\cite{SelfConsistency23,DynamicSelfConsistency24,InternalProbSelfConsistency25,EffectiveWithoutThinking25} and perplexity~\cite{LongContextPerplexity25}.
ii) \textit{sequential scaling}: this family of approaches addresses the overthinking/underthinking issues via early reasoning stopping~\cite{WithoutSelfDoubt25,DynamicEarlyExit25,ChainOfDraft25} and promoting for reinforcing reasoning~\cite{S1_25,UnderthinkingO1Like25}, respectively.
For example, 
\citet{ChainOfDraft25} proposes Chain of Draft, prompting LRMs to think fast strictly within 5 words to significantly reduce overthinking.
s1~\cite{S1_25} proposes to foster reasoning continuously via appending a slow-reasoning transition token ``\texttt{wait}'' multiple times when LRMs are about to end.
However, it is unclear if such monotonous reasoning increment or reduction is optimal, and the appropriate moment for slow thinking transitioning is still under-explored.
Hence, instead of test-time scaling with an automatic slowing down by LRMs themselves or simply increasing or reducing slow thinking, we are interested in finding:
\textit{Can we modulate reasoning progress universally, and develop a better slow thinking transitioning strategy with it?}

To answer this question, we present \textbf{\ours} (\oursshort), which efficiently scales LRMs at test time through a \textit{universal} reasoning progress modulation.
We introduce \textit{alpha moment}, parameterized by $\alpha\geq0$, where the thinking process is scaled by $\alpha$ times throughout the whole generation sequence.
To be specific, within a certain token length scaled by $\alpha$, we stochastically append the reasoning transition token ``\texttt{wait}'' after structural delimiters ``\texttt{\textbackslash n\textbackslash n}'' under $\mathrm{Bernoulli}(\pwait)$, inspired by the observation that these two frequently co-exist~\cite{SpeculativeThinking25}.
Here, $\pwait$ is scheduled to change over time to \textit{activate} slow thinking.
For example, a simple linear annealing over time indicates a slow thinking first, then fast thinking strategy.

However, we observe that amplifying slow thinking enables LRMs to sustain it automatically. Thus, when $\pwait$ reaches 0, we replace ``\texttt{wait}'' with ``\texttt{</think>}'' to deactivate slow thinking and switch to fast reasoning. 
In this fashion, \oursshort unifies prior methods like s1~\cite{S1_25}, where $\alpha$1 reduces to s1 if $\pwait$ is 1 or 0 at the end of a reasoning segment within a certain reasoning token length. However, different from these works that only explore sparse slow reasoning modulation, \oursshort modulates reasoning continuously, supporting both \textit{sparse} and \textit{dense} modulation strategies.

\paragraph{Takeaways} 
We present some insightful findings from evaluating three different \oursshort LRMs, ranging from 1.5B to 32B across six reasoning benchmarks, including math, code generation, and scientific problem reasoning:
i) \textit{Slow thinking first}, then fast thinking, leads to better LRM reasoning. Surprisingly, this differs from humans who commonly think fast, followed by slow thinking~\cite{ThinkingFastAndSlow}, emphasizing the requirement of dedicated test-time scaling strategies for LRMs.
ii) Slow thinking can bring \textit{efficient} test-time scaling. 
While slow thinking slows down reasoning, the overall token length is significantly reduced with $\alpha$1, inducing more informative reasoning progress brought by slow thinking.
iii) Slow thinking transitioning in \textit{high frequency} is helpful.
Interestingly, we find that \oursshort appending ``\texttt{wait}'' significantly more (\eg, over 2$\times$ more than s1) achieves much better results.
\section{Background \& Problem Statement}\label{sec:background}
\paragraph{Revisiting Reasoning Models}
Following the success of OpenAI's o1 model~\cite{OpenAIO1_24}, modern LRMs solve complex reasoning problems via a \textit{thinking-then-answering} paradigm~\cite{DeepSeekR125,QwenQwQ25,O1ReplicationJourneyII24}.
Generally, a special end-of-thinking token ``\texttt{</think>}'' is generated as a \textit{end-of-thinking moment}, transitioning from the thinking phase to the answering phase.
During the thinking process, LRMs automatically transit between slow thinking and fast thinking, utilizing self-reflection as chain of thoughts~\cite{CoT22}.

\paragraph{Slow Thinking Transitioning}
To leverage human-like system-2 slow thinking that helps solve complex reasoning problems, o1-style LRMs automatically transit between fast thinking and slow thinking.
To be specific, during the thinking process, LRMs frequently generate slow thinking transitioning tokens such as ``\texttt{wait}'', ``\texttt{hmm}'', and ``\texttt{alternatively}'', \etc.
Once these tokens are generated, LRMs slow down reasoning, where previous reasoning chains are self-reflected and corrected immediately.
Hence, reasoning following the transitioning token can be viewed as slow thinking, while the rest is generally fast thinking.

\paragraph{Reasoning Progress}
Let the overall answer sequence generation process be a \textit{reasoning progress} $\mathcal{P}\in[0, 1]$, where $0$ and $1$ indicate the start and the end of reasoning, respectively.
Notably, reasoning progress represents the overall problem-solving progress instead of the number of generated tokens, where a reasoning progress closer to $1$ represents the reasoning chain is more informative.
For example, the reasoning progress can be closer to $1$ while generating fewer tokens, indicating more efficient reasoning.
However, it is intractable to measure the exact progress obtained.
Hence, we define the reasoning progress following a \textit{reasoning velocity} assumption.
Given the total time $t=T>0$ spent on generating the whole sequence, the \textit{reasoning velocity} at timestep $t$, $\mathcal{V}_t$ is defined as $\frac{d\mathcal{P}}{dt}$, where $dt$ is the infinitesimal of time.
We assume:
\begin{assumption}
\textit{The reasoning velocity of slow thinking is smaller than that of fast thinking.}
\end{assumption}
\noindent See \cref{fig:teaser}, different reasoning strategies result in different reasoning progress achieved over time.
\subsection{Reasoning Progress Modulation: A Universal View of Test-Time Scaling}\label{sec:problem_statement}
There are mainly two components that must be modulated:
\textbf{i) Thinking phase budget.}
As discussed before, o1-like LRMs follow a ``think-then-answer'' paradigm.
Therefore, modulating reasoning via scaling up or down the thinking phase budget is required.
\textbf{ii) Slow thinking scheduling.}
Within the thinking phase, the transition to slow thinking should also be modulated, thus increasing or reducing slow thinking according to a certain plan specified by users (\eg, slow thinking first, and then fast thinking).
With user-defined scheduling, the modulation of slow thinking transitions vary arbitrarily, ranging from sparse modulation—where little is adjusted—to dense modulation, where adjustments are frequent and extensive.

Based on the above analysis, we establish a unified perspective on test-time scaling and identify key limitations in existing approaches—namely, their failure to consider both reasoning schedule and overall thinking budget jointly.
For instance, s1 modulates reasoning by sparsely increasing slow thinking (\ie, adding two \texttt{wait} tokens), but overlooks broader thinking budget adjustments~\cite{S1_25}.
Conversely, Chain-of-Draft (CoD) reduces the thinking budget while neglecting the scheduling of slow thinking~\cite{ChainOfDraft25}.
As a result, while LRMs are indirectly guided to reason more or less—sometimes achieving deeper reasoning or pruning unproductive thoughts—\textbf{we instead aim to explicitly and universally modulate the reasoning process by jointly considering both components}, as introduced next.

\section{\ours}
We introduce \oursbf (\oursshort), a universal reasoning progress modulation framework for test-time scaling of LRMs, which is illustrated in \cref{fig:framework}.
In the following, we first introduce $\alpha$ moment in \cref{sec:alpha_moment}, a moment that the thinking phase budget is scaled at least $\alpha\times$.
In \cref{sec:thinking_modulation} and \cref{sec:answer_modulation}, we detail how we modulate slow thinking scheduling pre-$\alpha$ moment and modulating fast thinking encouragement post-$\alpha$ moment, respectively.

\subsection{$\bs\alpha$ Moment for Universal Modulation}\label{sec:alpha_moment}
To modulate the thinking phase budget, we scale the thinking phase by at least $\alpha\times$, where $\alpha>1$ is a universal modulating parameter.
Formally, given the average thinking phase token length $\overline{N}_{\text{think}}>0$ generated by an LRM, we scale the thinking phase token length to $\alpha \overline{N}_{\text{think}}$, where the moment when the generated token length reaches $\alpha \overline{N}_{\text{think}}$ is dubbed as ``\textit{$\alpha$ moment}''.
In addition to scaling the thinking phase, we modulate the thinking phase via slow thinking scheduling before the $\alpha$ moment, thus achieving both controllable and scalable thinking.
Note that $\alpha$ moment does not represent the new thinking phase transitioning moment, because the thinking phase typically continues after $\alpha$ moment, which we will elaborate later.
\begin{table*}[ht!]
\caption{
\textbf{Systematic comparison of reasoning results} on mathematical, coding, and science reasoning  benchmarks with
DeepSeek-R1-Distill-Qwen-1.5B, DeepSeek-R1-Distill-Qwen-7B, and Qwen QwQ 32B. Additional results for other models are provided in \cref{app:addition_results}.
P@1: Pass@1 (\%); \#Tk: number of generated tokens;
$\overline{\Delta}_{\text{P@1}}$ (\%): average Pass@1 result boost over the base model. 
$^*$For a fair comparison, \sone~\cite{S1_25} directly applies budget forcing at test-time without supervised fine-tuning, which is same as CoD and our \oursshort that are training-free.
}
\label{table:main_results}
\centering
\renewcommand{\arraystretch}{1.3} 
\setlength{\tabcolsep}{1.5mm}
\resizebox{\textwidth}{!}{
\begin{tabular}{llllllllllllcc}
\toprule[0.95pt]
\multirow{3}{*}[-1.0ex]{\textbf{Method}} 
&
\multicolumn{8}{c}{\scshape\textbf{Mathematical}}
& \multicolumn{2}{c}{\scshape\textbf{Coding}}
& \multicolumn{2}{c}{\scshape\textbf{Science}}
\\
\cmidrule(lr){2-9}
\cmidrule(lr){10-11}
\cmidrule(lr){12-13}

&
\multicolumn{2}{c}{\textbf{AIME24}}
& \multicolumn{2}{c}{\textbf{AMC23}} 
& \multicolumn{2}{c}{\textbf{Minerva}}
& \multicolumn{2}{c}{\textbf{MATH500}}
& \multicolumn{2}{c}{\textbf{LiveCode}}
& \multicolumn{2}{c}{\textbf{Olympiad}}
\\
\cmidrule(lr){2-3}
\cmidrule(lr){4-5}
\cmidrule(lr){6-7}
\cmidrule(lr){8-9}
\cmidrule(lr){10-11}
\cmidrule(lr){12-13}
& 
P@1 & \#Tk &
P@1 & \#Tk &
P@1 & \#Tk &
P@1 & \#Tk &
P@1 & \#Tk &
P@1 & \#Tk &
$\overline{\Delta}_{\text{P@1}}$
\\
\midrule[0.6pt]
\multicolumn{14}{c}{\textit{\textbf{DeepSeek-R1-Distill-Qwen-1.5B}}}\\
\midrule[0.6pt]
\baselineexp  & 23.3 & 7280  
& 57.5 & \cellcolor{lightpink}{5339} 
& \cellcolor{lightpink}{32.0} & \cellcolor{lightpink}{4935} 
& 79.2 & \cellcolor{lightpink}{3773} 
& 17.8 & 6990 
& 38.8 & 5999
& N/A
\\

s1$^*$  & \cellcolor{lightpink}{26.7$_{\text{+3.4}}$} & 7798 & 57.5$_{\text{+0.0}}$ & 6418 & 31.6$_{\text{-0.4}}$ & 5826 & 78.2$_{\text{-1.0}}$ & 4733 & 17.0$_{\text{-0.8}}$ & 7025 & 38.5$_{\text{-0.3}}$ & 6673 & +0.15\\

\codexp  & \cellcolor{deeppink}\textbf{30.0}$_{\text{+6.7}}$ & \cellcolor{lightpink}{6994} & \cellcolor{lightpink}{65.0}$_{\text{+7.5}}$ & 5415 & 29.0$_{\text{-3.0}}$ & \cellcolor{deeppink}\textbf{4005} & \cellcolor{deeppink}\textbf{81.4}$_{\text{+2.2}}$ & \cellcolor{deeppink}\textbf{3136} & \cellcolor{lightpink}{20.3}$_{\text{+2.5}}$ & \cellcolor{lightpink}{6657} & \cellcolor{lightpink}{40.6}$_{\text{+1.8}}$ & \cellcolor{lightpink}{5651} & +2.95\\

\oursshortexp  & \cellcolor{deeppink}\textbf{30.0}$_{\text{\textcolor{runpei-orange}{\cellcolor{deeppink}\textbf{+6.7}}}}$ & \cellcolor{deeppink}\textbf{5916} & \cellcolor{deeppink}\textbf{70.0}$_{\text{\textcolor{runpei-orange}{\cellcolor{deeppink}\textbf{+12.5}}}}$ & \cellcolor{deeppink}\textbf{4952} & \cellcolor{deeppink}\textbf{34.2}$_{\text{\textcolor{runpei-orange}{\cellcolor{deeppink}\textbf{+2.2}}}}$ & 4586 & \cellcolor{lightpink}{81.0}$_{\text{\textcolor{runpei-orange}{\textbf{+1.8}}}}$ & 3852 & \cellcolor{deeppink}\textbf{24.8}$_{\text{\textcolor{runpei-orange}{\cellcolor{deeppink}\textbf{+7.0}}}}$ & \cellcolor{deeppink}\textbf{5426} & \cellcolor{deeppink}\textbf{45.5}$_{\text{\textcolor{runpei-orange}{\cellcolor{deeppink}\textbf{+6.7}}}}$ & \cellcolor{deeppink}\textbf{4944} & \textcolor{runpei-orange}{\textbf{+6.15}} \\

\midrule[0.6pt]
\multicolumn{14}{c}{\textit{\textbf{DeepSeek-R1-Distill-Qwen-7B}}}\\
\midrule[0.6pt]
\baselineexp  & \cellcolor{lightpink}{46.7} & \cellcolor{lightpink}{6648} 
& 82.5 & 4624 
& 40.4 & 4191 
& 87.6 & \cellcolor{lightpink}{3239} 
& 43.5 & 5885 
& 50.4 & 5385
& N/A
\\

s1$^*$  & \cellcolor{lightpink}{46.7$_{\text{+0.0}}$} & 7295 & 80.0$_{\text{-2.5}}$ & 5673 & \cellcolor{lightpink}{42.3$_{\text{+1.9}}$} & 6510 & \cellcolor{deeppink}\textbf{92.8}$_{\text{+5.2}}$ & 5848 & 44.0$_{\text{+0.5}}$ & 5979 & \cellcolor{lightpink}{54.2$_{\text{+3.8}}$} & 6007 & +1.48\\

\codexp  & 43.3$_{\text{-3.4}}$ & \cellcolor{deeppink}\textbf{6078} & \cellcolor{lightpink}{87.5}$_{\text{+5.0}}$ & \cellcolor{deeppink}\textbf{3594} & \cellcolor{deeppink}\textbf{43.4}$_{\text{+3.0}}$ & \cellcolor{deeppink}\textbf{2142} & 88.8$_{\text{+1.2}}$ & \cellcolor{deeppink}\textbf{2094} & \cellcolor{lightpink}{45.0$_{\text{+1.5}}$} & \cellcolor{lightpink}{5593} & 53.5$_{\text{+3.1}}$ & \cellcolor{deeppink}\textbf{4520} & +1.73\\

\oursshortexp  & \cellcolor{deeppink}\textbf{50.0}$_{\text{\textcolor{runpei-orange}{\cellcolor{deeppink}\textbf{+3.3}}}}$ & 6827 & \cellcolor{deeppink}\textbf{90.0}$_{\text{\textcolor{runpei-orange}{\cellcolor{deeppink}\textbf{+7.5}}}}$ & \cellcolor{lightpink}{4397} & \cellcolor{lightpink}{42.3}$_{\text{\textcolor{runpei-orange}{\textbf{+1.9}}}}$ & \cellcolor{lightpink}{4124} & \cellcolor{lightpink}{91.2}$_{\text{\textcolor{runpei-orange}{\textbf{+3.6}}}}$ & 4337 & \cellcolor{deeppink}\textbf{49.8}$_{\text{\textcolor{runpei-orange}{\cellcolor{deeppink}\textbf{+6.3}}}}$ & \cellcolor{deeppink}{\textbf{5067}} & \cellcolor{deeppink}\textbf{55.7}$_{\text{\textcolor{runpei-orange}{\cellcolor{deeppink}\textbf{+5.3}}}}$ & \cellcolor{lightpink}{4883} & \textcolor{runpei-orange}{\textbf{+4.65}}\\

\midrule[0.6pt]
\multicolumn{14}{c}{\textit{\textbf{Qwen QwQ-32B}}}\\
\midrule[0.6pt]
\baselineexp  & 40.0 & 4058 & 77.5 & 2901 & \cellcolor{deeppink}\textbf{47.8}\phantom{0} & 2199 & 90.2 & 1951 & {67.0} & \cellcolor{lightpink}{5092} & 53.6 & 3230 & N/A\\

s1$^*$   & 43.3$_{\text{+3.3}}$  & 4221 & 77.5$_{\text{+0.0}}$  & 3068 & 46.7$_{\text{-1.1}}$  & 2433 & \cellcolor{deeppink}\textbf{90.8}$_{\text{+0.6}}$  & 2218 & 66.5$_{\text{-0.5}}$  & 5260 & 55.1$_{\text{+1.5}}$  & 3454 & +0.63 \\

\codexp  & \cellcolor{lightpink}{46.7}$_{\text{+6.7}}$ & \cellcolor{lightpink}{3959} & \cellcolor{lightpink}{80.0}$_{\text{+2.5}}$ & \cellcolor{lightpink}{2400} & \cellcolor{lightpink}{47.4$_{\text{-0.4}}$} & \cellcolor{lightpink}{1464} & \cellcolor{lightpink}{90.6}$_{\text{+0.4}}$ & \cellcolor{deeppink}\textbf{1421} & \cellcolor{lightpink}{66.8$_{\text{-0.2}}$} & \cellcolor{deeppink}\textbf{4984} & \cellcolor{deeppink}{\textbf{57.2}}$_{\text{+3.6}}$ & \cellcolor{lightpink}{2844} & +2.10 \\

\oursshortexp & \cellcolor{deeppink}\textbf{53.3}$_{\text{\textcolor{runpei-orange}{\cellcolor{deeppink}\textbf{+13.3}}}}$ & \cellcolor{deeppink}\textbf{3141} & \cellcolor{deeppink}\textbf{87.5}$_{\text{\textcolor{runpei-orange}{\cellcolor{deeppink}\textbf{+10.0}}}}$ & \cellcolor{deeppink}\textbf{2286} & 46.0$_{\text{\textcolor{gray}{-1.8}}}$ & \cellcolor{deeppink}\textbf{1441} & 89.4$_{\text{\textcolor{gray}{-0.8}}}$ & \cellcolor{lightpink}{1668} & \cellcolor{deeppink}\textbf{75.8}$_{\text{\textcolor{runpei-orange}{\cellcolor{deeppink}\textbf{+8.8}}}}$ & 5824 & \cellcolor{lightpink}{56.1$_{\text{\textcolor{runpei-orange}{ \textbf{+2.5}}}}$} & \cellcolor{deeppink}\textbf{2504} & \textcolor{runpei-orange}{\textbf{+5.33}}\\
\bottomrule[0.95pt]
\end{tabular}
}
\end{table*}

\subsection{Pre-$\bs\alpha$ Moment Modulation}\label{sec:thinking_modulation}
Following previous works~\cite{SpeculativeThinking25,S1_25}, we activate slow thinking before $\alpha$ moment via appending ``\texttt{wait}'' after a frequently co-generated structural delimiters ``\texttt{\textbackslash n\textbackslash n}''.
Moreover, the activation of slow thinking is conducted following a user-specified scheduling plan, such as slow thinking, then fast thinking.

\noindent\textbf{Stochastic Reasoning Transitioning}~~
\oursshort achieves such scheduling by modeling the activation of slow thinking as a Bernoulli stochastic process.
Specifically, \oursshort appends ``\texttt{wait}'' following $\mathrm{Bernoulli}(\pwait)$. 
Let $t=0,1,\dots,T_m$ be the timestamps of generated tokens before $\alpha$ moment, where $T_m=\alpha \overline{N}_\text{think}$ represents the timestamp of $\alpha$ moment.
$\pwait$ is determined by a user-specified scheduling function $\mathcal{S}(t)$,
\begin{equation}\label{eq:St}
    \pwait := \mathcal{S}(t), t = 0, 1, \dots, T_m.
\end{equation}
$\mathcal{S}(t)$ can be an arbitrary function, such as linear annealing.
\oursshort adopts linear annealing, which we find the most effective and efficient (See \cref{sec:scheduling_strategy_analysis}).

\subsection{Post-$\bs\alpha$ Moment Modulation}\label{sec:answer_modulation}
While an LRM significantly increases slow thinking through pre-$\alpha$ modulation, this extended thinking phase often exhibits \textit{slow thinking inertia}, making it difficult to transition back to fast thinking.
Notably, without post-$\alpha$ moment modulation, the LRM substantially reduces the likelihood of generating ``\texttt{</think>}''. 
Furthermore, inserting a few ``\texttt{</think>}'' tokens does not effectively overcome the inertia, failing to fully restore fast thinking.

\noindent\textbf{Deterministic Reasoning Termination}
After the $\alpha$ moment, we guide \oursshort to transition into fast reasoning by disabling further slow thinking.
Specifically, any generated slow reasoning transition token ``\texttt{wait}'' is replaced with ``\texttt{</think>}'' to explicitly mark the end of the thinking phase, reinforcing a shift to fast thinking before entering the answering phase.
This deterministic termination strategy allows \oursshort to conclude reasoning naturally and consistently, enabling more efficient test-time scaling.

\section{Experiments}
\subsection{Experimental Setup}
\noindent\textbf{Benchmarks}~
To comprehensively evaluate the reasoning capability of LRMs, we conduct systematic evaluations on six benchmarks covering three reasoning categories:
i) mathematical reasoning, including AIME 2024 (AIME24)~\cite{AIME24}, , AMC23~\cite{AMC23}, and Minerva-Math (Minerva)~\cite{MinervaMath22};
ii) code generation, including LiveCodeBench (LiveCode)~\cite{LiveCodeBench25};
iii) scientific problems, including OlympiadBench (Olympiad)~\cite{OlympiadBench24}.
We report the problem-solving accuracy by average Pass@1 (\%), and the average number of generated tokens.

\vspace{2pt}
\noindent\textbf{Base Models}~
We use three o1-like open-source LRMs as the base model, including DeepSeek R1 distilled DeepSeek-R1-Distill-Qwen-1.5B and DeepSeek-R1-Distill-Qwen-7B~\cite{DeepSeekR125}, as well as a recently larger LRM Qwen QwQ 32B~\cite{QwenQwQ25}.

\vspace{2pt}
\noindent\textbf{Implementations}~
Without additional specifications, we use a temperature of 0.6, top-$p$ of 0.95, and the maximum token length is set to 8192.
We set $\alpha$ as 1.4, and we obtain the average thinking phase token length generated by an LRM on any benchmark by randomly sampling 10 test questions and averaging the generated token length before benchmarking.
See more details in \cref{app:more_impl}.

\vspace{2pt}
\noindent\textbf{Baselines}~
We compare our \oursshort against the vanilla LRM and two training-free, test-time scaling baselines.
i) {\scshape Base}: The original LRM that transitions between slow and fast thinking \textit{\textbf{automatically}}, without any external modulation.
ii) {\scshape s1}~\cite{S1_25}: A baseline that enforces a \textit{\textbf{monotonically increasing}} slow thinking pattern by appending approximately two ``\texttt{wait}'' tokens near the end of the reasoning phase to prolong slow thinking. For a fair comparison, we apply {\scshape s1} at test time without supervised fine-tuning used in its original implementation.
iii) {\scshape Chain of Draft (CoD)}~\cite{ChainOfDraft25}: A baseline that enforces a \textit{\textbf{monotonically decreasing}} slow thinking pattern by prompting the model to constrain each slow thinking step to no more than five words, thereby sharply reducing the thinking budget.

\subsection{Main Results}\label{sec:main_results}
\cref{table:main_results} shows the systematic comparison results of our \oursshort and baseline methods, and we observe:
i) \oursshort consistently yields a higher problem-solving accuracy than all baseline methods across all models and benchmarks.
Notably, compared to the base model, \oursshort improves the 1.5B LRM by a clear margin of \textbf{+6.15\%}, while reducing nearly \textbf{14\%} token length.
This demonstrates both the effectiveness and efficiency of \oursshort.
ii) Compared to baseline test-time scaling methods, including s1 and CoD, \oursshort still achieves significantly better results.
Specifically, the average accuracy boost over all benchmarks and models of \oursshort is \textbf{+3.12\%} and \textbf{+4.62\%} higher than CoD and s1, respectively.
iii) Surprisingly, we observe that while \oursshort modulates reasoning densely without restrictions on reducing the thinking budget (instead, we use $\alpha>1$ that increases the thinking budget), the average thinking phase token length generated by \oursshort is only about \textbf{+4.4\%} higher than the monotonically decreasing baseline CoD (4231 \textit{vs.} 4053), which is about \textbf{+21.0\%} more efficient than the monotonically increasing baseline s1 (4231 \textit{vs.} 5357).
This indicates that \oursshort achieves more efficient reasoning than baselines, which we provide analysis later.
\begin{figure*}[ht!]
  \includegraphics[width=1.0\linewidth]{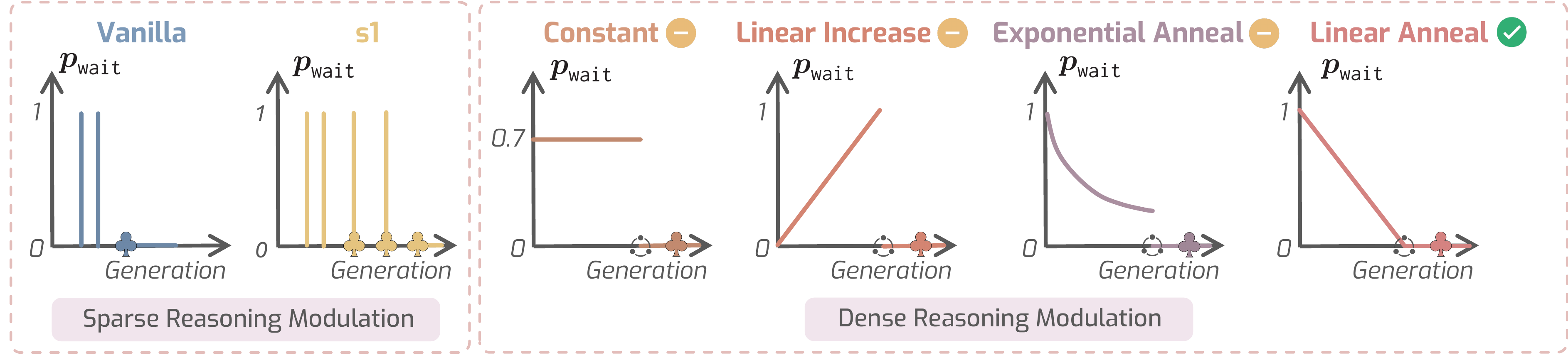}
  \caption {\textbf{Visualization of different scheduling strategies.} 
  We detail the functions in \cref{sec:scheduling_strategy_analysis}.
  Here \raisebox{-0.25em}{\includegraphics[height=1.2em]{figures/src/point.pdf}} represents $\alpha$ moment, which we elaborate in \cref{sec:alpha_moment}, and \raisebox{-0.18em}{\includegraphics[height=0.92em]{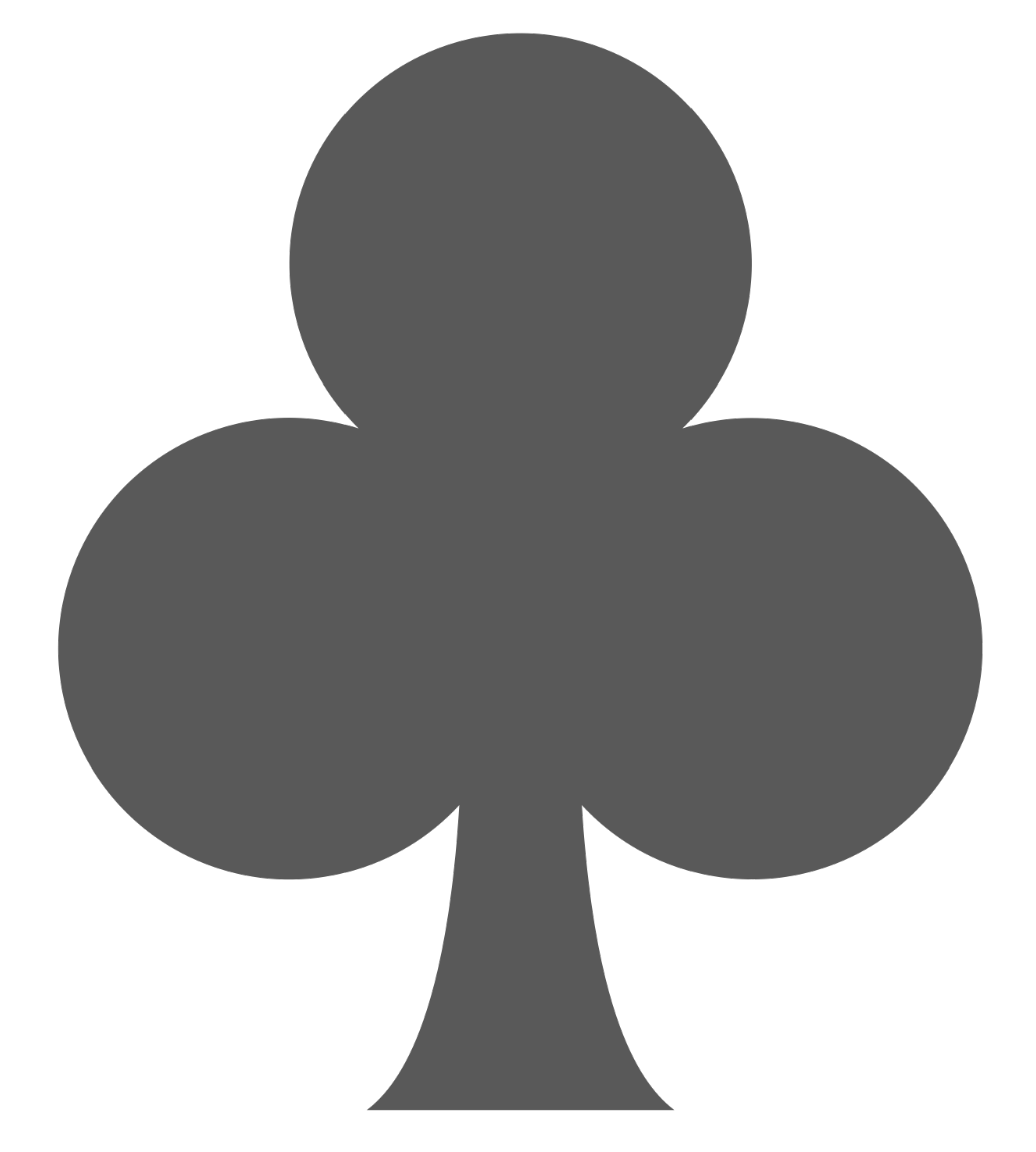}} denotes the end of the thinking phase.
  }\label{fig:scheduling_plot}
\end{figure*}
\begin{figure*}[t!]
  \includegraphics[width=1.0\linewidth]{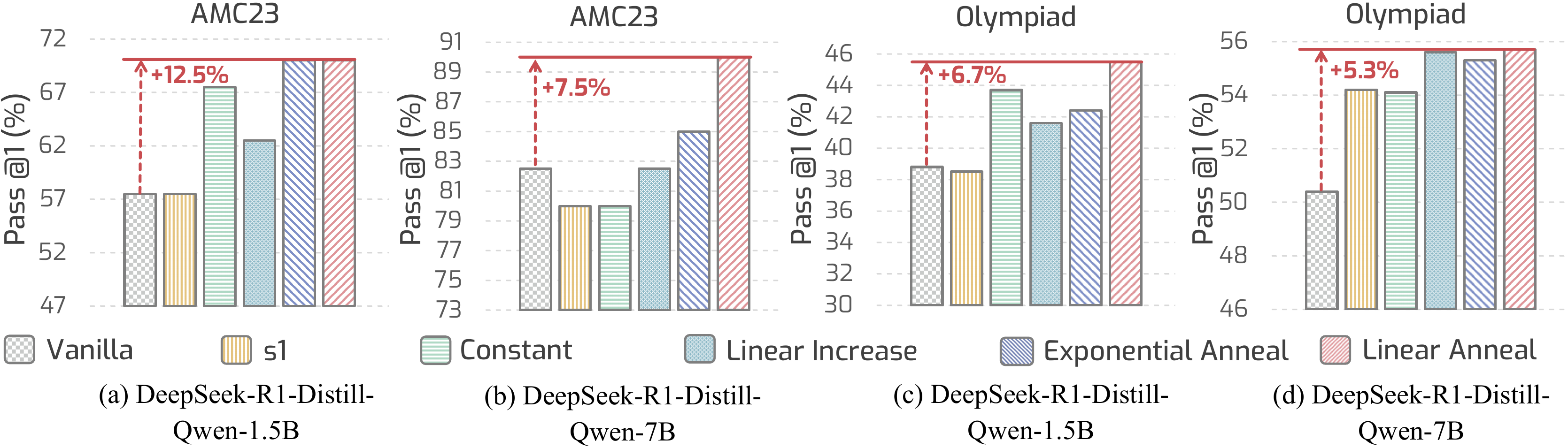}
  \caption{\textbf{Ablation study of different scheduling strategies} on (a-b) AMC23 and (c-d) OlympaidBench. 
  }
  \label{fig:scheduling_ablation}
\end{figure*}

\subsection{Analytic Results}\label{sec:analysis}
In this section, we analyze \oursshort by systematically addressing the following five questions:

\subsubsection{What scheduling strategy is better?}\label{sec:scheduling_strategy_analysis}
As shown in \cref{fig:scheduling_plot}, we study four variants of scheduling strategies for $\mathcal{S}(t)$ defined in \cref{eq:St}, where $T_m=\alpha \overline{N}_\text{think}$ represents the timestamp of $\alpha$ moment:
\begin{itemize}[leftmargin=*]
    \item \textbf{Constant:}
    \(\mathcal{S}(t):=\bs{p}_{\text{constant}}\), where $\bs{p}_{\text{constant}}\in [0,1]$ is a constant probability.
    This represents a consistently more slow thinking strategy, and the increase is large when $\bs{p}_{\text{constant}}$ is larger.
    Note that when $\bs{p}_{\text{constant}}=0$ and $\alpha=1$, it degenerates to vanilla reasoning models; and when $\bs{p}_{\text{constant}}<0.1$ and $\alpha>1$, it degenerates to s1-like model, where only about two ``\texttt{wait}'' are appended.
    
    \item \textbf{Linear increase:}
    \(\mathcal{S}(t) := \frac{1}{T_m} t\), where $t=\{0,1,\dots,T_m\}$ and $\frac{1}{T_m}>0$ indicates the increasing coefficient.
    This scheduling function indicates a fast-to-slow thinking strategy.
    
    \item \textbf{Exponential anneal:} 
    \(\mathcal{S}(t) := \exp(-\gamma{t})\), where $t=\{0,1,\dots,T_m\}$ and $\gamma>0$ is a hyper-parameter that controls annealing speed (here we use $\gamma=0.3$).
    This scheduling function indicates a slow-to-fast thinking strategy.
    
    \item \textbf{Linear anneal:}
    \(\mathcal{S}(t) := -\frac{1}{T_m} t + 1\), where $-\frac{1}{T_m} < 0$ indicates the annealing coefficient.
    Its modulation is similar to exponential anneal scheduling.
\end{itemize}
\begin{figure*}[ht!]
  \includegraphics[width=1.0\linewidth]{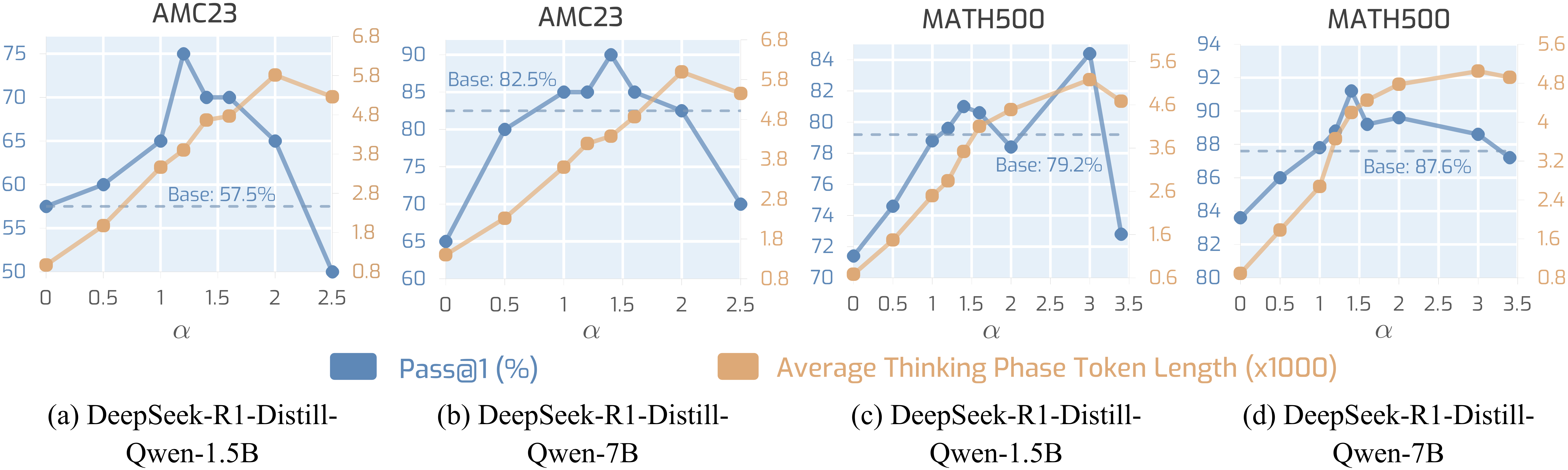}
  \caption{\textbf{Scaling property of $\bs\alpha$.}
  We scale $\alpha$ from $0$ to the maximum value restricted by the maximum token length, and plot the corresponding reasoning Pass@1 and average thinking phase token length on AMC23 and MATH500.
  }\label{fig:alpha-scaling}
\end{figure*}
\begin{figure}[t!]
  \includegraphics[width=1.0\columnwidth]
  {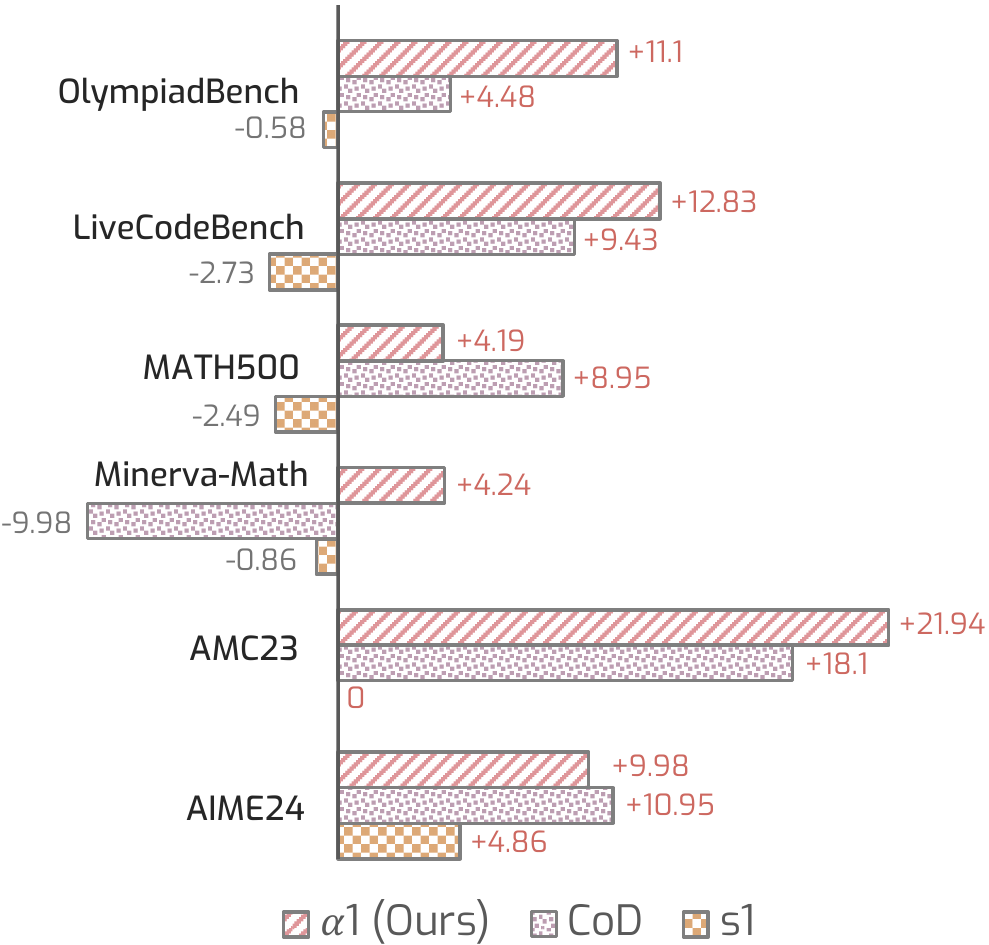}
  \caption{\textbf{Scaling efficiency analysis with REP} using Deepseek-R1-distill-Qwen-1.5B. 
  The REP metric is introduced in \cref{eq:REP}.
  }
  \label{fig:think-metric}
\end{figure}

\cref{fig:scheduling_ablation} shows the results of \oursshort using these four different scheduling strategies.
We observe:
i) Linear anneal consistently yields the highest reasoning accuracy, indicating that the \textit{slow thinking first, then fast thinking} is a better slow thinking scheduling strategy.
ii) Similar to linear anneal, exponential anneal also follows an annealing slow thinking scheduling, where the improvement on the 1.5B model further demonstrates the efficacy of the slow thinking, then fast thinking strategy.
However, such annealing scheduling may lead to an unstable performance boost compared to linear anneal.

\subsubsection{Can $\alpha$-moment scale the thinking phase budget?}\label{sec:alpha_scaling_law}
\cref{fig:alpha-scaling} shows the results of \oursshort with different $\alpha$-moments determined by scaling $\alpha$ from $0$ to a maximum value subject to the 8192 token length budget.
We observe:
i) $\alpha$-moment enables a \textit{scalable thinking phase budgeting}.
By scaling up $\alpha$, the average thinking phase token length is accordingly scaled up.
ii) Interestingly, while the thinking phase is scaled up, there exists a trade-off between the optimal value of $\alpha$ and the resulting reasoning accuracy.
This indicates that monotonously increasing the thinking phase budget does not consistently bring better reasoning performance, and it is critical to find the optimal $\alpha$-moment that results in a satisfactory improvement.

\subsubsection{Does \oursshort scale more efficiently?}\label{sec:efficiency_analysis}
To quantitatively evaluate how different methods trade off reasoning efficiency and accuracy, we introduce the $\mathcal{F}_{\text{REP}}(\mathcal{A}_{\text{method}};\mathcal{A}_{\text{base}},T_{\text{norm}})$ (Reasoning Efficiency-Performance, REP) metric. 
The $\text{REP}$ metric is defined as:
\begin{equation}\label{eq:REP}
\mathcal{F}_{\text{REP}}(\mathcal{A}_{\text{method}};\mathcal{A}_{\text{base}},T_{\text{norm}}) = \frac{\mathcal{A}_{\text{method}} - \mathcal{A}_{\text{base}}}{T_{\text{norm}}}
\end{equation}
where \(\mathcal{A}_{\text{method}}\) and \(\mathcal{A}_{\text{base}}\) denote the reasoning accuracy of the evaluated method and the base model, respectively. \(T_{\text{norm}}\) is the normalized thinking phase token length, computed by dividing the current thinking phase token length by the maximum token length. 
Higher REP indicates stronger performance with better reasoning efficiency.

We report the REP of CoD, s1, and \oursshort on six reasoning benchmarks with Deepseek-R1-distill-Qwen-1.5B. 
\cref{fig:think-metric} shows that \oursshort achieves higher REP on most benchmarks, indicating a more favorable balance between reasoning performance and efficiency. 
Notably, \oursshort outperforms CoD by \textbf{+6.62} and s1 by \textbf{+11.68} on Olympiad-Bench, and exceeds CoD by \textbf{+14.22} on Minerva-Math.

\begin{figure*}[ht!]
\includegraphics[width=1.0\linewidth]{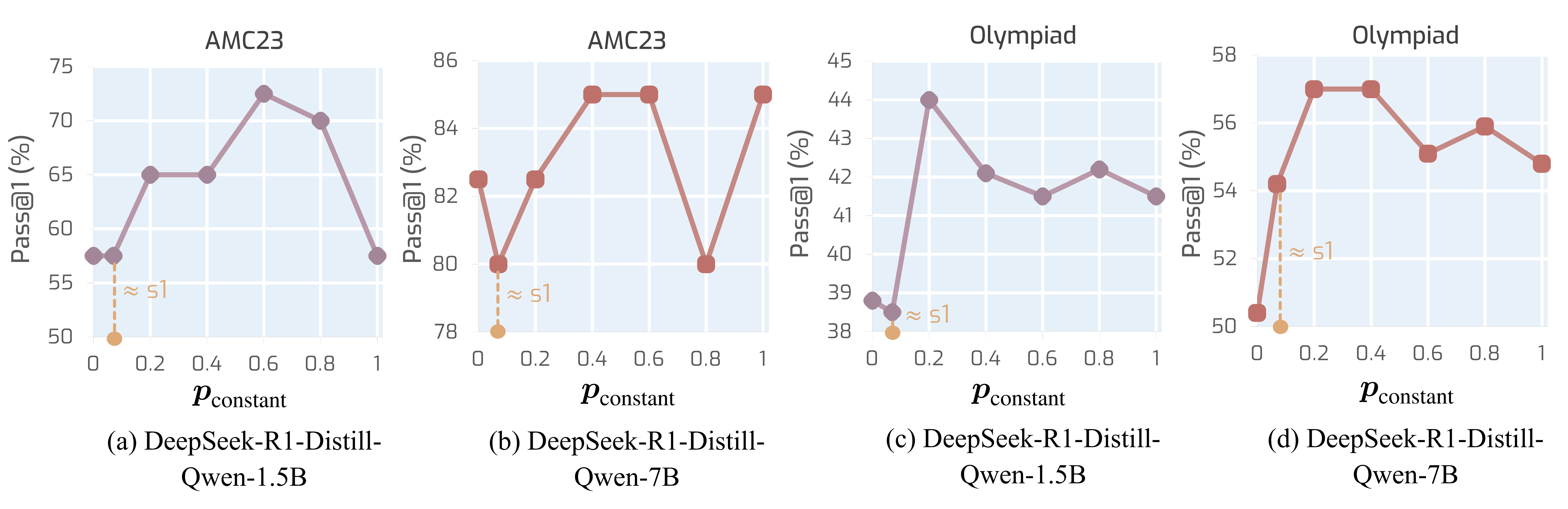}
  \caption{\textbf{Scaling property of ``\texttt{wait}'' frequency} under constant scheduling on AMC23 and OlympiadBench.
  Increasing $p_{\text{constant}}$ leads to a higher frequency of yielding ``\texttt{wait}'' in the Bernoulli process $\mathrm{Bernoulli}(\pwait)$.
  }
  \label{fig:wait_frequency}
\end{figure*}
\subsubsection{How frequent should slow thinking transitioning be?}
\oursshort modulate slow thinking transitioning via sampling from $\mathrm{Bernoulli}(\pwait)$, which leads to another question of how large should $\pwait$ be that can bring a better result.
To study this question, we use the constant scheduling function and scale $p_{\text{constant}}$ from $0$ to $1$ to increase the frequency of transitioning to slow thinking.
This is because the constant scheduling is a sampling process with a certain probability, and the value of the probability determines how frequently the slow thinking transitioning token will be sampled.
\cref{fig:wait_frequency} shows the results, from which we observe:
i) An extremely low or high frequency of transitioning to slow thinking brings unsatisfactory results (\eg, $p_{\text{constant}}=0.1$).
Similar to the scaling of the thinking phase dedget (\eg, modualting $\alpha$), the slow thinking frequency also needs to be carefully selected.
ii) While an extremely dense or sparse slow thinking transitioning leads to unsatisfactory results, the reasoning performance is decent across a large range of $p_{\text{constant}}$, demonstrating that increasing slow thinking generally brings improved reasoning.

\begin{table}[t!]
\centering
\caption{\textbf{Ablation study on post-$\bs\alpha$ moment modulation}. 
Without post-$\alpha$ modulation represents our $\alpha$1 without the suppression of the slow thinking inertia after the $\alpha$ moment.
}
\label{tab:suppress_wait_think_tokens}
\vspace{-10pt}
\small
\setlength{\tabcolsep}{4pt}
\renewcommand{\arraystretch}{1.3}
\resizebox{\columnwidth}{!}{
\begin{tabular}{lccccc}
\toprule
\multirow{2}{*}[-0.5ex]{\textbf{Method}} & \multirow{2}{*}[-0.5ex]{\textbf{\shortstack{Post-$\alpha$ Moment\\Modulation}}} & \multicolumn{2}{c}{\textbf{AIME24}} & \multicolumn{2}{c}{\textbf{AMC23}}\\
\cmidrule(lr){3-4} \cmidrule(lr){5-6}
& & P@1 & \#Tk & P@1 & \#Tk\\
\midrule[0.6pt]
\multicolumn{6}{c}{\textit{\textbf{DeepSeek-R1-Distill-Qwen-1.5B}}}\\
\midrule[0.6pt]
\baselineexp & N/A & 23.3 & \cellcolor{lightpink}7280 & \cellcolor{lightpink}57.5 & \cellcolor{lightpink}5339\\
\oursshortexp & $\times$ & \cellcolor{lightpink}26.7 & 7929 & 47.5 & 6903\\
\oursshortexp & $\checkmark$ & \cellcolor{deeppink}\textbf{30.0} & \cellcolor{deeppink}\textbf{5916} & \cellcolor{deeppink}\textbf{70.0} & \cellcolor{deeppink}\textbf{4951}\\
\midrule[0.6pt]
\multicolumn{6}{c}{\textit{\textbf{DeepSeek-R1-Distill-Qwen-7B}}}\\
\midrule[0.6pt]
\baselineexp & N/A & \cellcolor{lightpink}38.8 & \cellcolor{deeppink}\textbf{5999} & \cellcolor{lightpink}82.5 & \cellcolor{lightpink}4624\\
\oursshortexp & $\times$ & 30.0 & 7666 & 75.0 & 5878\\
\oursshortexp & $\checkmark$ & \cellcolor{deeppink}\textbf{50.0} & \cellcolor{lightpink}6826 & \cellcolor{deeppink}\textbf{90.0} & \cellcolor{deeppink}\textbf{4397}\\
\bottomrule
\end{tabular}
}
\end{table}
\subsubsection{Is post-$\alpha$ moment modulation necessary?}\label{sec:post_alpha}
Typical test-time scaling methods focus on the modulation of slow thinking within the thinking phase, while \oursshort consists of a post-$\alpha$ moment modulation that encourages fast thinking.
To validate its necessity of enforcing fast thinking in the end, we conduct an ablation study on utilizing the post-$\alpha$ moment modulation, shown in \cref{tab:suppress_wait_think_tokens}.
We observe:
i) Pre-$\alpha$ moment modulation of slow thinking is insufficient. When the post-$\alpha$ moment modulation is reduced to a single operation, the performance of \oursshort significantly drops.
This is because the increase of slow thinking during pre-$\alpha$ moment brings a \textit{slow thinking inertia} (as discussed before in \cref{sec:answer_modulation}), leading to a slow thinking intensive reasoning.
ii) By utilizing a post-$\alpha$ moment modulation, \oursshort successfully ends in a fast thinking, which demonstrates the necessity of \textit{combining both slow thinking and fast thinking}.

\section{Related Works}
\subsection{Large Reasoning Models}
Large Reasoning Models are rapidly emerging as a family of foundation models~\cite{FoundationModel21} that target human-level system-2 reasoning~\cite{ThinkingFastAndSlow}.
Starting from OpenAI's o1~\cite{OpenAIO1_24} in 2024, numerous efforts follow this ``thinking-then-answering'' paradigm. 
Notably, o1-like Large Language Models (LLMs) can solve increasingly complex reasoning problems after a thorough chain of thoughts~\cite {CoT22,TreeOfThoughts23,GraphOfThoughts24}, such as the IMO competition.
These advanced models are mainly developed via large-scale reinforcement learning (RL) to align human preference~\cite{RLHF17,PPO17,DeepSeekMath24,DeepSeekR125}, where a reward model is used to judge model answers~\cite{PRMandORM22,MATH500_24}.
Notable efforts replicating o1's success include DeepSeek R1, Qwen QwQ, and Phi-4~\cite{Phi4_24,DeepSeekR125,QwenQwQ25}, which typically utilize a special end-of-thinking token ``\texttt{</think>}'', after which a solution is output to the user.
Recently, some researchers have explored applying RL during post-training fine-tuning, where promising results have been obtained~\cite{InferenceAwareFinetuningBoN25,MetaRL25,TTRL25}.

\subsection{Reasoning with Test-Time Scaling}
Reasoning with test-time scaling has recently become a useful strategy that empowers LLMs with a scalable reasoning capability at test time.
The mainstream methods lie in two categories, \ie, i) parallel scaling and ii) sequential scaling.
The key idea of parallel scaling is Best-of-N (BoN) sampling, where the best choice is selected using uncertainty criteria like self-consistency~\cite{SelfConsistency23}, reward model~\cite{MATH500_24,TrainingVerifiers21}, or perplexity~\cite{LongContextPerplexity25}.
Specifically, one line of work focuses on sequence-level sampling~\cite{TrainingVerifiers21,BoNBoN24,DynamicSelfConsistency24,FastBestOfN24,InferenceAwareFinetuningBoN25,Bond25,VBoN25,InternalProbSelfConsistency25,RevisitTestTime25,SelfCertainty25}, while another line of work utilizes token-/step- level sampling including beam-/tree- based searching~\cite{StochasticBeams19,SelfEvalBeamSearch23,PlanningForCodeGeneration23,PlanningWithWorldModel23,TreeBoN24,ContrastiveMCTS24,OVMReasoning24,AlphaZeroTreeSearch24,AlphaMathAlmostZero24}.
Meanwhile, sequential scaling enhances or reduces slow thinking.
This technique typically relies on an iterative refinement and revision of answers generated by LLMs themselves~\cite{Star22,SelfRefine23} or external feedback~\cite{SelfDebug24,CRITIC24,LLMCannotSelfCorrect24,WhenCanLLMsSelfCorrect24,aliagent,SpeculativeDecoding25}.
Following this line of research, recent works have been devoted to addressing the underthinking and overthinking issues of modern LRMs via reinforcing~\cite{S1_25} and restricting~\cite{ChainOfDraft25,ElasticReasoning25} slow thinking, respectively.
Given the non-conflict between parallel scaling and sequential scaling, there exists another group of hybrid scaling methods that leverage both strategies~\cite{SstarCode25,RevisitTestTime25}.
\section{Conclusions}\label{sec:conclusions}
In this paper, we study the problem of test-time scaling of large reasoning models with our framework, \ours (\oursshort).
\ours starts from a universal view of reasoning modulation targeting two key aspects: thinking phase budgeting and slow thinking scheduling.
We introduce $\alpha$-moment, which is determined by $\alpha$ that scales the thinking phase budget by at least $\alpha\times$.
\ours operates by scheduling slow thinking before the $\alpha$-moment, and fast thinking after the $\alpha$-moment that eliminates slow thinking inertia.
Using \ours, we investigate the test-time scaling from various aspects, including the overall slow and fast thinking transitioning plan, thinking phase budget scaling property, and efficiency of test-time scaling, \etc.
Insightful findings are obtained, \eg, slow thinking first, then fast thinking leads to better reasoning capability of LRMs.

\section*{Limitations}
While \ours provides a universal view of test-time scaling of LRMs, and a significant performance boost has been achieved, we identify some possible limitations as follows.
i) \ours targets at o1-style LRMs, where tokens such as ``\texttt{wait}'' is proved effective in transitioning into slow thinking.
However, future LRMs may use a different slow thinking transitioning strategy, leading to a possibility of incompatibility with our framework.
ii) \ours relies on $\alpha$-moment throughout reasoning modulation, and the average thinking phase token length is typically required.
This paper obtains it by first running LRMs on 10 random samples, which requires marginal cost.
However, in case that no test questions are available, \ours can only rely on an empirical thinking phase length that may be suboptimal.

\section*{Broader Impact}
This work targets complex reasoning problems with LRMs, which we believe will lead to no ethical concerns.
However, since LRMs are modern variants of LLMs, any ethical concerns raised by LLMs can potentially exist.
\section*{Acknowledgments}
Huan Zhang is partially funded by the AI2050 program at Schmidt Sciences (AI2050 Early Career Fellowship). 
The authors thank Heng Dong for his valuable suggestions on this project.

\bibliography{main}

@String(CVPR  = {CVPR})

@String(ECCV  = {ECCV})

@String(ICLR  = {ICLR})

@String(ICML  = {ICML})

@String(AAAI  = {AAAI})

@String(EMNLP = {EMNLP})

@String(CVPR= {IEEE/CVF Conf. Comput. Vis. Pattern Recog. (CVPR)})

@String(ECCV= {Eur. Conf. Comput. Vis. (ECCV)})

@String(NeurIPS= {Adv. Neural Inform. Process. Syst. (NeurIPS)})

@String(ICLR = {Int. Conf. Learn. Represent. (ICLR)})

@String(ICML = {Int. Conf. Mach. Learn. (ICML)})

@String(AAAI = {AAAI Conf. Artif. Intell. (AAAI)})

@String(EMNLP = {Empir. Method. Nat. Lang. Process. (EMNLP)})

@article{Phi4_25,
  title={Phi-4-reasoning technical report},
  author={Abdin, Marah and Agarwal, Sahaj and Awadallah, Ahmed and Balachandran, Vidhisha and Behl, Harkirat and Chen, Lingjiao and de Rosa, Gustavo and Gunasekar, Suriya and Javaheripi, Mojan and Joshi, Neel and others},
  journal={arXiv preprint arXiv:2504.21318},
  year={2025}
}

@article{ReproducibleReasoning25,
  title={Give Me FP32 or Give Me Death? Challenges and Solutions for Reproducible Reasoning},
  author={Yuan, Jiayi and Li, Hao and Ding, Xinheng and Xie, Wenya and Li, Yu-Jhe and Zhao, Wentian and Wan, Kun and Shi, Jing and Hu, Xia and Liu, Zirui},
  journal={arXiv preprint arXiv:2506.09501},
  year={2025}
}

@article{ElasticReasoning25,
  title={Scalable chain of thoughts via elastic reasoning},
  author={Xu, Yuhui and Dong, Hanze and Wang, Lei and Sahoo, Doyen and Li, Junnan and Xiong, Caiming},
  journal={arXiv preprint arXiv:2505.05315},
  year={2025}
}

@article{SpeculativeDecoding25,
  title={Reward-guided speculative decoding for efficient llm reasoning},
  author={Liao, Baohao and Xu, Yuhui and Dong, Hanze and Li, Junnan and Monz, Christof and Savarese, Silvio and Sahoo, Doyen and Xiong, Caiming},
  journal={arXiv preprint arXiv:2501.19324},
  year={2025}
}

@article{FoundationModel21,
  author    = {Rishi Bommasani and
               Drew A. Hudson and
               Ehsan Adeli and
               Russ Altman and
               Simran Arora and
               Sydney von Arx and
               Michael S. Bernstein and
               Jeannette Bohg and
               Antoine Bosselut and
               Emma Brunskill and
               Erik Brynjolfsson and
               Shyamal Buch and
               Dallas Card and
               Rodrigo Castellon and
               Niladri S. Chatterji and
               Annie S. Chen and
               Kathleen Creel and
               Jared Quincy Davis and
               Dorottya Demszky and
               Chris Donahue and
               Moussa Doumbouya and
               Esin Durmus and
               Stefano Ermon and
               John Etchemendy and
               Kawin Ethayarajh and
               Li Fei{-}Fei and
               Chelsea Finn and
               Trevor Gale and
               Lauren Gillespie and
               Karan Goel and
               Noah D. Goodman and
               Shelby Grossman and
               Neel Guha and
               Tatsunori Hashimoto and
               Peter Henderson and
               John Hewitt and
               Daniel E. Ho and
               Jenny Hong and
               Kyle Hsu and
               Jing Huang and
               Thomas Icard and
               Saahil Jain and
               Dan Jurafsky and
               Pratyusha Kalluri and
               Siddharth Karamcheti and
               Geoff Keeling and
               Fereshte Khani and
               Omar Khattab and
               Pang Wei Koh and
               Mark S. Krass and
               Ranjay Krishna and
               Rohith Kuditipudi and
               et al.},
  title     = {On the Opportunities and Risks of Foundation Models},
  journal   = {CoRR},
  volume    = {abs/2108.07258},
  year      = {2021},
  eprinttype = {arXiv},
  eprint    = {2108.07258},
  bibsource = {dblp computer science bibliography, https://dblp.org}
}

@inproceedings{Flamingo22,
  author       = {Jean{-}Baptiste Alayrac and
                  Jeff Donahue and
                  Pauline Luc and
                  Antoine Miech and
                  Iain Barr and
                  Yana Hasson and
                  Karel Lenc and
                  Arthur Mensch and
                  Katherine Millican and
                  Malcolm Reynolds and
                  Roman Ring and
                  Eliza Rutherford and
                  Serkan Cabi and
                  Tengda Han and
                  Zhitao Gong and
                  Sina Samangooei and
                  Marianne Monteiro and
                  Jacob L. Menick and
                  Sebastian Borgeaud and
                  Andy Brock and
                  Aida Nematzadeh and
                  Sahand Sharifzadeh and
                  Mikolaj Binkowski and
                  Ricardo Barreira and
                  Oriol Vinyals and
                  Andrew Zisserman and
                  Kar{\'{e}}n Simonyan},
  editor       = {Sanmi Koyejo and
                  S. Mohamed and
                  A. Agarwal and
                  Danielle Belgrave and
                  K. Cho and
                  A. Oh},
  title        = {Flamingo: a Visual Language Model for Few-Shot Learning},
  booktitle    = {Advances in Neural Information Processing Systems 35: Annual Conference
                  on Neural Information Processing Systems 2022, NeurIPS 2022, New Orleans,
                  LA, USA, November 28 - December 9, 2022},
  year         = {2022},
  url          = {http://papers.nips.cc/paper\_files/paper/2022/hash/960a172bc7fbf0177ccccbb411a7d800-Abstract-Conference.html},
  bibsource    = {dblp computer science bibliography, https://dblp.org}
}

@inproceedings{LLaVA23,
  author       = {Haotian Liu and
                  Chunyuan Li and
                  Qingyang Wu and
                  Yong Jae Lee},
  editor       = {Alice Oh and
                  Tristan Naumann and
                  Amir Globerson and
                  Kate Saenko and
                  Moritz Hardt and
                  Sergey Levine},
  title        = {Visual Instruction Tuning},
  booktitle    = {Advances in Neural Information Processing Systems 36: Annual Conference
                  on Neural Information Processing Systems 2023, NeurIPS 2023, New Orleans,
                  LA, USA, December 10 - 16, 2023},
  year         = {2023},
  url          = {http://papers.nips.cc/paper\_files/paper/2023/hash/6dcf277ea32ce3288914faf369fe6de0-Abstract-Conference.html},
  bibsource    = {dblp computer science bibliography, https://dblp.org}
}

@inproceedings{RLHF17,
  author       = {Paul F. Christiano and
                  Jan Leike and
                  Tom B. Brown and
                  Miljan Martic and
                  Shane Legg and
                  Dario Amodei},
  editor       = {Isabelle Guyon and
                  Ulrike von Luxburg and
                  Samy Bengio and
                  Hanna M. Wallach and
                  Rob Fergus and
                  S. V. N. Vishwanathan and
                  Roman Garnett},
  title        = {Deep Reinforcement Learning from Human Preferences},
  booktitle    = {Advances in Neural Information Processing Systems 30: Annual Conference
                  on Neural Information Processing Systems 2017, December 4-9, 2017,
                  Long Beach, CA, {USA}},
  pages        = {4299--4307},
  year         = {2017},
  url          = {https://proceedings.neurips.cc/paper/2017/hash/d5e2c0adad503c91f91df240d0cd4e49-Abstract.html},
  bibsource    = {dblp computer science bibliography, https://dblp.org}
}

@article{PPO17,
  author       = {John Schulman and
                  Filip Wolski and
                  Prafulla Dhariwal and
                  Alec Radford and
                  Oleg Klimov},
  title        = {Proximal Policy Optimization Algorithms},
  journal      = {CoRR},
  volume       = {abs/1707.06347},
  year         = {2017},
  url          = {http://arxiv.org/abs/1707.06347},
  eprinttype    = {arXiv},
  eprint       = {1707.06347},
  bibsource    = {dblp computer science bibliography, https://dblp.org}
}

@article{Chameleon24,
  author       = {Chameleon Team},
  title        = {Chameleon: Mixed-Modal Early-Fusion Foundation Models},
  journal      = {CoRR},
  volume       = {abs/2405.09818},
  year         = {2024},
  url          = {https://doi.org/10.48550/arXiv.2405.09818},
  doi          = {10.48550/ARXIV.2405.09818},
  eprinttype    = {arXiv},
  eprint       = {2405.09818},
  bibsource    = {dblp computer science bibliography, https://dblp.org}
}

@article{T2IR125,
  title={T2i-r1: Reinforcing image generation with collaborative semantic-level and token-level cot},
  author={Jiang, Dongzhi and Guo, Ziyu and Zhang, Renrui and Zong, Zhuofan and Li, Hao and Zhuo, Le and Yan, Shilin and Heng, Pheng-Ann and Li, Hongsheng},
  journal={arXiv preprint arXiv:2505.00703},
  URL={https://arxiv.org/abs/2505.00703},
  year={2025}
}

@article{EmbodiedBench25,
  author       = {Rui Yang and
                  Hanyang Chen and
                  Junyu Zhang and
                  Mark Zhao and
                  Cheng Qian and
                  Kangrui Wang and
                  Qineng Wang and
                  Teja Venkat Koripella and
                  Marziyeh Movahedi and
                  Manling Li and
                  Heng Ji and
                  Huan Zhang and
                  Tong Zhang},
  title        = {EmbodiedBench: Comprehensive Benchmarking Multi-modal Large Language
                  Models for Vision-Driven Embodied Agents},
  journal      = {CoRR},
  volume       = {abs/2502.09560},
  year         = {2025},
  url          = {https://doi.org/10.48550/arXiv.2502.09560},
  doi          = {10.48550/ARXIV.2502.09560},
  eprinttype    = {arXiv},
  eprint       = {2502.09560},
  bibsource    = {dblp computer science bibliography, https://dblp.org}
}

@misc{ThinkingWithImages25,
  author       = {{OpenAI}},
  title        = {Thinking with Images},
  year         = {2025},
  month        = {April},
  url          = {https://openai.com/index/thinking-with-images/},
  note         = {Accessed: 2025-05-25}
}

@inproceedings{VisualCoT24,
title={Visual CoT: Advancing Multi-Modal Language Models with a Comprehensive Dataset and Benchmark for Chain-of-Thought Reasoning},
author={Hao Shao and Shengju Qian and Han Xiao and Guanglu Song and Zhuofan Zong and Letian Wang and Yu Liu and Hongsheng Li},
booktitle={The Thirty-eight Conference on Neural Information Processing Systems Datasets and Benchmarks Track},
year={2024},
url={https://openreview.net/forum?id=aXeiCbMFFJ}
}

@article{PerceptionInReflection25,
  author       = {Yana Wei and
                  Liang Zhao and
                  Kangheng Lin and
                  En Yu and
                  Yuang Peng and
                  Runpei Dong and
                  Jianjian Sun and
                  Haoran Wei and
                  Zheng Ge and
                  Xiangyu Zhang and
                  Vishal M. Patel},
  title        = {Perception in Reflection},
  journal      = {CoRR},
  volume       = {abs/2504.07165},
  year         = {2025},
  url          = {https://doi.org/10.48550/arXiv.2504.07165},
  doi          = {10.48550/ARXIV.2504.07165},
  eprinttype    = {arXiv},
  eprint       = {2504.07165},
  bibsource    = {dblp computer science bibliography, https://dblp.org}
}

@article{LLaVACritic24,
  author       = {Tianyi Xiong and
                  Xiyao Wang and
                  Dong Guo and
                  Qinghao Ye and
                  Haoqi Fan and
                  Quanquan Gu and
                  Heng Huang and
                  Chunyuan Li},
  title        = {LLaVA-Critic: Learning to Evaluate Multimodal Models},
  journal      = {CoRR},
  volume       = {abs/2410.02712},
  year         = {2024},
  url          = {https://doi.org/10.48550/arXiv.2410.02712},
  doi          = {10.48550/ARXIV.2410.02712},
  eprinttype    = {arXiv},
  eprint       = {2410.02712},
  bibsource    = {dblp computer science bibliography, https://dblp.org}
}

@article{PerceptionR125,
  author       = {En Yu and
                  Kangheng Lin and
                  Liang Zhao and
                  Jisheng Yin and
                  Yana Wei and
                  Yuang Peng and
                  Haoran Wei and
                  Jianjian Sun and
                  Chunrui Han and
                  Zheng Ge and
                  Xiangyu Zhang and
                  Daxin Jiang and
                  Jingyu Wang and
                  Wenbing Tao},
  title        = {Perception-R1: Pioneering Perception Policy with Reinforcement Learning},
  journal      = {CoRR},
  volume       = {abs/2504.07954},
  year         = {2025},
  url          = {https://doi.org/10.48550/arXiv.2504.07954},
  doi          = {10.48550/ARXIV.2504.07954},
  eprinttype    = {arXiv},
  eprint       = {2504.07954},
  bibsource    = {dblp computer science bibliography, https://dblp.org}
}

@book{WorldISee23,
  title={The Worlds I See: Curiosity, Exploration, and Discovery at the Dawn of AI},
  author={Fei-Fei, Li},
  year={2023},
  url={https://www.google.com/books/edition/The_Worlds_I_See/6BmhEAAAQBAJ?hl=en&gbpv=0},
  publisher={Flatiron books: a moment of lift book}
}

@inproceedings{CoT22,
  author       = {Jason Wei and
                  Xuezhi Wang and
                  Dale Schuurmans and
                  Maarten Bosma and
                  Brian Ichter and
                  Fei Xia and
                  Ed H. Chi and
                  Quoc V. Le and
                  Denny Zhou},
  editor       = {Sanmi Koyejo and
                  S. Mohamed and
                  A. Agarwal and
                  Danielle Belgrave and
                  K. Cho and
                  A. Oh},
  title        = {Chain-of-Thought Prompting Elicits Reasoning in Large Language Models},
  booktitle    = {Advances in Neural Information Processing Systems 35: Annual Conference
                  on Neural Information Processing Systems 2022, NeurIPS 2022, New Orleans,
                  LA, USA, November 28 - December 9, 2022},
  year         = {2022},
  url          = {http://papers.nips.cc/paper\_files/paper/2022/hash/9d5609613524ecf4f15af0f7b31abca4-Abstract-Conference.html},
  bibsource    = {dblp computer science bibliography, https://dblp.org}
}

@inproceedings{SelfConsistency23,
  author       = {Xuezhi Wang and
                  Jason Wei and
                  Dale Schuurmans and
                  Quoc V. Le and
                  Ed H. Chi and
                  Sharan Narang and
                  Aakanksha Chowdhery and
                  Denny Zhou},
  title        = {Self-Consistency Improves Chain of Thought Reasoning in Language Models},
  booktitle    = {The Eleventh International Conference on Learning Representations,
                  {ICLR} 2023, Kigali, Rwanda, May 1-5, 2023},
  publisher    = {OpenReview.net},
  year         = {2023},
  url          = {https://openreview.net/forum?id=1PL1NIMMrw},
  bibsource    = {dblp computer science bibliography, https://dblp.org}
}

@inproceedings{SelfRefine23,
  author       = {Aman Madaan and
                  Niket Tandon and
                  Prakhar Gupta and
                  Skyler Hallinan and
                  Luyu Gao and
                  Sarah Wiegreffe and
                  Uri Alon and
                  Nouha Dziri and
                  Shrimai Prabhumoye and
                  Yiming Yang and
                  Shashank Gupta and
                  Bodhisattwa Prasad Majumder and
                  Katherine Hermann and
                  Sean Welleck and
                  Amir Yazdanbakhsh and
                  Peter Clark},
  editor       = {Alice Oh and
                  Tristan Naumann and
                  Amir Globerson and
                  Kate Saenko and
                  Moritz Hardt and
                  Sergey Levine},
  title        = {Self-Refine: Iterative Refinement with Self-Feedback},
  booktitle    = {Advances in Neural Information Processing Systems 36: Annual Conference
                  on Neural Information Processing Systems 2023, NeurIPS 2023, New Orleans,
                  LA, USA, December 10 - 16, 2023},
  year         = {2023},
  url          = {http://papers.nips.cc/paper\_files/paper/2023/hash/91edff07232fb1b55a505a9e9f6c0ff3-Abstract-Conference.html},
  bibsource    = {dblp computer science bibliography, https://dblp.org}
}

@inproceedings{aliagent,
  author       = {Jingnan Zheng and
                  Han Wang and
                  An Zhang and
                  Tai D. Nguyen and
                  Jun Sun and
                  Tat{-}Seng Chua},
  editor       = {Amir Globersons and
                  Lester Mackey and
                  Danielle Belgrave and
                  Angela Fan and
                  Ulrich Paquet and
                  Jakub M. Tomczak and
                  Cheng Zhang},
  title        = {ALI-Agent: Assessing LLMs' Alignment with Human Values via Agent-based
                  Evaluation},
  booktitle    = {Advances in Neural Information Processing Systems 38: Annual Conference
                  on Neural Information Processing Systems 2024, NeurIPS 2024, Vancouver,
                  BC, Canada, December 10 - 15, 2024},
  year         = {2024},
  url          = {http://papers.nips.cc/paper\_files/paper/2024/hash/b35c38f70065ac6c694089ca93a015bb-Abstract-Conference.html},
  bibsource    = {dblp computer science bibliography, https://dblp.org}
}

@article{steering24,
  author       = {Han Wang and
                  Gang Wang and
                  Huan Zhang},
  title        = {Steering Away from Harm: An Adaptive Approach to Defending Vision
                  Language Model Against Jailbreaks},
  journal      = {CoRR},
  volume       = {abs/2411.16721},
  year         = {2024},
  url          = {https://doi.org/10.48550/arXiv.2411.16721},
  doi          = {10.48550/ARXIV.2411.16721},
  eprinttype    = {arXiv},
  eprint       = {2411.16721},
  bibsource    = {dblp computer science bibliography, https://dblp.org}
}

@inproceedings{DreamLLM24,
  title={Dream{LLM}: Synergistic Multimodal Comprehension and Creation},
  author={Runpei Dong and Chunrui Han and Yuang Peng and Zekun Qi and Zheng Ge and Jinrong Yang and Liang Zhao and Jianjian Sun and Hongyu Zhou and Haoran Wei and Xiangwen Kong and Xiangyu Zhang and Kaisheng Ma and Li Yi},
  booktitle={The Twelfth International Conference on Learning Representations},
  year={2024},
  url={https://openreview.net/forum?id=y01KGvd9Bw}
}

@inproceedings{ShapeLLM24,
  author       = {Zekun Qi and
                  Runpei Dong and
                  Shaochen Zhang and
                  Haoran Geng and
                  Chunrui Han and
                  Zheng Ge and
                  Li Yi and
                  Kaisheng Ma},
  editor       = {Ales Leonardis and
                  Elisa Ricci and
                  Stefan Roth and
                  Olga Russakovsky and
                  Torsten Sattler and
                  G{\"{u}}l Varol},
  title        = {ShapeLLM: Universal 3D Object Understanding for Embodied Interaction},
  booktitle    = {Computer Vision - {ECCV} 2024 - 18th European Conference, Milan, Italy,
                  September 29-October 4, 2024, Proceedings, Part {XLIII}},
  series       = {Lecture Notes in Computer Science},
  volume       = {15101},
  pages        = {214--238},
  publisher    = {Springer},
  year         = {2024},
  url          = {https://doi.org/10.1007/978-3-031-72775-7\_13},
  doi          = {10.1007/978-3-031-72775-7\_13},
}

@article{SoFar25,
  author       = {Zekun Qi and
                  Wenyao Zhang and
                  Yufei Ding and
                  Runpei Dong and
                  Xinqiang Yu and
                  Jingwen Li and
                  Lingyun Xu and
                  Baoyu Li and
                  Xialin He and
                  Guofan Fan and
                  Jiazhao Zhang and
                  Jiawei He and
                  Jiayuan Gu and
                  Xin Jin and
                  Kaisheng Ma and
                  Zhizheng Zhang and
                  He Wang and
                  Li Yi},
  title        = {SoFar: Language-Grounded Orientation Bridges Spatial Reasoning and
                  Object Manipulation},
  journal      = {CoRR},
  volume       = {abs/2502.13143},
  year         = {2025},
  url          = {https://doi.org/10.48550/arXiv.2502.13143},
  doi          = {10.48550/ARXIV.2502.13143},
  eprinttype    = {arXiv},
  eprint       = {2502.13143},
}

@article{GPT4o24,
  author    = {OpenAI},
  title     = {Introducing GPT-4o and more tools to ChatGPT free users},
  year      = {2024},
  url       = {https://openai.com/index/gpt-4o-and-more-tools-to-chatgpt-free/}
}

@inproceedings{VStar24,
  author       = {Penghao Wu and
                  Saining Xie},
  title        = {V*: Guided Visual Search as a Core Mechanism in Multimodal LLMs},
  booktitle    = {{IEEE/CVF} Conference on Computer Vision and Pattern Recognition,
                  {CVPR} 2024, Seattle, WA, USA, June 16-22, 2024},
  pages        = {13084--13094},
  publisher    = {{IEEE}},
  year         = {2024},
  url          = {https://doi.org/10.1109/CVPR52733.2024.01243},
  doi          = {10.1109/CVPR52733.2024.01243},
  bibsource    = {dblp computer science bibliography, https://dblp.org}
}

@inproceedings{Volcano24,
  author       = {Seongyun Lee and
                  Sue Hyun Park and
                  Yongrae Jo and
                  Minjoon Seo},
  editor       = {Kevin Duh and
                  Helena G{\'{o}}mez{-}Adorno and
                  Steven Bethard},
  title        = {Volcano: Mitigating Multimodal Hallucination through Self-Feedback
                  Guided Revision},
  booktitle    = {Proceedings of the 2024 Conference of the North American Chapter of
                  the Association for Computational Linguistics: Human Language Technologies
                  (Volume 1: Long Papers), {NAACL} 2024, Mexico City, Mexico, June 16-21,
                  2024},
  pages        = {391--404},
  publisher    = {Association for Computational Linguistics},
  year         = {2024},
  url          = {https://doi.org/10.18653/v1/2024.naacl-long.23},
  doi          = {10.18653/V1/2024.NAACL-LONG.23},
  bibsource    = {dblp computer science bibliography, https://dblp.org}
}

@article{SlowPerception24,
  author       = {Haoran Wei and
                  Youyang Yin and
                  Yumeng Li and
                  Jia Wang and
                  Liang Zhao and
                  Jianjian Sun and
                  Zheng Ge and
                  Xiangyu Zhang},
  title        = {Slow Perception: Let's Perceive Geometric Figures Step-by-step},
  journal      = {CoRR},
  volume       = {abs/2412.20631},
  year         = {2024},
  url          = {https://doi.org/10.48550/arXiv.2412.20631},
  doi          = {10.48550/ARXIV.2412.20631},
  eprinttype    = {arXiv},
  eprint       = {2412.20631},
  bibsource    = {dblp computer science bibliography, https://dblp.org}
}

@article{CoCoNut24,
  author       = {Shibo Hao and
                  Sainbayar Sukhbaatar and
                  DiJia Su and
                  Xian Li and
                  Zhiting Hu and
                  Jason Weston and
                  Yuandong Tian},
  title        = {Training Large Language Models to Reason in a Continuous Latent Space},
  journal      = {CoRR},
  volume       = {abs/2412.06769},
  year         = {2024},
  url          = {https://doi.org/10.48550/arXiv.2412.06769},
  doi          = {10.48550/ARXIV.2412.06769},
  eprinttype    = {arXiv},
  eprint       = {2412.06769},
  bibsource    = {dblp computer science bibliography, https://dblp.org}
}

@article{S1_25,
  author       = {Niklas Muennighoff and
                  Zitong Yang and
                  Weijia Shi and
                  Xiang Lisa Li and
                  Li Fei{-}Fei and
                  Hannaneh Hajishirzi and
                  Luke Zettlemoyer and
                  Percy Liang and
                  Emmanuel J. Cand{\`{e}}s and
                  Tatsunori Hashimoto},
  title        = {s1: Simple test-time scaling},
  journal      = {CoRR},
  volume       = {abs/2501.19393},
  year         = {2025},
  url          = {https://doi.org/10.48550/arXiv.2501.19393},
  doi          = {10.48550/ARXIV.2501.19393},
  eprinttype    = {arXiv},
  eprint       = {2501.19393},
  bibsource    = {dblp computer science bibliography, https://dblp.org}
}

@inproceedings{LiveCodeBench25,
  author       = {Naman Jain and
                  King Han and
                  Alex Gu and
                  Wen{-}Ding Li and
                  Fanjia Yan and
                  Tianjun Zhang and
                  Sida Wang and
                  Armando Solar{-}Lezama and
                  Koushik Sen and
                  Ion Stoica},
  title        = {LiveCodeBench: Holistic and Contamination Free Evaluation of Large
                  Language Models for Code},
  booktitle    = {The Thirteenth International Conference on Learning Representations,
                  {ICLR} 2025, Singapore, April 24-28, 2025},
  publisher    = {OpenReview.net},
  year         = {2025},
  url          = {https://openreview.net/forum?id=chfJJYC3iL},
  bibsource    = {dblp computer science bibliography, https://dblp.org}
}

@inproceedings{MinervaMath22,
  author       = {Aitor Lewkowycz and
                  Anders Andreassen and
                  David Dohan and
                  Ethan Dyer and
                  Henryk Michalewski and
                  Vinay V. Ramasesh and
                  Ambrose Slone and
                  Cem Anil and
                  Imanol Schlag and
                  Theo Gutman{-}Solo and
                  Yuhuai Wu and
                  Behnam Neyshabur and
                  Guy Gur{-}Ari and
                  Vedant Misra},
  editor       = {Sanmi Koyejo and
                  S. Mohamed and
                  A. Agarwal and
                  Danielle Belgrave and
                  K. Cho and
                  A. Oh},
  title        = {Solving Quantitative Reasoning Problems with Language Models},
  booktitle    = {Advances in Neural Information Processing Systems 35: Annual Conference
                  on Neural Information Processing Systems 2022, NeurIPS 2022, New Orleans,
                  LA, USA, November 28 - December 9, 2022},
  year         = {2022},
  url          = {http://papers.nips.cc/paper\_files/paper/2022/hash/18abbeef8cfe9203fdf9053c9c4fe191-Abstract-Conference.html},
  bibsource    = {dblp computer science bibliography, https://dblp.org}
}

@misc{AMC23,
  author       = {{AI-MO}},
  title        = {{AIMO Validation Dataset - AMC}},
  year         = {2024},
  url          = {https://huggingface.co/datasets/AI-MO/aimo-validation-amc},
  note         = {Accessed: 2025-05-19},
  howpublished = {\url{https://huggingface.co/datasets/AI-MO/aimo-validation-amc}},
  publisher    = {Hugging Face}
}

@article{DynamicEarlyExit25,
  title={Dynamic Early Exit in Reasoning Models},
  author={Yang, Chenxu and Si, Qingyi and Duan, Yongjie and Zhu, Zheliang and Zhu, Chenyu and Lin, Zheng and Cao, Li and Wang, Weiping},
  journal={arXiv preprint arXiv:2504.15895},
  url={https://arxiv.org/abs/2504.15895},
  year={2025}
}

@inproceedings{Dynamath,
  author       = {Chengke Zou and
                  Xingang Guo and
                  Rui Yang and
                  Junyu Zhang and
                  Bin Hu and
                  Huan Zhang},
  title        = {DynaMath: {A} Dynamic Visual Benchmark for Evaluating Mathematical
                  Reasoning Robustness of Vision Language Models},
  booktitle    = {The Thirteenth International Conference on Learning Representations,
                  {ICLR} 2025, Singapore, April 24-28, 2025},
  publisher    = {OpenReview.net},
  year         = {2025},
  url          = {https://openreview.net/forum?id=VOAMTA8jKu},
  bibsource    = {dblp computer science bibliography, https://dblp.org}
}

@article{DeepSeekR125,
  author       = {DeepSeek{-}AI and
                  Daya Guo and
                  Dejian Yang and
                  Haowei Zhang and
                  Junxiao Song and
                  Ruoyu Zhang and
                  Runxin Xu and
                  Qihao Zhu and
                  Shirong Ma and
                  Peiyi Wang and
                  Xiao Bi and
                  Xiaokang Zhang and
                  Xingkai Yu and
                  Yu Wu and
                  Z. F. Wu and
                  Zhibin Gou and
                  Zhihong Shao and
                  Zhuoshu Li and
                  Ziyi Gao and
                  Aixin Liu and
                  Bing Xue and
                  Bingxuan Wang and
                  Bochao Wu and
                  Bei Feng and
                  Chengda Lu and
                  Chenggang Zhao and
                  Chengqi Deng and
                  Chenyu Zhang and
                  Chong Ruan and
                  Damai Dai and
                  Deli Chen and
                  Dongjie Ji and
                  Erhang Li and
                  Fangyun Lin and
                  Fucong Dai and
                  Fuli Luo and
                  Guangbo Hao and
                  Guanting Chen and
                  Guowei Li and
                  H. Zhang and
                  Han Bao and
                  Hanwei Xu and
                  Haocheng Wang and
                  Honghui Ding and
                  Huajian Xin and
                  Huazuo Gao and
                  Hui Qu and
                  Hui Li and
                  Jianzhong Guo and
                  Jiashi Li and
                  Jiawei Wang and
                  Jingchang Chen and
                  Jingyang Yuan and
                  Junjie Qiu and
                  Junlong Li and
                  J. L. Cai and
                  Jiaqi Ni and
                  Jian Liang and
                  Jin Chen and
                  Kai Dong and
                  Kai Hu and
                  Kaige Gao and
                  Kang Guan and
                  Kexin Huang and
                  Kuai Yu and
                  Lean Wang and
                  Lecong Zhang and
                  Liang Zhao and
                  Litong Wang and
                  Liyue Zhang and
                  Lei Xu and
                  Leyi Xia and
                  Mingchuan Zhang and
                  Minghua Zhang and
                  Minghui Tang and
                  Meng Li and
                  Miaojun Wang and
                  Mingming Li and
                  Ning Tian and
                  Panpan Huang and
                  Peng Zhang and
                  Qiancheng Wang and
                  Qinyu Chen and
                  Qiushi Du and
                  Ruiqi Ge and
                  Ruisong Zhang and
                  Ruizhe Pan and
                  Runji Wang and
                  R. J. Chen and
                  R. L. Jin and
                  Ruyi Chen and
                  Shanghao Lu and
                  Shangyan Zhou and
                  Shanhuang Chen and
                  Shengfeng Ye and
                  Shiyu Wang and
                  Shuiping Yu and
                  Shunfeng Zhou and
                  Shuting Pan and
                  S. S. Li},
  title        = {DeepSeek-R1: Incentivizing Reasoning Capability in LLMs via Reinforcement
                  Learning},
  journal      = {CoRR},
  volume       = {abs/2501.12948},
  year         = {2025},
  url          = {https://doi.org/10.48550/arXiv.2501.12948},
  doi          = {10.48550/ARXIV.2501.12948},
  eprinttype    = {arXiv},
  eprint       = {2501.12948},
  bibsource    = {dblp computer science bibliography, https://dblp.org}
}

@misc{QwenQwQ25,
  author       = {{Qwen Team}},
  title        = {Preview of qwen qwen1.5-32b},
  howpublished = {\url{https://qwenlm.github.io/blog/qwq-32b-preview/}},
  year         = {2025},
  note         = {Accessed: 2025-03-20}
}

@article{SpeculativeThinking25,
  title={Speculative Thinking: Enhancing Small-Model Reasoning with Large Model Guidance at Inference Time},
  author={Yang, Wang and Yue, Xiang and Chaudhary, Vipin and Han, Xiaotian},
  journal={arXiv preprint arXiv:2504.12329},
  url={https://arxiv.org/abs/2504.12329},
  year={2025}
}

@article{DynamicSelfConsistency24,
  author       = {Guangya Wan and
                  Yuqi Wu and
                  Jie Chen and
                  Sheng Li},
  title        = {Dynamic Self-Consistency: Leveraging Reasoning Paths for Efficient
                  {LLM} Sampling},
  journal      = {CoRR},
  volume       = {abs/2408.17017},
  year         = {2024},
  url          = {https://doi.org/10.48550/arXiv.2408.17017},
  doi          = {10.48550/ARXIV.2408.17017},
  eprinttype    = {arXiv},
  eprint       = {2408.17017},
  bibsource    = {dblp computer science bibliography, https://dblp.org}
}

@inproceedings{LongContextPerplexity25,
  author       = {Lizhe Fang and
                  Yifei Wang and
                  Zhaoyang Liu and
                  Chenheng Zhang and
                  Stefanie Jegelka and
                  Jinyang Gao and
                  Bolin Ding and
                  Yisen Wang},
  title        = {What is Wrong with Perplexity for Long-context Language Modeling?},
  booktitle    = {The Thirteenth International Conference on Learning Representations,
                  {ICLR} 2025, Singapore, April 24-28, 2025},
  publisher    = {OpenReview.net},
  year         = {2025},
  url          = {https://openreview.net/forum?id=fL4qWkSmtM},
  bibsource    = {dblp computer science bibliography, https://dblp.org}
}

@article{InternalProbSelfConsistency25,
  author       = {Zhi Zhou and
                  Tan Yuhao and
                  Zenan Li and
                  Yuan Yao and
                  Lan{-}Zhe Guo and
                  Xiaoxing Ma and
                  Yu{-}Feng Li},
  title        = {Bridging Internal Probability and Self-Consistency for Effective and
                  Efficient {LLM} Reasoning},
  journal      = {CoRR},
  volume       = {abs/2502.00511},
  year         = {2025},
  url          = {https://doi.org/10.48550/arXiv.2502.00511},
  doi          = {10.48550/ARXIV.2502.00511},
  eprinttype    = {arXiv},
  eprint       = {2502.00511},
  bibsource    = {dblp computer science bibliography, https://dblp.org}
}

@misc{AIME24,
  author       = {{Mathematical Association of America}},
  title        = {{American Invitational Mathematics Examination -- AIME}},
  howpublished = {\emph{American Invitational Mathematics Examination -- AIME 2024}},
  month        = feb,
  year         = 2024,
  note         = {Accessed: 2025-05-15},
  url          = {https://maa.org/math-competitions/american-invitational-mathematics-examination-aime}
}

@inproceedings{MATH500_24,
  author       = {Hunter Lightman and
                  Vineet Kosaraju and
                  Yuri Burda and
                  Harrison Edwards and
                  Bowen Baker and
                  Teddy Lee and
                  Jan Leike and
                  John Schulman and
                  Ilya Sutskever and
                  Karl Cobbe},
  title        = {Let's Verify Step by Step},
  booktitle    = {The Twelfth International Conference on Learning Representations,
                  {ICLR} 2024, Vienna, Austria, May 7-11, 2024},
  publisher    = {OpenReview.net},
  year         = {2024},
  url          = {https://openreview.net/forum?id=v8L0pN6EOi},
  bibsource    = {dblp computer science bibliography, https://dblp.org}
}

@article{OpenAIO1_24,
  author       = {Aaron Jaech and
                  Adam Kalai and
                  Adam Lerer and
                  Adam Richardson and
                  Ahmed El{-}Kishky and
                  Aiden Low and
                  Alec Helyar and
                  Aleksander Madry and
                  Alex Beutel and
                  Alex Carney and
                  Alex Iftimie and
                  Alex Karpenko and
                  Alex Tachard Passos and
                  Alexander Neitz and
                  Alexander Prokofiev and
                  Alexander Wei and
                  Allison Tam and
                  Ally Bennett and
                  Ananya Kumar and
                  Andre Saraiva and
                  Andrea Vallone and
                  Andrew Duberstein and
                  Andrew Kondrich and
                  Andrey Mishchenko and
                  Andy Applebaum and
                  Angela Jiang and
                  Ashvin Nair and
                  Barret Zoph and
                  Behrooz Ghorbani and
                  Ben Rossen and
                  Benjamin Sokolowsky and
                  Boaz Barak and
                  Bob McGrew and
                  Borys Minaiev and
                  Botao Hao and
                  Bowen Baker and
                  Brandon Houghton and
                  Brandon McKinzie and
                  Brydon Eastman and
                  Camillo Lugaresi and
                  Cary Bassin and
                  Cary Hudson and
                  Chak Ming Li and
                  Charles de Bourcy and
                  Chelsea Voss and
                  Chen Shen and
                  Chong Zhang and
                  Chris Koch and
                  Chris Orsinger and
                  Christopher Hesse and
                  Claudia Fischer and
                  Clive Chan and
                  Dan Roberts and
                  Daniel Kappler and
                  Daniel Levy and
                  Daniel Selsam and
                  David Dohan and
                  David Farhi and
                  David Mely and
                  David Robinson and
                  Dimitris Tsipras and
                  Doug Li and
                  Dragos Oprica and
                  Eben Freeman and
                  Eddie Zhang and
                  Edmund Wong and
                  Elizabeth Proehl and
                  Enoch Cheung and
                  Eric Mitchell and
                  Eric Wallace and
                  Erik Ritter and
                  Evan Mays and
                  Fan Wang and
                  Felipe Petroski Such and
                  Filippo Raso and
                  Florencia Leoni and
                  Foivos Tsimpourlas and
                  Francis Song and
                  Fred von Lohmann and
                  Freddie Sulit and
                  Geoff Salmon and
                  Giambattista Parascandolo and
                  Gildas Chabot and
                  Grace Zhao and
                  Greg Brockman and
                  Guillaume Leclerc and
                  Hadi Salman and
                  Haiming Bao and
                  Hao Sheng and
                  Hart Andrin and
                  Hessam Bagherinezhad and
                  Hongyu Ren and
                  Hunter Lightman and
                  Hyung Won Chung and
                  Ian Kivlichan and
                  Ian O'Connell and
                  Ian Osband and
                  Ignasi Clavera Gilaberte and
                  Ilge Akkaya},
  title        = {OpenAI o1 System Card},
  journal      = {CoRR},
  volume       = {abs/2412.16720},
  year         = {2024},
  url          = {https://doi.org/10.48550/arXiv.2412.16720},
  doi          = {10.48550/ARXIV.2412.16720},
  eprinttype    = {arXiv},
  eprint       = {2412.16720},
  bibsource    = {dblp computer science bibliography, https://dblp.org}
}

@inproceedings{Star22,
  author       = {Eric Zelikman and
                  Yuhuai Wu and
                  Jesse Mu and
                  Noah D. Goodman},
  editor       = {Sanmi Koyejo and
                  S. Mohamed and
                  A. Agarwal and
                  Danielle Belgrave and
                  K. Cho and
                  A. Oh},
  title        = {STaR: Bootstrapping Reasoning With Reasoning},
  booktitle    = {Advances in Neural Information Processing Systems 35: Annual Conference
                  on Neural Information Processing Systems 2022, NeurIPS 2022, New Orleans,
                  LA, USA, November 28 - December 9, 2022},
  year         = {2022},
  url          = {http://papers.nips.cc/paper\_files/paper/2022/hash/639a9a172c044fbb64175b5fad42e9a5-Abstract-Conference.html},
  bibsource    = {dblp computer science bibliography, https://dblp.org}
}

@article{ContrastiveMCTS24,
  author       = {Zitian Gao and
                  Boye Niu and
                  Xuzheng He and
                  Haotian Xu and
                  Hongzhang Liu and
                  Aiwei Liu and
                  Xuming Hu and
                  Lijie Wen},
  title        = {Interpretable Contrastive Monte Carlo Tree Search Reasoning},
  journal      = {CoRR},
  volume       = {abs/2410.01707},
  year         = {2024},
  url          = {https://doi.org/10.48550/arXiv.2410.01707},
  doi          = {10.48550/ARXIV.2410.01707},
  eprinttype    = {arXiv},
  eprint       = {2410.01707},
  bibsource    = {dblp computer science bibliography, https://dblp.org}
}

@book{ThinkingFastAndSlow,
  title={Thinking, fast and slow},
  author={Kahneman, Daniel},
  year={2011},
  publisher={macmillan},
  url={https://www.google.com/books/edition/_/SHvzzuCnuv8C?hl=en}
}

@article{O1ReplicationJourneyII24,
  author       = {Zhen Huang and
                  Haoyang Zou and
                  Xuefeng Li and
                  Yixiu Liu and
                  Yuxiang Zheng and
                  Ethan Chern and
                  Shijie Xia and
                  Yiwei Qin and
                  Weizhe Yuan and
                  Pengfei Liu},
  title        = {{O1} Replication Journey - Part 2: Surpassing O1-preview through Simple
                  Distillation, Big Progress or Bitter Lesson?},
  journal      = {CoRR},
  volume       = {abs/2411.16489},
  year         = {2024},
  url          = {https://doi.org/10.48550/arXiv.2411.16489},
  doi          = {10.48550/ARXIV.2411.16489},
  eprinttype    = {arXiv},
  eprint       = {2411.16489},
  bibsource    = {dblp computer science bibliography, https://dblp.org}
}

@article{PRMandORM22,
  author       = {Jonathan Uesato and
                  Nate Kushman and
                  Ramana Kumar and
                  H. Francis Song and
                  Noah Y. Siegel and
                  Lisa Wang and
                  Antonia Creswell and
                  Geoffrey Irving and
                  Irina Higgins},
  title        = {Solving math word problems with process- and outcome-based feedback},
  journal      = {CoRR},
  volume       = {abs/2211.14275},
  year         = {2022},
  url          = {https://doi.org/10.48550/arXiv.2211.14275},
  doi          = {10.48550/ARXIV.2211.14275},
  eprinttype    = {arXiv},
  eprint       = {2211.14275},
  bibsource    = {dblp computer science bibliography, https://dblp.org}
}

@article{Phi4_24,
  title={Phi-4 technical report},
  author={Marah Abdin and Jyoti Aneja and Harkirat Behl and Sébastien Bubeck and Ronen Eldan and Suriya Gunasekar and Michael Harrison and Russell J. Hewett and Mojan Javaheripi and Piero Kauffmann and James R. Lee and Yin Tat Lee and Yuanzhi Li and Weishung Liu and Caio C. T. Mendes and Anh Nguyen and Eric Price and Gustavo de Rosa and Olli Saarikivi and Adil Salim and Shital Shah and Xin Wang and Rachel Ward and Yue Wu and Dingli Yu and Cyril Zhang and Yi Zhang},
  url={https://arxiv.org/abs/2412.08905},
  journal      = {CoRR},
  volume       = {abs/2412.08905},
  eprinttype    = {arXiv},
  eprint       = {2412.08905},
  year={2024}
}

@article{EffectiveWithoutThinking25,
  title={Reasoning Models Can Be Effective Without Thinking},
  author={Ma, Wenjie and He, Jingxuan and Snell, Charlie and Griggs, Tyler and Min, Sewon and Zaharia, Matei},
  journal={arXiv preprint arXiv:2504.09858},
  url={https://arxiv.org/abs/2504.09858},
  year={2025}
}

@article{StopOverthinking25,
  author       = {Yang Sui and
                  Yu{-}Neng Chuang and
                  Guanchu Wang and
                  Jiamu Zhang and
                  Tianyi Zhang and
                  Jiayi Yuan and
                  Hongyi Liu and
                  Andrew Wen and
                  Shaochen Zhong and
                  Hanjie Chen and
                  Xia Ben Hu},
  title        = {Stop Overthinking: {A} Survey on Efficient Reasoning for Large Language
                  Models},
  journal      = {CoRR},
  volume       = {abs/2503.16419},
  year         = {2025},
  url          = {https://doi.org/10.48550/arXiv.2503.16419},
  doi          = {10.48550/ARXIV.2503.16419},
  eprinttype    = {arXiv},
  eprint       = {2503.16419},
  bibsource    = {dblp computer science bibliography, https://dblp.org}
}

@article{BetweenUnderThinkingandOverthinking25,
  title={Between Underthinking and Overthinking: An Empirical Study of Reasoning Length and correctness in LLMs},
  author={Su, Jinyan and Healey, Jennifer and Nakov, Preslav and Cardie, Claire},
  journal={arXiv preprint arXiv:2505.00127},
  url={https://arxiv.org/abs/2505.00127},
  year={2025}
}

@article{ThoughtTerminatorOverThinking25,
  title={THOUGHTTERMINATOR: Benchmarking, Calibrating, and Mitigating Overthinking in Reasoning Models},
  author={Pu, Xiao and Saxon, Michael and Hua, Wenyue and Wang, William Yang},
  journal={arXiv preprint arXiv:2504.13367},
  url={https://arxiv.org/abs/2504.13367},
  year={2025}
}

@inproceedings{TreeOfThoughts23,
  author       = {Shunyu Yao and
                  Dian Yu and
                  Jeffrey Zhao and
                  Izhak Shafran and
                  Tom Griffiths and
                  Yuan Cao and
                  Karthik Narasimhan},
  editor       = {Alice Oh and
                  Tristan Naumann and
                  Amir Globerson and
                  Kate Saenko and
                  Moritz Hardt and
                  Sergey Levine},
  title        = {Tree of Thoughts: Deliberate Problem Solving with Large Language Models},
  booktitle    = {Advances in Neural Information Processing Systems 36: Annual Conference
                  on Neural Information Processing Systems 2023, NeurIPS 2023, New Orleans,
                  LA, USA, December 10 - 16, 2023},
  year         = {2023},
  url          = {http://papers.nips.cc/paper\_files/paper/2023/hash/271db9922b8d1f4dd7aaef84ed5ac703-Abstract-Conference.html},
  bibsource    = {dblp computer science bibliography, https://dblp.org}
}

@article{OverThinkingO1Like24,
  author       = {Xingyu Chen and
                  Jiahao Xu and
                  Tian Liang and
                  Zhiwei He and
                  Jianhui Pang and
                  Dian Yu and
                  Linfeng Song and
                  Qiuzhi Liu and
                  Mengfei Zhou and
                  Zhuosheng Zhang and
                  Rui Wang and
                  Zhaopeng Tu and
                  Haitao Mi and
                  Dong Yu},
  title        = {Do {NOT} Think That Much for 2+3=? On the Overthinking of o1-Like
                  LLMs},
  journal      = {CoRR},
  volume       = {abs/2412.21187},
  year         = {2024},
  url          = {https://doi.org/10.48550/arXiv.2412.21187},
  doi          = {10.48550/ARXIV.2412.21187},
  eprinttype    = {arXiv},
  eprint       = {2412.21187},
  bibsource    = {dblp computer science bibliography, https://dblp.org}
}

@article{UnderthinkingO1Like25,
  author       = {Yue Wang and
                  Qiuzhi Liu and
                  Jiahao Xu and
                  Tian Liang and
                  Xingyu Chen and
                  Zhiwei He and
                  Linfeng Song and
                  Dian Yu and
                  Juntao Li and
                  Zhuosheng Zhang and
                  Rui Wang and
                  Zhaopeng Tu and
                  Haitao Mi and
                  Dong Yu},
  title        = {Thoughts Are All Over the Place: On the Underthinking of o1-Like LLMs},
  journal      = {CoRR},
  volume       = {abs/2501.18585},
  year         = {2025},
  url          = {https://doi.org/10.48550/arXiv.2501.18585},
  doi          = {10.48550/ARXIV.2501.18585},
  eprinttype    = {arXiv},
  eprint       = {2501.18585},
  bibsource    = {dblp computer science bibliography, https://dblp.org}
}

@article{ThinkingOptimalScaling25,
  author       = {Wenkai Yang and
                  Shuming Ma and
                  Yankai Lin and
                  Furu Wei},
  title        = {Towards Thinking-Optimal Scaling of Test-Time Compute for {LLM} Reasoning},
  journal      = {CoRR},
  volume       = {abs/2502.18080},
  year         = {2025},
  url          = {https://doi.org/10.48550/arXiv.2502.18080},
  doi          = {10.48550/ARXIV.2502.18080},
  eprinttype    = {arXiv},
  eprint       = {2502.18080},
  bibsource    = {dblp computer science bibliography, https://dblp.org}
}

@inproceedings{WithoutSelfDoubt25,
  title={Reasoning without self-doubt: More efficient chain-of-thought through certainty probing},
  author={Fu, Yichao and Chen, Junda and Zhuang, Yonghao and Fu, Zheyu and Stoica, Ion and Zhang, Hao},
  booktitle={ICLR 2025 Workshop on Foundation Models in the Wild},
  url={https://openreview.net/pdf?id=wpK4IMJfdX},
  year={2025}
}

@article{ChainOfDraft25,
  author       = {Silei Xu and
                  Wenhao Xie and
                  Lingxiao Zhao and
                  Pengcheng He},
  title        = {Chain of Draft: Thinking Faster by Writing Less},
  journal      = {CoRR},
  volume       = {abs/2502.18600},
  year         = {2025},
  url          = {https://doi.org/10.48550/arXiv.2502.18600},
  doi          = {10.48550/ARXIV.2502.18600},
  eprinttype    = {arXiv},
  eprint       = {2502.18600},
  bibsource    = {dblp computer science bibliography, https://dblp.org}
}

@inproceedings{GraphOfThoughts24,
  author       = {Maciej Besta and
                  Nils Blach and
                  Ales Kubicek and
                  Robert Gerstenberger and
                  Michal Podstawski and
                  Lukas Gianinazzi and
                  Joanna Gajda and
                  Tomasz Lehmann and
                  Hubert Niewiadomski and
                  Piotr Nyczyk and
                  Torsten Hoefler},
  editor       = {Michael J. Wooldridge and
                  Jennifer G. Dy and
                  Sriraam Natarajan},
  title        = {Graph of Thoughts: Solving Elaborate Problems with Large Language
                  Models},
  booktitle    = {Thirty-Eighth {AAAI} Conference on Artificial Intelligence, {AAAI}
                  2024, Thirty-Sixth Conference on Innovative Applications of Artificial
                  Intelligence, {IAAI} 2024, Fourteenth Symposium on Educational Advances
                  in Artificial Intelligence, {EAAI} 2014, February 20-27, 2024, Vancouver,
                  Canada},
  pages        = {17682--17690},
  publisher    = {{AAAI} Press},
  year         = {2024},
  url          = {https://doi.org/10.1609/aaai.v38i16.29720},
  doi          = {10.1609/AAAI.V38I16.29720},
  bibsource    = {dblp computer science bibliography, https://dblp.org}
}

@article{RevisitTestTime25,
  author       = {Zhiyuan Zeng and
                  Qinyuan Cheng and
                  Zhangyue Yin and
                  Yunhua Zhou and
                  Xipeng Qiu},
  title        = {Revisiting the Test-Time Scaling of o1-like Models: Do they Truly
                  Possess Test-Time Scaling Capabilities?},
  journal      = {CoRR},
  volume       = {abs/2502.12215},
  year         = {2025},
  url          = {https://doi.org/10.48550/arXiv.2502.12215},
  doi          = {10.48550/ARXIV.2502.12215},
  eprinttype    = {arXiv},
  eprint       = {2502.12215},
  bibsource    = {dblp computer science bibliography, https://dblp.org}
}

@article{TrainingVerifiers21,
  author       = {Karl Cobbe and
                  Vineet Kosaraju and
                  Mohammad Bavarian and
                  Mark Chen and
                  Heewoo Jun and
                  Lukasz Kaiser and
                  Matthias Plappert and
                  Jerry Tworek and
                  Jacob Hilton and
                  Reiichiro Nakano and
                  Christopher Hesse and
                  John Schulman},
  title        = {Training Verifiers to Solve Math Word Problems},
  journal      = {CoRR},
  volume       = {abs/2110.14168},
  year         = {2021},
  url          = {https://arxiv.org/abs/2110.14168},
  eprinttype    = {arXiv},
  eprint       = {2110.14168},
  bibsource    = {dblp computer science bibliography, https://dblp.org}
}

@inproceedings{FastBestOfN24,
title={Fast Best-of-N Decoding via Speculative Rejection},
author={Hanshi Sun and Momin Haider and Ruiqi Zhang and Huitao Yang and Jiahao Qiu and Ming Yin and Mengdi Wang and Peter Bartlett and Andrea Zanette},
booktitle={The Thirty-eighth Annual Conference on Neural Information Processing Systems},
year={2024},
url={https://openreview.net/forum?id=348hfcprUs}
}

@inproceedings{BoNBoN24,
  author       = {Lin Gui and
                  Cristina Garbacea and
                  Victor Veitch},
  editor       = {Amir Globersons and
                  Lester Mackey and
                  Danielle Belgrave and
                  Angela Fan and
                  Ulrich Paquet and
                  Jakub M. Tomczak and
                  Cheng Zhang},
  title        = {BoNBoN Alignment for Large Language Models and the Sweetness of Best-of-n
                  Sampling},
  booktitle    = {Advances in Neural Information Processing Systems 38: Annual Conference
                  on Neural Information Processing Systems 2024, NeurIPS 2024, Vancouver,
                  BC, Canada, December 10 - 15, 2024},
  year         = {2024},
  url          = {http://papers.nips.cc/paper\_files/paper/2024/hash/056521a35eacd9d2127b66a7d3c499c5-Abstract-Conference.html},
  bibsource    = {dblp computer science bibliography, https://dblp.org}
}

@inproceedings{VBoN25,
  author       = {Afra Amini and
                  Tim Vieira and
                  Elliott Ash and
                  Ryan Cotterell},
  title        = {Variational Best-of-N Alignment},
  booktitle    = {The Thirteenth International Conference on Learning Representations,
                  {ICLR} 2025, Singapore, April 24-28, 2025},
  publisher    = {OpenReview.net},
  year         = {2025},
  url          = {https://openreview.net/forum?id=W9FZEQj3vv},
  bibsource    = {dblp computer science bibliography, https://dblp.org}
}

@inproceedings{Bond25,
  author       = {Pier Giuseppe Sessa and
                  Robert Dadashi{-}Tazehozi and
                  L{\'{e}}onard Hussenot and
                  Johan Ferret and
                  Nino Vieillard and
                  Alexandre Ram{\'{e}} and
                  Bobak Shahriari and
                  Sarah Perrin and
                  Abram L. Friesen and
                  Geoffrey Cideron and
                  Sertan Girgin and
                  Piotr Stanczyk and
                  Andrea Michi and
                  Danila Sinopalnikov and
                  Sabela Ramos Garea and
                  Am{\'{e}}lie H{\'{e}}liou and
                  Aliaksei Severyn and
                  Matthew Hoffman and
                  Nikola Momchev and
                  Olivier Bachem},
  title        = {{BOND:} Aligning LLMs with Best-of-N Distillation},
  booktitle    = {The Thirteenth International Conference on Learning Representations,
                  {ICLR} 2025, Singapore, April 24-28, 2025},
  publisher    = {OpenReview.net},
  year         = {2025},
  url          = {https://openreview.net/forum?id=0tAXMiSufG},
  bibsource    = {dblp computer science bibliography, https://dblp.org}
}

@inproceedings{InferenceAwareFinetuningBoN25,
  author       = {Yinlam Chow and
                  Guy Tennenholtz and
                  Izzeddin Gur and
                  Vincent Zhuang and
                  Bo Dai and
                  Aviral Kumar and
                  Rishabh Agarwal and
                  Sridhar Thiagarajan and
                  Craig Boutilier and
                  Aleksandra Faust},
  title        = {Inference-Aware Fine-Tuning for Best-of-N Sampling in Large Language
                  Models},
  booktitle    = {The Thirteenth International Conference on Learning Representations,
                  {ICLR} 2025, Singapore, April 24-28, 2025},
  publisher    = {OpenReview.net},
  year         = {2025},
  url          = {https://openreview.net/forum?id=77gQUdQhE7},
  bibsource    = {dblp computer science bibliography, https://dblp.org}
}

@article{TreeBoN24,
  author       = {Jiahao Qiu and
                  Yifu Lu and
                  Yifan Zeng and
                  Jiacheng Guo and
                  Jiayi Geng and
                  Huazheng Wang and
                  Kaixuan Huang and
                  Yue Wu and
                  Mengdi Wang},
  title        = {TreeBoN: Enhancing Inference-Time Alignment with Speculative Tree-Search
                  and Best-of-N Sampling},
  journal      = {CoRR},
  volume       = {abs/2410.16033},
  year         = {2024},
  url          = {https://doi.org/10.48550/arXiv.2410.16033},
  doi          = {10.48550/ARXIV.2410.16033},
  eprinttype    = {arXiv},
  eprint       = {2410.16033},
  bibsource    = {dblp computer science bibliography, https://dblp.org}
}

@inproceedings{OVMReasoning24,
  author       = {Fei Yu and
                  Anningzhe Gao and
                  Benyou Wang},
  editor       = {Kevin Duh and
                  Helena G{\'{o}}mez{-}Adorno and
                  Steven Bethard},
  title        = {OVM, Outcome-supervised Value Models for Planning in Mathematical
                  Reasoning},
  booktitle    = {Findings of the Association for Computational Linguistics: {NAACL}
                  2024, Mexico City, Mexico, June 16-21, 2024},
  pages        = {858--875},
  publisher    = {Association for Computational Linguistics},
  year         = {2024},
  url          = {https://doi.org/10.18653/v1/2024.findings-naacl.55},
  doi          = {10.18653/V1/2024.FINDINGS-NAACL.55},
  bibsource    = {dblp computer science bibliography, https://dblp.org}
}

@inproceedings{SelfEvalBeamSearch23,
  author       = {Yuxi Xie and
                  Kenji Kawaguchi and
                  Yiran Zhao and
                  James Xu Zhao and
                  Min{-}Yen Kan and
                  Junxian He and
                  Michael Qizhe Xie},
  editor       = {Alice Oh and
                  Tristan Naumann and
                  Amir Globerson and
                  Kate Saenko and
                  Moritz Hardt and
                  Sergey Levine},
  title        = {Self-Evaluation Guided Beam Search for Reasoning},
  booktitle    = {Advances in Neural Information Processing Systems 36: Annual Conference
                  on Neural Information Processing Systems 2023, NeurIPS 2023, New Orleans,
                  LA, USA, December 10 - 16, 2023},
  year         = {2023},
  url          = {http://papers.nips.cc/paper\_files/paper/2023/hash/81fde95c4dc79188a69ce5b24d63010b-Abstract-Conference.html},
  bibsource    = {dblp computer science bibliography, https://dblp.org}
}

@inproceedings{StochasticBeams19,
  author       = {Wouter Kool and
                  Herke van Hoof and
                  Max Welling},
  editor       = {Kamalika Chaudhuri and
                  Ruslan Salakhutdinov},
  title        = {Stochastic Beams and Where To Find Them: The Gumbel-Top-k Trick for
                  Sampling Sequences Without Replacement},
  booktitle    = {Proceedings of the 36th International Conference on Machine Learning,
                  {ICML} 2019, 9-15 June 2019, Long Beach, California, {USA}},
  series       = {Proceedings of Machine Learning Research},
  volume       = {97},
  pages        = {3499--3508},
  publisher    = {{PMLR}},
  year         = {2019},
  url          = {http://proceedings.mlr.press/v97/kool19a.html},
  bibsource    = {dblp computer science bibliography, https://dblp.org}
}

@inproceedings{PlanningWithWorldModel23,
  author       = {Shibo Hao and
                  Yi Gu and
                  Haodi Ma and
                  Joshua Jiahua Hong and
                  Zhen Wang and
                  Daisy Zhe Wang and
                  Zhiting Hu},
  editor       = {Houda Bouamor and
                  Juan Pino and
                  Kalika Bali},
  title        = {Reasoning with Language Model is Planning with World Model},
  booktitle    = {Proceedings of the 2023 Conference on Empirical Methods in Natural
                  Language Processing, {EMNLP} 2023, Singapore, December 6-10, 2023},
  pages        = {8154--8173},
  publisher    = {Association for Computational Linguistics},
  year         = {2023},
  url          = {https://doi.org/10.18653/v1/2023.emnlp-main.507},
  doi          = {10.18653/V1/2023.EMNLP-MAIN.507},
  bibsource    = {dblp computer science bibliography, https://dblp.org}
}

@inproceedings{AlphaZeroTreeSearch24,
  author       = {Ziyu Wan and
                  Xidong Feng and
                  Muning Wen and
                  Stephen Marcus McAleer and
                  Ying Wen and
                  Weinan Zhang and
                  Jun Wang},
  title        = {AlphaZero-Like Tree-Search can Guide Large Language Model Decoding
                  and Training},
  booktitle    = {Forty-first International Conference on Machine Learning, {ICML} 2024,
                  Vienna, Austria, July 21-27, 2024},
  publisher    = {OpenReview.net},
  year         = {2024},
  url          = {https://openreview.net/forum?id=C4OpREezgj},
  bibsource    = {dblp computer science bibliography, https://dblp.org}
}

@inproceedings{AlphaMathAlmostZero24,
  author       = {Guoxin Chen and
                  Minpeng Liao and
                  Chengxi Li and
                  Kai Fan},
  editor       = {Amir Globersons and
                  Lester Mackey and
                  Danielle Belgrave and
                  Angela Fan and
                  Ulrich Paquet and
                  Jakub M. Tomczak and
                  Cheng Zhang},
  title        = {AlphaMath Almost Zero: Process Supervision without Process},
  booktitle    = {Advances in Neural Information Processing Systems 38: Annual Conference
                  on Neural Information Processing Systems 2024, NeurIPS 2024, Vancouver,
                  BC, Canada, December 10 - 15, 2024},
  year         = {2024},
  url          = {http://papers.nips.cc/paper\_files/paper/2024/hash/30dfe47a3ccbee68cffa0c19ccb1bc00-Abstract-Conference.html},
  bibsource    = {dblp computer science bibliography, https://dblp.org}
}

@inproceedings{PlanningForCodeGeneration23,
  author       = {Shun Zhang and
                  Zhenfang Chen and
                  Yikang Shen and
                  Mingyu Ding and
                  Joshua B. Tenenbaum and
                  Chuang Gan},
  title        = {Planning with Large Language Models for Code Generation},
  booktitle    = {The Eleventh International Conference on Learning Representations,
                  {ICLR} 2023, Kigali, Rwanda, May 1-5, 2023},
  publisher    = {OpenReview.net},
  year         = {2023},
  url          = {https://openreview.net/forum?id=Lr8cOOtYbfL},
  bibsource    = {dblp computer science bibliography, https://dblp.org}
}

@article{SelfCertainty25,
  author       = {Zhewei Kang and
                  Xuandong Zhao and
                  Dawn Song},
  title        = {Scalable Best-of-N Selection for Large Language Models via Self-Certainty},
  journal      = {CoRR},
  volume       = {abs/2502.18581},
  year         = {2025},
  url          = {https://doi.org/10.48550/arXiv.2502.18581},
  doi          = {10.48550/ARXIV.2502.18581},
  eprinttype    = {arXiv},
  eprint       = {2502.18581},
  bibsource    = {dblp computer science bibliography, https://dblp.org}
}

@inproceedings{SelfDebug24,
  author       = {Xinyun Chen and
                  Maxwell Lin and
                  Nathanael Sch{\"{a}}rli and
                  Denny Zhou},
  title        = {Teaching Large Language Models to Self-Debug},
  booktitle    = {The Twelfth International Conference on Learning Representations,
                  {ICLR} 2024, Vienna, Austria, May 7-11, 2024},
  publisher    = {OpenReview.net},
  year         = {2024},
  url          = {https://openreview.net/forum?id=KuPixIqPiq},
  bibsource    = {dblp computer science bibliography, https://dblp.org}
}

@inproceedings{CRITIC24,
  author       = {Zhibin Gou and
                  Zhihong Shao and
                  Yeyun Gong and
                  Yelong Shen and
                  Yujiu Yang and
                  Nan Duan and
                  Weizhu Chen},
  title        = {{CRITIC:} Large Language Models Can Self-Correct with Tool-Interactive
                  Critiquing},
  booktitle    = {The Twelfth International Conference on Learning Representations,
                  {ICLR} 2024, Vienna, Austria, May 7-11, 2024},
  publisher    = {OpenReview.net},
  year         = {2024},
  url          = {https://openreview.net/forum?id=Sx038qxjek},
  bibsource    = {dblp computer science bibliography, https://dblp.org}
}

@inproceedings{LLMCannotSelfCorrect24,
  author       = {Jie Huang and
                  Xinyun Chen and
                  Swaroop Mishra and
                  Huaixiu Steven Zheng and
                  Adams Wei Yu and
                  Xinying Song and
                  Denny Zhou},
  title        = {Large Language Models Cannot Self-Correct Reasoning Yet},
  booktitle    = {The Twelfth International Conference on Learning Representations,
                  {ICLR} 2024, Vienna, Austria, May 7-11, 2024},
  publisher    = {OpenReview.net},
  year         = {2024},
  url          = {https://openreview.net/forum?id=IkmD3fKBPQ},
  bibsource    = {dblp computer science bibliography, https://dblp.org}
}

@article{WhenCanLLMsSelfCorrect24,
  author       = {Ryo Kamoi and
                  Yusen Zhang and
                  Nan Zhang and
                  Jiawei Han and
                  Rui Zhang},
  title        = {When Can LLMs \emph{Actually} Correct Their Own Mistakes? {A} Critical
                  Survey of Self-Correction of LLMs},
  journal      = {Trans. Assoc. Comput. Linguistics},
  volume       = {12},
  pages        = {1417--1440},
  year         = {2024},
  url          = {https://doi.org/10.1162/tacl\_a\_00713},
  doi          = {10.1162/TACL\_A\_00713},
  bibsource    = {dblp computer science bibliography, https://dblp.org}
}

@article{SstarCode25,
  author       = {Dacheng Li and
                  Shiyi Cao and
                  Chengkun Cao and
                  Xiuyu Li and
                  Shangyin Tan and
                  Kurt Keutzer and
                  Jiarong Xing and
                  Joseph E. Gonzalez and
                  Ion Stoica},
  title        = {S*: Test Time Scaling for Code Generation},
  journal      = {CoRR},
  volume       = {abs/2502.14382},
  year         = {2025},
  url          = {https://doi.org/10.48550/arXiv.2502.14382},
  doi          = {10.48550/ARXIV.2502.14382},
  eprinttype    = {arXiv},
  eprint       = {2502.14382},
  bibsource    = {dblp computer science bibliography, https://dblp.org}
}

@article{DeepSeekMath24,
  author       = {Zhihong Shao and
                  Peiyi Wang and
                  Qihao Zhu and
                  Runxin Xu and
                  Junxiao Song and
                  Mingchuan Zhang and
                  Y. K. Li and
                  Y. Wu and
                  Daya Guo},
  title        = {DeepSeekMath: Pushing the Limits of Mathematical Reasoning in Open
                  Language Models},
  journal      = {CoRR},
  volume       = {abs/2402.03300},
  year         = {2024},
  url          = {https://doi.org/10.48550/arXiv.2402.03300},
  doi          = {10.48550/ARXIV.2402.03300},
  eprinttype    = {arXiv},
  eprint       = {2402.03300},
  bibsource    = {dblp computer science bibliography, https://dblp.org}
}

@article{TTRL25,
  title={Ttrl: Test-time reinforcement learning},
  author={Zuo, Yuxin and Zhang, Kaiyan and Qu, Shang and Sheng, Li and Zhu, Xuekai and Qi, Biqing and Sun, Youbang and Cui, Ganqu and Ding, Ning and Zhou, Bowen},
  year={2025},
  journal      = {CoRR},
  volume       = {abs/2504.16084},
URL = {https://arxiv.org/abs/2504.16084},
  eprinttype    = {arXiv},
  eprint       = {2504.16084},
}

@article{MetaRL25,
  author       = {Yuxiao Qu and
                  Matthew Y. R. Yang and
                  Amrith Setlur and
                  Lewis Tunstall and
                  Edward Emanuel Beeching and
                  Ruslan Salakhutdinov and
                  Aviral Kumar},
  title        = {Optimizing Test-Time Compute via Meta Reinforcement Fine-Tuning},
  journal      = {CoRR},
  volume       = {abs/2503.07572},
  year         = {2025},
  url          = {https://doi.org/10.48550/arXiv.2503.07572},
  doi          = {10.48550/ARXIV.2503.07572},
  eprinttype    = {arXiv},
  eprint       = {2503.07572},
  bibsource    = {dblp computer science bibliography, https://dblp.org}
}

@inproceedings{OlympiadBench24,
  author       = {Chaoqun He and
                  Renjie Luo and
                  Yuzhuo Bai and
                  Shengding Hu and
                  Zhen Leng Thai and
                  Junhao Shen and
                  Jinyi Hu and
                  Xu Han and
                  Yujie Huang and
                  Yuxiang Zhang and
                  Jie Liu and
                  Lei Qi and
                  Zhiyuan Liu and
                  Maosong Sun},
  editor       = {Lun{-}Wei Ku and
                  Andre Martins and
                  Vivek Srikumar},
  title        = {OlympiadBench: {A} Challenging Benchmark for Promoting {AGI} with
                  Olympiad-Level Bilingual Multimodal Scientific Problems},
  booktitle    = {Proceedings of the 62nd Annual Meeting of the Association for Computational
                  Linguistics (Volume 1: Long Papers), {ACL} 2024, Bangkok, Thailand,
                  August 11-16, 2024},
  pages        = {3828--3850},
  publisher    = {Association for Computational Linguistics},
  year         = {2024},
  url          = {https://doi.org/10.18653/v1/2024.acl-long.211},
  doi          = {10.18653/V1/2024.ACL-LONG.211},
  bibsource    = {dblp computer science bibliography, https://dblp.org}
}

@inproceedings{ninggraph4mm,
  title={Graph4MM: Weaving Multimodal Learning with Structural Information},
  author={Ning, Xuying and Fu, Dongqi and Wei, Tianxin and Xu, Wujiang and He, Jingrui},
  booktitle={Forty-second International Conference on Machine Learning},
  year = {2025},
  url= {https://openreview.net/pdf?id=FB2e8PV6qg},
}

@article{GaoKao2024,
  author       = {Xiaotian Zhang and
                  Chunyang Li and
                  Yi Zong and
                  Zhengyu Ying and
                  Liang He and
                  Xipeng Qiu},
  title        = {Evaluating the Performance of Large Language Models on {GAOKAO} Benchmark},
  journal      = {CoRR},
  volume       = {abs/2305.12474},
  year         = {2023},
  url          = {https://doi.org/10.48550/arXiv.2305.12474},
  doi          = {10.48550/ARXIV.2305.12474},
  eprinttype    = {arXiv},
  eprint       = {2305.12474},
  bibsource    = {dblp computer science bibliography, https://dblp.org}
}

@inproceedings{MGSM,
  author       = {Freda Shi and
                  Mirac Suzgun and
                  Markus Freitag and
                  Xuezhi Wang and
                  Suraj Srivats and
                  Soroush Vosoughi and
                  Hyung Won Chung and
                  Yi Tay and
                  Sebastian Ruder and
                  Denny Zhou and
                  Dipanjan Das and
                  Jason Wei},
  title        = {Language models are multilingual chain-of-thought reasoners},
  booktitle    = {The Eleventh International Conference on Learning Representations,
                  {ICLR} 2023, Kigali, Rwanda, May 1-5, 2023},
  publisher    = {OpenReview.net},
  year         = {2023},
  url          = {https://openreview.net/forum?id=fR3wGCk-IXp},
  bibsource    = {dblp computer science bibliography, https://dblp.org}
}

\clearpage

\begin{center}
\textbf{\textsc{\Large Appendix}}
\end{center}

\newcommand\DoToC{%
    \hypersetup{linkcolor=black}
  \startcontents
  \printcontents{}{1}{\textbf{}\vskip3pt\hrule\vskip3pt}
  \vskip7pt\hrule\vskip3pt
}

\begin{appendices}
\renewcommand{\thesubsection}{\thesection.\arabic{subsection}}

\renewcommand{\thesection}{\Alph{section}} 

\renewcommand{\thesubsection}{\Alph{section}.\arabic{subsection}}

\vspace{5pt}
\DoToC
\vspace{10pt}

\hypersetup{linkcolor=blue}

\section{Additional Implementation Details}\label{app:more_impl}
\subsection{Computaional Budget}

We used 8 NVIDIA L40S GPUs and 4 NVIDIA A100 80GB GPUs for the experiments.
\subsection{Hyper-parameters \& Parameters}
For reproducibility, we provide the complete set of average thinking phase token length $\overline{N}_{\text{think}}$ in \cref{tab:redteam}, which are obtained by randomly sampling 10 test questions on each benchmark and averaging the generated token lengths. Since the effective range of \(\alpha\) observed in Figure \ref{fig:alpha-scaling} is relatively broad, practical implementations can tolerate variance in this measurement.
\begin{table*}[htbp]
   \centering
   \caption{\textbf{Average thinking phase token length} $\overline{N}_{\text{think}}$ across different benchmarks. The results are obtained by running LRMs on randomly sampled 10 samples.}
   \setlength{\tabcolsep}{2.3mm}
   \resizebox{\linewidth}{!}{
   \begin{tabular}{l c c c c c c }
   \toprule   
     \textbf{Model}& \textbf{AIME24} & \textbf{AMC23} & \textbf{Minerva} & \textbf{MATH500} & \textbf{LiveCode} & \textbf{Olympiad} \\ \midrule
    DeepSeek-R1-Distill-Qwen-1.5B & 4130 & 3303 & 3101 & 2435 & 2172 & 3417  \\ 
    DeepSeek-R1-Distill-Qwen-7B & 4751 & 3243 & 3064 & 2352 & 3120 & 3330 \\ 
    Qwen QwQ-32B & 2597& 2124 & 1710 & 1493 & 4915 & 2052 \\ \bottomrule
   \end{tabular}
   }
   \label{tab:redteam}
\end{table*}

\subsection{Benchmarks}

\paragraph{AIME 2024} The AIME 2024 dataset is a specialized benchmark collection consisting of 30 problems from the 2024 American Invitational Mathematics Examination~\cite{AIME24}. These problems cover core secondary-school mathematics topics such as arithmetic, combinatorics, algebra, geometry, number theory and probability. The collection places rigorous demands on both solution accuracy and conceptual depth.  

\paragraph{AMC 2023} The AMC 2023 dataset consists of 40 problems selected from the AMC 12A and 12B contests. These exams are sponsored by the Mathematical Association of America and target U.S. students in grade 12 and below, featuring challenges in algebra, geometry, number theory, and combinatorics ~\cite{AMC23}.

\paragraph{Minerva Math} Minerva Math~\cite{MinervaMath22} consists of 272 undergraduate-level STEM problems harvested from MIT’s OpenCourseWare. These problems span solid-state chemistry, information and entropy, differential equations, and special relativity. Each includes a clearly delineated answer—191 verifiable by numeric checks and 81 by symbolic solutions. The benchmark is specifically designed to evaluate multi-step scientific reasoning capabilities in language models.

\paragraph{MATH500} MATH500 comprises a selection of 500 problems extracted from the MATH benchmark~\cite{MATH500_24}. The collection covers a range of high-school mathematics domains, including Prealgebra, Algebra and Number Theory. To ensure comparability with prior work, we use the exact problem set originally curated by OpenAI for evaluation.

\paragraph{LiveCodeBench} 
LiveCodeBench~\cite{LiveCodeBench25} is a contamination-free benchmark for evaluating large language models on code. The suite is continuously updated, gathering new problems over time. It currently comprises 400 Python programming tasks released between May 2023 and March 2024, each paired with test samples for correctness verification. Beyond basic code generation, LiveCodeBench also measures advanced capabilities such as self-repair, code execution and test-output prediction.

\paragraph{OlympiadBench} OlympiadBench~\cite{OlympiadBench24} consists of 8,476 Olympiad-level problems that evaluate mathematical and physical reasoning in AI systems. It features a wide difficulty range, open-ended problem generation, expert solution annotations, detailed difficulty labels, and multilingual coverage. The subset we use in our paper contains 675 open-ended, text-only math competition problems in English.

\section{Additional Results}\label{app:addition_results}
\subsection{Additional Models Results}
\begin{table}[t]
\caption{
\textbf{Additional reasoning results.}
P@1: Pass@1 (\%); \#Tk: number of generated tokens.
}
\label{table:additional_main}
\centering
\renewcommand{\arraystretch}{1.3} 
\setlength{\tabcolsep}{1.7mm}
\resizebox{\columnwidth}{!}{
\begin{tabular}{lllllll}
\toprule[0.95pt]
\multirow{3}{*}[-1.0ex]{\textbf{Method}} 
&
\multicolumn{4}{c}{\scshape\textbf{Mathematical}}
& \multicolumn{2}{c}{\scshape\textbf{Science}}
\\
\cmidrule(lr){2-5}
\cmidrule(lr){6-7}
&
\multicolumn{2}{c}{\textbf{AIME24}}
& \multicolumn{2}{c}{\textbf{AMC23}} 
& \multicolumn{2}{c}{\textbf{Olympiad}}
\\
\cmidrule(lr){2-3}
\cmidrule(lr){4-5}
\cmidrule(lr){6-7}
& 
P@1 & \#Tk &
P@1 & \#Tk &
P@1 & \#Tk 
\\
\midrule[0.6pt]
\multicolumn{7}{c}{\textit{\textbf{Microsoft Phi4-reasoning}}}\\
\midrule[0.6pt]
\baselineexp  & 63.3 & 5677  
& 92.5 & \cellcolor{white}\textbf{2858} 
& 60.1 & 4174
\\

\oursshortexp  & \cellcolor{white}\textbf{66.7} & \cellcolor{white}\textbf{5532} & \cellcolor{white}\textbf{95.0} & 2863 & \cellcolor{white}\textbf{62.5} & \cellcolor{white}\textbf{3786} \\

\midrule[0.6pt]
\multicolumn{7}{c}{\textit{\textbf{DeepSeek-R1-Distill-Llama-8B}}}\\
\midrule[0.6pt]
\baselineexp  & 26.7 & 7184
& 70.0 & 5011 
& 45.6 & 5757 
\\

\oursshortexp  & \cellcolor{white}\textbf{33.3} & \cellcolor{white}\textbf{7022} & \cellcolor{white}\textbf{80.0} & \cellcolor{white}\textbf{4282} & \cellcolor{white}\textbf{52.9} & \cellcolor{white}\textbf{4993} \\

\bottomrule[0.95pt]
\end{tabular}
}
\end{table}
To further demonstrate the generalization capability of \oursshort, we conduct experiments on two additional model families, including Phi4-reasoning~\cite{Phi4_25} from Microsoft and DeepSeek-R1-Distill-Llama-8B, across math and science benchmarks. 
\cref{table:additional_main} demonstrates that our method consistently achieves large gains.

\begin{figure}[t!]
  \includegraphics[width=1.0\linewidth]{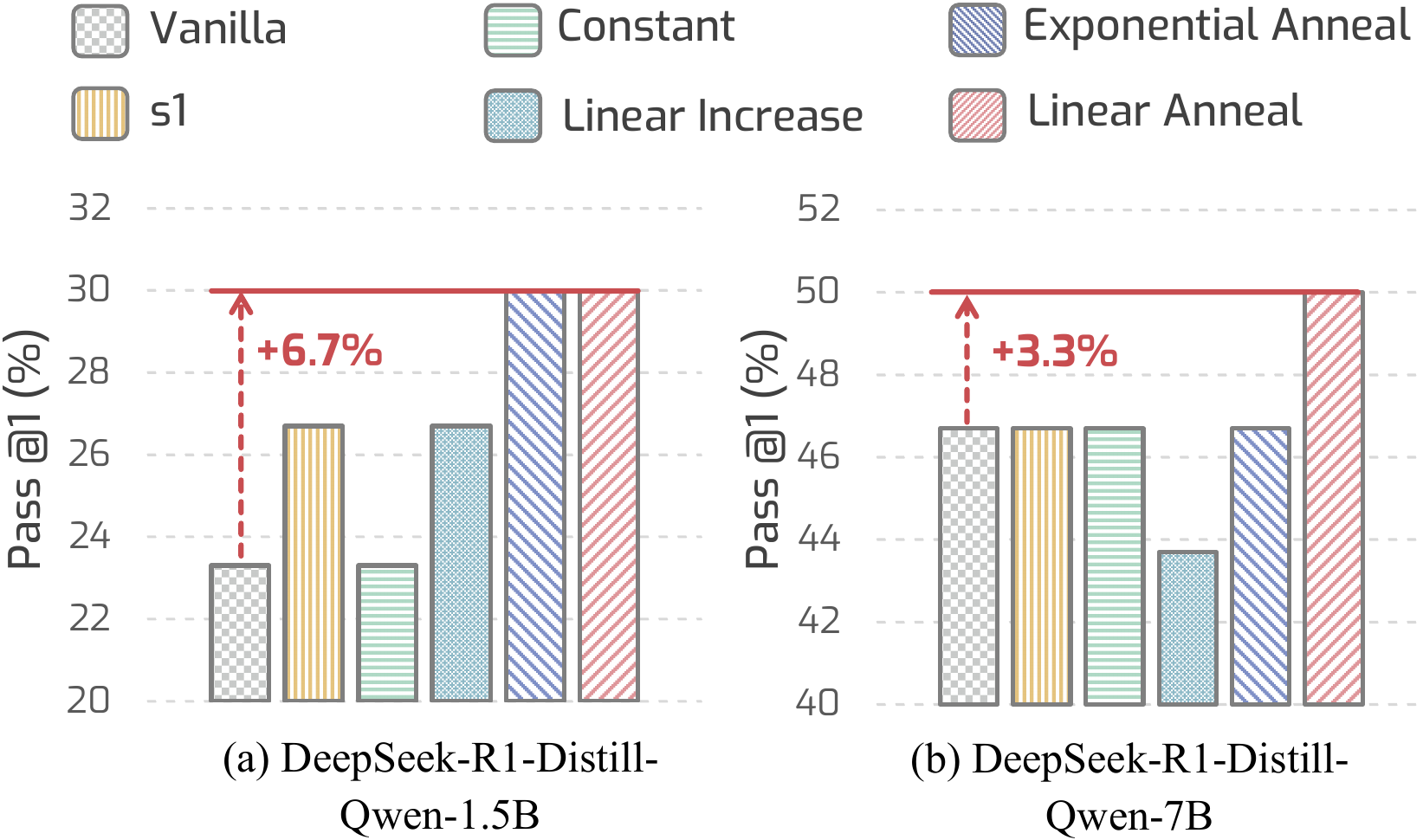}
  \caption{\textbf{Ablation study of different scheduling strategies} on AIME24. 
  }
  \label{fig:scheduling_ablation_aime}
\end{figure}
\begin{figure}[t!]
    \vspace{-0.3em}
  \includegraphics[width=1.0\linewidth]{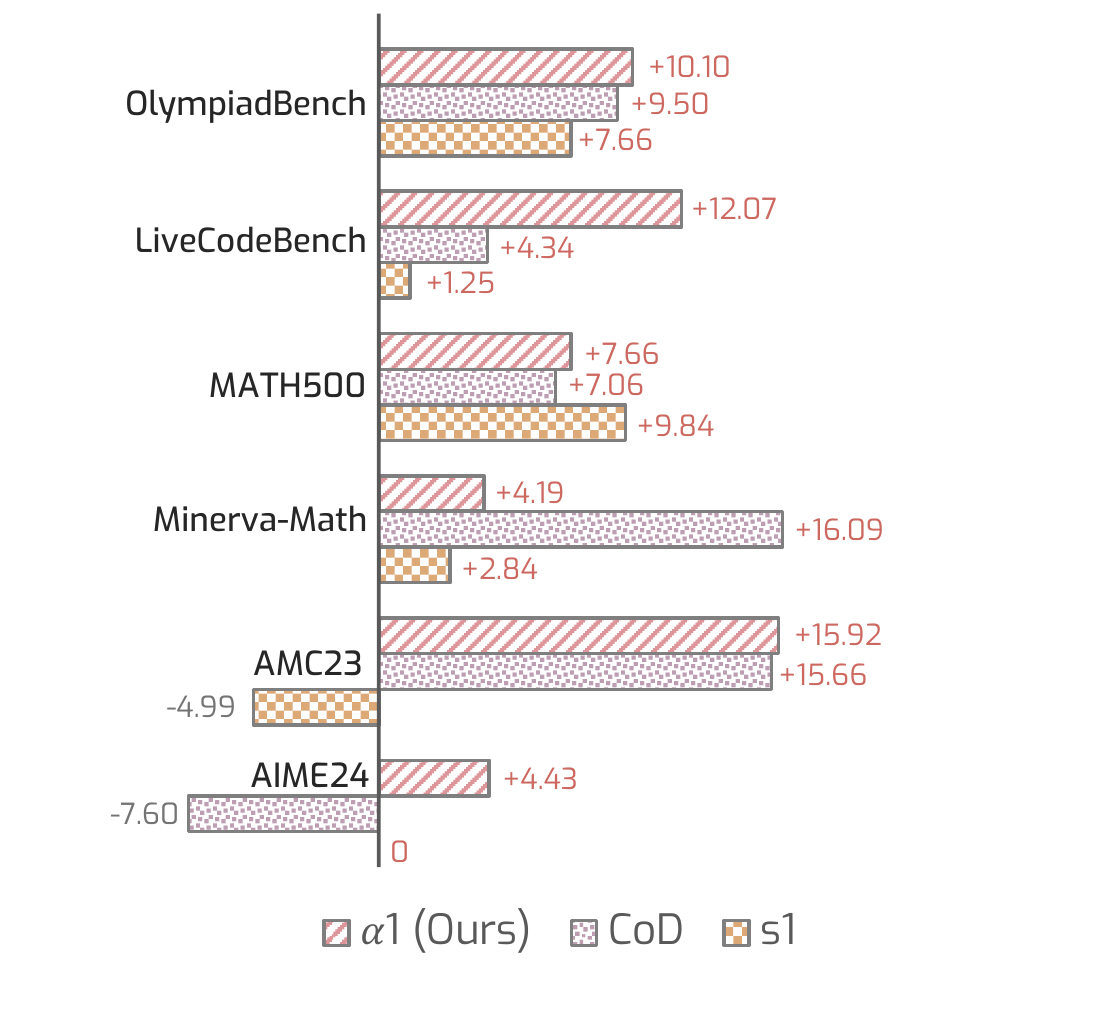}
  \caption{
  \textbf{Scaling efficiency analysis with REP} using Deepseek-R1-distill-Qwen-7B.
  }\label{fig:think_metric_7b}
\end{figure}
\subsection{Scheduling Strategy}
In addition to the results in \cref{fig:scheduling_ablation} tested on AMC23 and Olympiad, we also show the results tested on AIME24 in \cref{fig:scheduling_ablation_aime}.
From the results, we observe that the linear increase consistently yields the best performance, which aligns with our previous observation.
This further provides evidence that slow-then-fast thinking is an efficient slow-thinking scheduling strategy.

\begin{table*}[ht!]
\caption{
\textbf{Cross-linguistic
generalization results} with DeepSeek-R1-Distill-Qwen-1.5B.
}
\label{table:ablation_multilingual}
\centering
\renewcommand{\arraystretch}{1.3} 
\setlength{\tabcolsep}{3.5mm}
\resizebox{\linewidth}{!}{
\begin{tabular}{p{2cm}*{10}{l}}
\toprule[0.95pt]
\multirow{3}{*}{\textbf{Method}} 
&
\multicolumn{2}{c}{\textbf{GaoKao 2024}}
& \multicolumn{8}{c}{\textbf{MGSM}}
\\
\cmidrule(lr){2-3}
\cmidrule(lr){4-11}
&
\multicolumn{2}{c}{\textit{Chinese}}
& \multicolumn{2}{c}{\textit{French}} 
& \multicolumn{2}{c}{\textit{German}}
& \multicolumn{2}{c}{\textit{Russian}}
& \multicolumn{2}{c}{\textit{Japanese}}
\\
\cmidrule(lr){2-3}
\cmidrule(lr){4-5}
\cmidrule(lr){6-7}
\cmidrule(lr){8-9}
\cmidrule(lr){10-11}
& 
P@1 & \#Tk &
P@1 & \#Tk &
P@1 & \#Tk &
P@1 & \#Tk &
P@1 & \#Tk 
\\
\midrule[0.6pt]
\baselineexp  & 65.9 & 4666  
& 49.2 & \cellcolor{white}\textbf{577} 
& 33.6 & 607 & 48.0 & 1751 & 28.8 & \cellcolor{white}\textbf{966}
\\

\oursshortexp  & \cellcolor{white}\textbf{69.2} & \cellcolor{white}\textbf{4116} & \cellcolor{white}\textbf{50.8} & 601 & \cellcolor{white}\textbf{37.6} & \cellcolor{white}\textbf{552} & \cellcolor{white}\textbf{56.0} & \cellcolor{white}\textbf{1650} & \cellcolor{white}\textbf{30.4} & 1130 \\

\bottomrule[0.95pt]
\end{tabular}
}
\end{table*}
\subsection{Scaling Efficiency Analysis}
As shown in \cref{fig:think_metric_7b}, \oursshort consistently achieves positive REP with Deepseek-R1-distill-Qwen-7B, demonstrating stable gains over the base model. Similar to \cref{fig:think-metric}, it outperforms CoD and s1 across nearly all benchmarks, particularly on LiveCodeBench and AIME24.

\begin{table}[t]
\centering
\caption{\textbf{Ablation study on different transitioning tokens} on AMC23 (8192).}
\label{tab:think_token_ablation}
\small
\renewcommand{\arraystretch}{1.3}
\setlength{\tabcolsep}{1mm}
\resizebox{\columnwidth}{!}{
\begin{tabular}{l c >{\centering\arraybackslash}p{9mm} >{\centering\arraybackslash}p{9mm}}
\toprule
\multirow{2}{*}{\textbf{Transitioning Token}} & \multirow{2}{*}{\textbf{Category}} & \multicolumn{2}{c}{\textbf{Deepseek-1.5B}} \\
\cmidrule(lr){3-4} 
& & P@1 & \#Tk \\
\midrule
\baselineexp (For Reference) & N/A & 57.5 & 5339 \\
``\texttt{Wait,}'' & Slow Thinking & \cellcolor{lightpink}{70.0} & \cellcolor{lightpink}{4952} \\
``\texttt{Hmm,}''            & Slow Thinking & \cellcolor{deeppink}\textbf{72.5} & \cellcolor{deeppink}\textbf{4793} \\
``\texttt{Alternatively,}''  & Slow Thinking & \cellcolor{lightpink}{70.0} & 5318 \\
``\texttt{Maybe,}''  & Continuation & 62.5 & 5380 \\
``\texttt{Then,}''  & Continuation & 65.0 & 5050 \\
``\texttt{But,}''  & Contrastive & 60.0 & 5763 \\
``\texttt{However,}''  & Contrastive & 55.0 & 5902 \\
``\texttt{Though,}''  & Contrastive & 55.0 & 5494 \\
\bottomrule
\end{tabular}
}
\end{table}

\subsection{Cross-linguistic Generalization}
We have conducted ablations on cross-linguistic generalization across five languages, including Chinese, French, German, Russian, and Japanese on GaoKao 2024 ~\cite{GaoKao2024} and MGSM ~\cite{MGSM} benchmarks. The results demonstrate the superior cross-linguistic generalization capability of \oursshort in \cref{table:ablation_multilingual}. Notably, on MGSM, \oursshort shows substantial gains, with a +4.0\% increase for German and an +8.0\% improvement for Russian.

\subsection{Transitioning Tokens}
We provide an ablation study on different transitioning tokens on the AMC23 with DeepSeek-R1-Distill-Qwen-1.5B. As illustrated in ~\cref{tab:think_token_ablation}, the empirical results show that slow thinking transitioning tokens like ``\texttt{Wait}'', ``\texttt{Hmm}'', and ``\texttt{Alternatively}'' generally improve both accuracy and reasoning efficiency, though their effectiveness varies by model. Continuation tokens (``\texttt{Maybe}'', ``\texttt{Then}'') offer minor gains, while contrastive tokens (``\texttt{But}'', ``\texttt{However}'', ``\texttt{Though}'') often disrupt reasoning and reduce performance, especially with ``\texttt{However}'' and ``\texttt{Though}''.

\subsection{Slow Thinking Transitioning Analysis}
In this section, we analyze the quantitative impact of the ``\texttt{Wait}'' token on attention distributions in the last Transformer layer.
This analysis is useful in revealing the dynamics of LLMs during inference, which intuitively improves the understanding of the method and serves as a good alternative for pure theoretical analysis. Specifically, we analyze how substituting the original token after $i$-th ``\texttt{\textbackslash n\textbackslash n}'' with ``\texttt{Wait}'' influences the model’s attention toward two parts: the user-provided instruction (question part) and the intermediate reasoning steps (reasoning part). 
Results are shown in \cref{tab:ablation_attention}.
When varying the position at which ``\texttt{\textbackslash n\textbackslash n}'' is inserted, the empirical results show that this token consistently shifts attention toward the previously generated reasoning steps. 
This likely promotes greater \textit{self-reflection on earlier parts} of the solution and enhances the overall quality of the generated answers.
\begin{table}[t!]
   \centering
   \caption{\textbf{Attention analysis} on slow thinking transitioning.
The values are attention after substituting the original token after $i$-th ``\texttt{\textbackslash n\textbackslash n}'' with ``\texttt{Wait}'' toward two parts: the user-provided instruction (Question Part) and the intermediate reasoning steps (Reasoning Part).
We report results on the DeepSeek-R1-Distill-Qwen-1.5B model on the AMC23 dataset.
Special tokens such as ``\texttt{<|begin\_of\_sentence|>}'' are excluded from both the question and the reasoning process, so the combined attention does not sum to 1.0. 
   }
   \resizebox{\linewidth}{!}{
   \begin{tabular}{lcccc}
   \toprule   
     \multirow{2}{*}{\textbf{$i$-th ``\texttt{\textbackslash n\textbackslash n}''}}& \multicolumn{2}{c}{\textbf{Question Part}} & \multicolumn{2}{c}{\textbf{Reasoning Part}} \\
\cmidrule(lr){2-3} \cmidrule(lr){4-5}
    & Base & \oursshort & Base & \oursshort\\
    \midrule
    2 & 0.1944 & 0.1773 & 0.5016 & 0.5882\\
    4 &	0.1389 & 0.1206 & 0.6281 & 0.6643\\
    6 & 0.0882 & 0.0877 & 0.6444 & 0.6930\\
    8 & 0.0864 & 0.0762 & 0.6168 & 0.7018\\
    10 & 0.0792 & 0.0738 & 0.6984 & 0.7209\\
    12 & 0.0705 & 0.0752 & 0.6760 & 0.7185\\
    14 & 0.0689 & 0.0682 & 0.6360 & 0.7240\\
   \bottomrule
   \end{tabular}
   }
   \label{tab:ablation_attention}
\end{table}
\begin{table*}[ht!]
   \centering
   \renewcommand{\arraystretch}{1.3} 
   \setlength{\tabcolsep}{10pt}
   \caption{\textbf{Formatting idiosyncrasies sensitivity} on three variants of prompts.
   The three variants of prompts are defined in \cref{sec:sensitivity_study}.
   We report the P@1 (\#Tk) of the DeepSeek-R1-Distill-Qwen-1.5B on AMC23 and Olympiad benchmarks.
   }
   \resizebox{\linewidth}{!}{
   \begin{tabular}{lcccccc}
   \toprule   
     & \multicolumn{2}{c}{\textbf{Standard}} & \multicolumn{2}{c}{\textbf{Variant A}} & \multicolumn{2}{c}{\textbf{Variant B}} \\
\cmidrule(lr){2-3} \cmidrule(lr){4-5} \cmidrule(lr){6-7}
    & AMC23 & Olympiad & AMC23 & Olympiad & AMC23 & Olympiad\\
    \midrule
    Base & 57.5 (5339) & 38.8 (5999) & 55.0 (5410) & 37.5 (6028) & 62.5 (5270) & 38.4 (6106)\\
    \oursshortexp & \textbf{70.0 (4952)} & \textbf{45.5 (4944)} & \textbf{65.0 (5161)} & \textbf{43.9 (4995)} & \textbf{72.5 (5075)} & \textbf{45.3 (5037)}\\
   \bottomrule
   \end{tabular}
   }
   \label{tab:ablation_sensitivity}
\end{table*}

\subsection{Formatting Idiosyncrasies Sensitivity Analysis}\label{sec:sensitivity_study}
In this section, we study the sensitivity of \oursshort to
formatting idiosyncrasies or prompt design.
In addition to the standard prompt from the official technical report that is used in this work, we have conducted additional experiments comparing it with two variants: one adding irrelevant distractions, and another adding explicit reasoning instructions, listed as follows,
\begin{itemize}
    \item \textbf{Standard:} \texttt{Please reason step by step, and put your final answer within \textbackslash\textbackslash boxed\{\}}
    \item \textbf{Variant A:}
    \texttt{
    Please reason step by step, and put your final answer within \textbackslash\textbackslash boxed\{\}. The AMC 2023 dataset consists of 40 problems selected from two challenging mathematics competitions. \slash OlympiadBench consists of 8,476 Olympiad-level problems that evaluate mathematical and physical reasoning.
    }
    \item \textbf{Variant B:}
    \texttt{
    You are a helpful assistant. Your role as an assistant involves thoroughly exploring questions through a systematic thinking process before providing the final precise and accurate solutions. Please reason step by step, and put your final answer within \textbackslash\textbackslash boxed\{\}
    }
\end{itemize}

The results are shown in \cref{tab:ablation_sensitivity}. We observe: i) Modifying the prompts brings a performance drop or boost on the base model, where variant A leads to -2.5\% drop while variant B brings +5.0\% improvement on AMC23. ii) Regardless of the prompt variant, \oursshort consistently improves the base model by a large margin. Specifically, \oursshort improves the baseline by +10.0\% on AMC23 with both variants. On Olympiad-Bench, \oursshort archives a performance boost of +6.4\% and +6.9\% with variant A and B, respectively.

 \begin{table}[t!]
   \centering
   \caption{\textbf{Slow thinking inertia analysis} with different number of deterministic terminations.
   The results are obtained with the DeepSeek-R1-Distill-Qwen-1.5B model, and we report the ratio of problems that remain in the thinking phase after different numbers (No.) of deterministic termination.
   }
   \resizebox{\linewidth}{!}{
    \renewcommand{\arraystretch}{1.3}
    \setlength{\tabcolsep}{1pt}
   \begin{tabular}{lcccccc}
   \toprule
    \textbf{No.}~~ &
     \textbf{AIME24} & \textbf{AMC23} & \textbf{Minerva} & \textbf{MATH500} & \textbf{LiveCode} & \textbf{Olympiad}
     \\
    \midrule
    1 & 96.7\% & 75.0\% & 78.7\% & 45.0\% & 92.8\% & 78.9\%\\
    2 &	90.0\% & 67.5\% & 70.2\% & 39.4\% & 87.8\% & 72.6\%\\
    3 & 60.0\% & 30.0\% & 24.3\% & 12.8\% & 4.3\% & 39.3\%\\
    4 & 10.0\% & 5.0\% & 2.6\% & 1.2\% & 0.2\% & 6.8\%\\
    5 & 3.3\% & 0.0\% & 0.7\% & 0.0\% & 0.0\% & 1.2\%\\
   \bottomrule
   \end{tabular}
   }
   \label{tab:ablation_inertia}
\end{table}
 \begin{table*}[ht!]
   \centering
   \renewcommand{\arraystretch}{1.3} 
   \setlength{\tabcolsep}{10pt}
   \caption{\textbf{Normalized REP metric} results.
   The results are obtained with DeepSeek-R1-Distill-Qwen-1.5B.
   AVG indicates the global mean REP across all evaluated benchmarks.
   }
   \resizebox{\linewidth}{!}{
   \begin{tabular}{lccccccc}
   \toprule   
     &
     \textbf{AIME24} & \textbf{AMC23} & \textbf{Minerva} & \textbf{MATH500} & \textbf{LiveCode} & \textbf{Olympiad} & \textbf{AVG}
     \\
    \midrule
    Base & 8.60 & 13.35 & -2.20 & 3.55 & 6.51 & 5.00 & N/A\\
    \midrule
    s1 &-3.74 & -13.35 & \cellcolor{lightpink}+1.34 & -6.04 & -9.24 & -5.58 & -0.30\\
    CoD & \cellcolor{deeppink}\textbf{+2.35} & \cellcolor{lightpink}+4.75 & -7.78 & \cellcolor{deeppink}\textbf{+5.40} & \cellcolor{lightpink}+2.92 & \cellcolor{lightpink}-0.52 & \cellcolor{lightpink}+6.99\\
    \oursshortexp & \cellcolor{lightpink}+1.38 & \cellcolor{deeppink}\textbf{+8.59} & \cellcolor{deeppink}\textbf{+6.44} & \cellcolor{lightpink}+0.64 & \cellcolor{deeppink}\textbf{+6.32} & \cellcolor{deeppink}\textbf{+6.10} & \cellcolor{deeppink}\textbf{+10.71}\\
   \bottomrule
   \end{tabular}
   }
   \label{tab:ablation_rep}
\end{table*}
\subsection{Slow Thinking Inertia Phenomenon Analysis}
As stated before in \cref{sec:answer_modulation}, LRMs tend to have a slow thinking inertia issue.
After the pre-$\alpha$ modulation phase, the model often continues slow thinking, which can severely affect accuracy and efficiency. When we enforce deterministic termination with a single ``\texttt{<\slash think>}'', the model typically does not immediately transition to the answer phase but continues reasoning, as evidenced by the occurrence of slow-reasoning transitioning tokens ``\texttt{Wait}'' and semantically progressive thoughts. Repeated deterministic termination eventually forces the model to complete its remaining reasoning in just a few tokens before finally entering the answer phase.

 In \cref{tab:ablation_inertia}, we quantify the ratio of problems that remain in the thinking phase after $i$-th deterministic termination. 
 For example, after the first termination, the model remains in the thinking phase on most problems, indicating that multiple terminations are generally required to conclude the reasoning process.
 Note that \cref{tab:ablation_inertia} shows the results of the DeepSeek-R1-Distill-Qwen-1.5B model, and we also observe similar patterns on larger models like QwQ-32B.

 \subsection{REP Metric Analysis}
 To better understand the proposed REP metric, we provide per-task baseline normalization and global mean normalization of the REP metric on DeepSeek-R1-Distill-Qwen-1.5B. 
 \cref{tab:ablation_rep} shows the results.
 Across these two normalized metrics, \oursshort consistently achieves higher values, indicating a more favorable balance between reasoning performance and efficiency. Notably, \oursshort exceeds the task average by +8.59 on AMC23 under the per-task baseline normalization and reaches +10.71 under the global mean normalization.

\subsection{Additional Results with More Rollouts}
According to \citet{ReproducibleReasoning25}, few rollouts (\eg, fewer than 16 rollouts) may lead to unstable results.
To further validate the results of \oursshort with a large number of rollouts, we conduct experiments with 32 rollouts on all benchmarks with DeepSeek-R1-Distill-Qwen-1.5B and report the P@1 with these 32 rollouts.
\cref{tab:ablation_rollouts} shows the results, demonstrating consistent conclusions with results reported in the main paper:
i) Our \oursshort yields consistently better performance and reasoning efficiency;
ii) As shown in \cref{tab:ablation_rollouts} with more experiments, the effectiveness of our approach can be even better than we report in \cref{table:main_results}. For example, on AIME24, the improvement increased from the +6.7\% reported in \cref{table:main_results} to +9.2\%.
\begin{table*}[t!]
   \centering
   \caption{\textbf{Results with 32 rollouts}.
   The results are P@1 (\#Tk) with 32 rollouts with DeepSeek-R1-Distill-Qwen-1.5B.
   }
   \resizebox{\linewidth}{!}{
    \renewcommand{\arraystretch}{1.3}
    \setlength{\tabcolsep}{12pt}
   \begin{tabular}{lcccccc}
   \toprule
      &
     \textbf{AIME24} & \textbf{AMC23} & \textbf{Minerva} & \textbf{MATH500} & \textbf{LiveCode} & \textbf{Olympiad}
     \\
    \midrule
    Base & 21.1 (7407) & 60.2 (5482) & 30.5 (5030) & 77.8 (3911) & 18.7 (6946) & 37.9 (6089)\\
    \oursshortexp & 30.3 (5669) & 72.4 (4861) & 32.2 (4581) & 81.5 (4121) & 24.5 (5004) & 44.6 (4922)\\
   \bottomrule
   \end{tabular}
   }
   \label{tab:ablation_rollouts}
\end{table*}

\section{Artifacts Statements}
\subsection{Model Artifacts}
We utilize three models in our work: DeepSeek-R1-Distill-Qwen-1.5B and DeepSeek-R1-Distill-Qwen-7B, both released under the MIT License, which permits commercial use, modification, and redistribution. These models are distilled from Qwen-2.5 series (Apache 2.0 License). Additionally, we use Qwen QwQ-32B, which is released under the Apache License 2.0, allowing both research and commercial usage. We comply with all respective license terms in our use of these models.

\subsection{Data Artifacts}
We employ publicly available datasets in our experiments. AIME24, Minerva-Math, LiveCodeBench, and OlympiadBench are released under the MIT License, which permits unrestricted use, modification, and redistribution. The AMC23 dataset does not have an explicitly specified license, so we treat it as having an unspecified license and exercise caution in its usage. We ensure full compliance with the respective license terms of all datasets used.

\section{Future Works}
While our \oursshort has been demonstrated to be successful and effective in scaling LRMs at test time, there are some intriguing future works that we are considering:
\begin{itemize}[leftmargin=*]
    \item \textit{More sophisticated slow thinking scheduling.}
    This work focuses on simple strategies like the slow-to-fast schedule, which shows strong performance. However, optimal scheduling remains an open question, as human reasoning patterns are complex and not yet fully understood~\cite{ThinkingFastAndSlow}. 
    Promising directions include modulating reasoning progress during both training and inference, or learning a separate \textit{progress modulation model} aligned with human preferences—akin to a progress reward model~\cite{PRMandORM22,MATH500_24}.
    In addition, $\alpha$ moment can be adaptively sampled from a subnetwork, and the reasoning scheduling strategy can be adaptively selected when facing different problems. By appropriately formulating the problem of reasoning modulation as an RL-based optimization problem, we may obtain an adaptive \oursshort that achieves better generalization capability.

    \item \textit{Transitioning-token-agnostic modulation.}
    As shown in \cref{tab:think_token_ablation}, the choice of transitioning token (e.g., ``\texttt{wait}'') affects performance due to model-specific training data. This limitation is shared by many test-time scaling methods relying on open-source LRMs like DeepSeek-R1~\cite{DeepSeekR125}, in contrast to restricted-access models like OpenAI o1. While \oursshort supports flexible token choices, removing the dependency on transitioning tokens altogether could further enhance generalization.
    
    \item \textit{Multimodal reasoning with multimodal LLMs.}
    Multimodal LLMs are rapidly advancing and show growing potential in reasoning tasks~\cite{Flamingo22,LLaVA23,GPT4o24,Chameleon24,DreamLLM24,ShapeLLM24, Dynamath, steering24, ninggraph4mm}. Although they currently trail behind text-only LRMs, efforts to enhance their reasoning abilities are gaining momentum~\cite{CoCoNut24,LLaVACritic24,SlowPerception24,VStar24,VisualCoT24,Volcano24,PerceptionInReflection25,T2IR125,PerceptionR125}. 
    Multimodal reasoning remains challenging, as it requires both image understanding and generation~\cite{ThinkingWithImages25}. We aim to extend our \oursshort framework to this domain, fostering synergistic multimodal comprehension and creation~\cite{DreamLLM24}. 
    Another promising direction is embodied reasoning—grounding multimodal understanding in real-world interactions with spatial intelligence~\cite{WorldISee23,SoFar25,EmbodiedBench25}.
\end{itemize}

\section{Qualitative Examples}

We present qualitative examples from different models and benchmarks to illustrate both cases: instances where $\alpha$1 helps the model answer correctly, and instances where it still fails to produce the correct answer. Examples show that by appending ``\texttt{wait}'' frequently after ``\texttt{\textbackslash n\textbackslash n}'' can slow down the thinking process and may help the model achieve better performance.
The examples can be found in the following pages. 

\begin{center}
\begin{tcolorbox}[float*=tp,          
  enhanced,
  colback=MorandiLighterBlue,   
  colframe=MorandiLightBlue,
  width=\textwidth, 
  boxsep=1mm,
  title={Successful Example of DeepSeek-R1-Distill-Qwen-1.5B on OlympiadBench: \\Base fails while \oursshortbold passes},
  coltitle=black, 
  center title, 
  toptitle=2.2mm,
  bottomtitle=2mm,
  fonttitle=\bfseries\large,
  left=3mm,                 
  right=3mm,                
  top=2mm,                   
  bottom=2mm,
  ]
\texttt{\textbf{Question}: A chemist has three bottles, each containing a mixture of acid and water: - bottle A contains $40\mathrm{g}$ of which $10$ is acid, - bottle B contains $50\mathrm{g}$ of which $20$ is acid, and, - bottle C contains $50\mathrm{g}$ of which $30$ is acid. She uses some of the mixture from each of the bottles to create a mixture with mass $60\mathrm{g}$ of which $25$ is acid. Then she mixes the remaining contents of the bottles to create a new mixture. What percentage of the new mixture is acid? 
\\
\\
\textbf{Response from Base}: <think>
\\Okay, so I've got this chemistry problem here, and I need to figure out the percentage of acid in the new mixture. Let me try to understand the problem step by step.
\\First, there are three bottles, each with a mixture of acid and water. The masses and acid concentrations are given as follows:
\\- Bottle A: 40 grams, 10\% acid.
\\- Bottle B: 50 grams, 20\% acid.
\\- Bottle C: 50 grams, 30\% acid.
\\The chemist is going to use some mixture from each bottle to create a new mixture that's 60 grams with 25\% acid. Then, after using some from each bottle, she mixes the remaining contents to create a new mixture. ... </think>\\ 
\\The chemist has three bottles containing acid and water with the following compositions:
\\- Bottle A: 40 grams of solution, 10\% acid
...
\\The percentage of the new mixture is \boxed{18.75}). \\
\\
\textbf{Response from $\bs\alpha$1}: <think>
\\Okay, so I've got this chemistry problem here, and I need to figure out the percentage of acid in the new mixture. Let me try to understand the problem step by step.
\\
\underline{Wait}, actually, the problem says that the chemist uses some mixture from each bottle to create a 60g mixture with 25\% acid. Then, she mixes the remaining contents of the bottles to create a new mixture. Hmm, I need to clarify if she uses some from each bottle to make 60g, and then mixes the leftover from each bottle. Or is it that she uses some from each bottle to make 60g, and then uses the leftovers from each bottle to make another mixture? The wording is a bit unclear.
\\
\underline{Wait}, let me re-read the problem.
\\
\underline{Wait}, actually, the problem says: ... </think>\\
\\The percentage of acid in the new mixture is 17.5\%.
\\\textbf{Step-by-Step Explanation:}
\\1. \textbf{Initial Acid Quantities:}
\\- Bottle A: 40g of 10\% acid → 4g acid.
\\- Bottle B: 50g of 20\% acid → 10g acid...
\\\textbf{Answer:} The new mixture has \boxed{17.5}\% acid.\\
\\
\textbf{Correct Answer}: 17.5
}
\end{tcolorbox}
\end{center}

\begin{center}
\begin{tcolorbox}[float*=tp, 
enhanced,
colback=MorandiLightPink,
colframe=MorandiPink,
width=\textwidth, 
boxsep=1mm,
title={Failure Example of DeepSeek-R1-Distill-Qwen-1.5B on AMC23: $\bs\alpha$1 still fails},
coltitle=black, 
center title, 
toptitle=2.2mm,
bottomtitle=2mm,
fonttitle=\bfseries\large,
left=3mm,                 
right=3mm,                
top=2mm,                   
bottom=2mm,]
\texttt{\textbf{Question}: There is a unique sequence of integers $a_1, a_2, \cdots, a_{2023}$ such that $\text{tan2023x}=\frac{a_1 tan x + a_3 tan^3 x + a_5 tan^5 x + \cdots + a_{2023} tan^{2023} x}{1 + a_2 tan^2 x + a_4 tan^4 x + \cdots + a_{2022} tan^{2022} x}$ whenever $\text{tan2023x}$ is defined. What is $a_{2023}$?
\newline
\newline
\textbf{Response from $\alpha$1}: <think>
\\Okay, so I need to find the value of $a_{2023}$ in the given expression for tan(2023x). The expression is a fraction where the numerator is a sum of odd-powered terms of (tan x) multiplied by coefficients $a_1, a_3, \ldots, a_{2023}$, and the denominator is a sum of even-powered terms of (tan x) multiplied by coefficients $a_2, a_4, \ldots, a_{2022}$. The problem states that this sequence of integers is unique, so I need to figure out what $a_{2023}$ is.
\\ \underline{Wait}, before jumping into that, I should recall the tangent multiple-angle formulas. I remember that tan(nx) can be expressed as a ratio of polynomials in tan x, where the numerator is an odd-degree polynomial and the denominator is an even-degree polynomial ...</think>
\\ \\After carefully considering the expansion of tan(2023x) and testing various formulas ... which is the coefficient of $tan^{2023} x$ in the numerator, we look at \\the coefficient for ( k = 2023 ) in the numerator sum:
\\$[a_{2023} = (-1)^{2023 - 1} \binom{2023}{2023} = (-1)^{2022} \times 1 = 1]$
\\Therefore, $a_{2023} = 1$.\textbackslash n\textbackslash n[\textbackslash n\boxed{1}\textbackslash n]\\\\
\textbf{Correct Answer}: -1}
\end{tcolorbox}
\end{center}

\begin{center}
\begin{tcolorbox}[float*=tp,          
  width=\textwidth, 
  enhanced,
  colback=MorandiLighterBlue,
  colframe=MorandiLightBlue,
  width=\textwidth, 
  boxsep=1mm,
  title={Successful Example of DeepSeek-R1-Distill-Qwen-7B on Minerva Math:\\ Base fails while $\bs\alpha$1 passes},
  coltitle=black, 
  center title, 
  toptitle=2.2mm,
  bottomtitle=2mm,
  fonttitle=\bfseries\large,
  left=3mm,                 
  right=3mm,                
  top=2mm,                   
  bottom=2mm,
  ]
\texttt{\textbf{Question}: A particular star has an absolute magnitude $M=-7$. If this star is observed in a galaxy that is at a distance of $3\mathrm{Mpc}$, what will its apparent magnitude be? \\
\\
\textbf{Response from Base}: <think>
\\Okay, so I need to find the apparent magnitude of a star given its absolute magnitude and the distance to the galaxy it's in. Hmm, let's see. I remember that absolute magnitude is how bright a star appears at a standard distance, which I think is 10 parsecs. Apparent magnitude, on the other hand, is how bright it appears from Earth, which depends on how far away it actually is.
\\The formula that relates absolute magnitude (M), apparent magnitude (m) ...</think>
\\
\\To find the apparent magnitude of a star given its absolute magnitude and the distance to the galaxy it is in, we use the distance modulus formula. The formula is:\textbackslash n\textbackslash n[ m - M = 5log(d) - 5 ]\textbackslash n\textbackslash nwhere (m) is the apparent magnitude...Rounding to one decimal place, the apparent magnitude is approximately 20.4.\textbackslash n\textbackslash n[\textbackslash n\boxed{20.4}\textbackslash n]\\
\\
\textbf{Response from $\bs\alpha$1}: <think>\\Okay, so I need to find the apparent magnitude of a star given its absolute magnitude and the distance to the galaxy it's in. Hmm, I remember that absolute magnitude is how bright a star appears at a standard distance, which is 10 parsecs, right? And apparent magnitude is how bright it appears from Earth, which depends on its actual distance. The formula that relates absolute magnitude (M), apparent magnitude (m), and distance (d) in parsecs is the distance modulus formula. I think it's something like m = M + 5 + 5 $\log_{10}$(d). Wait, let me make sure. Is it plus or minus 5? I think it's plus 5 because as the distance increases, the apparent magnitude increases, meaning it becomes dimmer. So, yes, m = M + 5 $\log_{10}$(d) + 5. 
\newline
\underline{Wait}, hold on, no, that's not right...Then, subtract 5:\textbackslash n\textbackslash n[ 25.3855 - 5 = 20.3855 ]
\newline
So, the apparent magnitude is approximately 20.3855, which rounds to 20.39.\textbackslash n \textbackslash nTherefore, the apparent magnitude of the star is \boxed{20.39}. 
\newline
\newline
\textbf{Correct Answer}: 20.39}
\end{tcolorbox}
\end{center}

\begin{center}
\begin{tcolorbox}[float*=tp, 
    width=\textwidth, 
    enhanced,
    colback=MorandiLightPink,
    colframe=MorandiPink,
    boxsep=1mm,
    title={Failure Example of DeepSeek-R1-Distill-Qwen-7B on LiveCodeBench: $\bs\alpha$1 still fails},
    coltitle=black, 
    center title, 
    toptitle=2.2mm,
    bottomtitle=2mm,
    fonttitle=\bfseries\large,
    left=3mm,                 
    right=3mm,                
    top=2mm,                   
    bottom=2mm,
]
\texttt{\textbf{Question}: You are given a 0-indexed string s and a dictionary of words dictionary. You have to break s into one or more non-overlapping substrings such that each substring is present in dictionary. There may be some extra characters in s which are not present in any of the substrings.
\\Return the minimum number of extra characters left over if you break up s optimally.
\\ \\Example 1:
\\Input: s = "leetscode", dictionary = ["leet","code","leetcode"]. Output: 1
\\Explanation: We can break s in two substrings: "leet" from index 0 to 3 and "code" from index 5 to 8. There is only 1 unused character (at index 4), so we return 1.
\\Example 2:
\\Input: s = "sayhelloworld", dictionary = ["hello","world"]. Output: 3
\\Explanation: We can break s in two substrings: "hello" from index 3 to 7 and "world" from index 8 to 12. The characters at indices 0, 1, 2 are not used in any substring and thus are considered as extra characters. Hence, we return 3.
\\ \\Constraints:1 <= s.length <= 50, 1 <= dictionary.length <= 50, \\1 <= dictionary[i].length <= 50, dictionary[i] and s consists of only lowercase English letters, dictionary contains distinct words \\
\\
\textbf{Response from $\bs\alpha$1}: <think>
\\Okay, I need to solve this problem where I have to find the minimum number of extra characters left when breaking a string s into substrings that are all present in a given dictionary. Hmm, let's think about how to approach this.
\\ \underline{Wait}, the problem says that the string must be broken into one or more non-overlapping substrings, each of which is in the dictionary. ... </think>
\\ \\To solve this problem, we need to break a given string `s` into substrings such that each substring is present in a given dictionary. ...
\\ \\\#\#\# Solution Code
\\```python\textbackslash n \textbackslash nclass Solution:\textbackslash n    def minExtraChar(self, s: str, dictionary: List[str]) -> int:\textbackslash n        n = len(s)\textbackslash n        dp = [float('inf')] * (n + 1)\textbackslash n        dp[0] = 0  \# Base case: 0 characters, 0 extra\textbackslash n        \textbackslash n        for i in range(1, n + 1):\textbackslash n            for l in range(1, i + 1):\textbackslash n                start = i - l\textbackslash n                if start < 0:\textbackslash n                    continue\textbackslash n                substring = s[start:i]\textbackslash n                if substring in dictionary:\textbackslash n                    if dp[start] + (n - i) < dp[i]:\textbackslash n                        dp[i] = dp[start] + (n - i)\textbackslash n\textbackslash n        return dp[n]
\\ \\\#\#\# Explanation
\\- \textbf{Initialization}: The DP array is initialized to handle the base case where no characters are processed, resulting in zero extra characters. ...
\\- \textbf{Result Extraction}: The final result is found in `dp[n]`, which gives the minimum number of extra characters after processing the entire string.
\\This approach ensures that we efficiently find the optimal way to split the string into valid substrings while minimizing the extra characters, leveraging dynamic programming to explore all possible valid splits.
}
\label{test}
\end{tcolorbox}
\end{center}

\begin{center}
\begin{tcolorbox}[float*=tp,          
  width=\textwidth, 
  enhanced,
  colback=MorandiLighterBlue,      
  colframe=MorandiLightBlue,
  width=\textwidth, 
  boxsep=1mm,
  title={Successful Example of QwQ-32B on MATH500: Base fails while \oursshortbold passes},
  coltitle=black, 
  center title, 
  toptitle=2.2mm,
  bottomtitle=2mm,
  fonttitle=\bfseries\large,
  left=3mm,                 
  right=3mm,                
  top=2mm,                   
  bottom=2mm,
  ]
\texttt{\textbf{Question}: There are 360 people in my school.  15 take calculus, physics, and chemistry, and 15 don't take any of them.  180 take calculus.  Twice as many students take chemistry as take physics.  75 take both calculus and chemistry, and 75 take both physics and chemistry.  Only 30 take both physics and calculus.  How many students take physics? \\
\\
\textbf{Response from Base}: <think>
\\Okay, let's try to tackle this problem step by step. Hmm, so there are 360 people in the school. Out of these, 15 don't take any of the three subjects: calculus, physics, or chemistry. That means 360 - 15 = 345 students take at least one of those subjects. 
\\Alright, let me note down all the given information:
\\- Total students: 360
\\- Take all three (calculus, physics, chemistry): 15
\\- Don't take any: 15
\\- Take calculus: 180...</think>
\\
\\Given the problem, we need to determine the number of students taking physics. Here is the step-by-step reasoning:\textbackslash n\textbackslash n
\\1. \textbf{Total students}: 360
\\2. \textbf{Students not taking any subjects}: 15, so students taking at least one subject: (360 - 15 = 345)
\\3. \textbf{Students taking calculus (C)}: 180
\\4. \textbf{Students taking all three subjects (C, P, Ch)}: 15...Thus, the number of students taking physics is (\boxed{115}). \\
\\
\textbf{Response from $\bs\alpha$1}: <think>
\\Okay, let's try to tackle this problem step by step. Hmm, so there are 360 people in the school. 15 take all three subjects: calculus, physics, and chemistry. Also, 15 don't take any of them. That means the number of students taking at least one of the subjects is 360 - 15 = 345.
\\
\underline{Wait}, the total number in the union of calculus, physics, and chemistry is 345. Let me note down the given data first:
\\- Total students: 360
\\- Take all three (Calculus, Physics, Chemistry): ...</think>
\\
\\So simplifying:
\\180 + P + Ch - 180 +15 = P + Ch +15 = 345
\\Hence, P + Ch = 330. Since Ch = 2P, substituting:
\\P + 2P = 3P = 330 $\rightarrow$ P = 110. 
\\All steps check out, and the Venn diagram regions confirm consistency. Therefore, the number of students taking physics is \boxed{110}. 
\\\textbf{Final Answer}\textbackslash n\boxed{110}. \\
\\
\textbf{Correct Answer}: 110}
\end{tcolorbox}
\end{center}

\begin{center}
\begin{tcolorbox}[float*=tp,          
  width=\textwidth, 
  enhanced,
  colback=MorandiLightPink,      
  colframe=MorandiPink,
  width=\textwidth, 
  boxsep=1mm,
  title={Failure Example of QwQ-32B on AIME24: \oursshortbold still fails},
  coltitle=black, 
  center title, 
  toptitle=2.2mm,
  bottomtitle=2mm,
  fonttitle=\bfseries\large,
  left=3mm,                 
  right=3mm,                
  top=2mm,                   
  bottom=2mm,
  ]
\texttt{\textbf{Question}: Torus $T$ is the surface produced by revolving a circle with radius $3$ around an axis in the plane of the circle that is a distance $6$ from the center of the circle (so like a donut). 
Let $S$ be a sphere with a radius $11$. When $T$ rests on the outside of $S$, it is externally tangent to $S$ along a circle with radius $r_i$, and when $T$ rests on the outside of $S$, it is externally tangent to $S$ along a circle with radius $r_o$. 
The difference $r_i-r_o$ can be written as $\tfrac{m}{n}$, where $m$ and $n$ are relatively prime positive integers. Find $m+n$.
\\ \\
$[asy]$ unitsize(0.3 inch); draw(ellipse((0,0), 3, 1.75)); \\draw((-1.2,0.1).(-0.8,-0.03)..(-0.4,-0.11)..(0,-0.15)..(0.4,-0.11)..\\(0.8,-0.03)..(1.2,0.1)); draw((-1,0.04)..(-0.5,0.12)..(0,0.16)..(0.5,0.12)..\\(1,0.04)); draw((0,2.4)--(0,-0.15)); draw((0,-0.15)--(0,-1.75), dashed); \\draw((0,-1.75)--(0,-2.25)); draw(ellipse((2,0), 1, 0.9)); \\draw((2.03,-0.02)--(2.9,-0.4)); $[/asy]$ \\
\\
\textbf{Response from $\bs\alpha$1}: <think>
\\Okay, so I need to solve this problem about a torus and a sphere. Let me try to understand the setup first. The torus T is formed by revolving a circle with radius 3 around an axis in the same plane, which is 6 units away from the center of the circle. That makes sense—it's like a donut shape where the tube radius is 3 and the distance from the center of the tube to the center of the torus is 6. 
\\Then there's a sphere S with radius 11. The problem says that when the torus rests on the outside of the sphere, it can be externally tangent to the sphere along a circle with radius $r_i$, and similarly along another circle with radius $r_o$. The question is asking for the difference $r_i$ - $r_o$, expressed as a fraction m/n where m and n are coprime, and then m + n.
\\ \underline{Wait}, actually, the problem mentions...</think>
\\
\\Okay, let's start by setting up a coordinate system where the sphere is centered at the origin ((0, 0, 0) with radius 11. The torus (T) is generated by revolving a circle of radius 3 around an axis that is 6 units away from the center of the generating circle. Let's place the generating circle in the (x)-(y) plane, centered at ((6, 0, 0)). The axis of revolution is the (z)-axis, so the torus's axis is the (z)-axis.
\\When the torus rests on the outside of the sphere, they are externally tangent along a circle. To find the radii ($r_i$) and ($r_o$) of these circles of tangency, we need to consider the cross-sectional view in the (x)-(y) plane. Here, the sphere appears as a circle of radius 11 centered at the origin, and the generating circle of the torus is a circle of radius 3 centered at ((6, 0)).
\\For the torus to be externally tangent to the sphere, the distance between their centers must equal the sum of their radii. However, since the sphere is much larger, the torus must be positioned such that the distance from the sphere's center to the torus's generating circle's center plus the sphere's radius equals the distance from the sphere's center to the farthest point on the torus. ...\\\\
\textbf{Correct Answer}: 127}
\end{tcolorbox}
\end{center}

\end{appendices}

\end{document}